\pdfoutput=1
\documentclass[preprint,11pt]{elsarticle}
\usepackage[margin=1in]{geometry}
\usepackage{amsmath,amssymb,amsthm,bm,booktabs,graphicx,float,placeins,array}
\usepackage[T1]{fontenc}
\usepackage{lmodern}
\usepackage{xcolor}
\usepackage[hidelinks]{hyperref}
\floatstyle{ruled}\newfloat{algorithm}{tbp}{loa}\floatname{algorithm}{Algorithm}
\floatstyle{plain}

\newcommand{\R}{\mathbb R}
\newcommand{\norm}[1]{\left\lVert#1\right\rVert}
\newcommand{\dd}{\mathrm d}
\newtheorem{theorem}{Theorem}
\newtheorem{proposition}{Proposition}
\newtheorem{lemma}{Lemma}
\theoremstyle{remark}
\newtheorem{remark}{Remark}

\biboptions{sort&compress}
\makeatletter
\def\ps@pprintTitle{\let\@oddhead\@empty\let\@evenhead\@empty
  \def\@oddfoot{\hfil\thepage\hfil}\let\@evenfoot\@oddfoot}
\makeatother
\date{November 2026}

\begin{document}
\begin{frontmatter}

\title{Escaping two moving sensing disks: Explicit characteristics and
the complete value of a two-pursuer detection game}

\author[cicese]{Johnny Rubio}
\author[cicese]{Ubaldo Ruiz\corref{cor1}}
\ead{uruiz@cicese.edu.mx}
\cortext[cor1]{Corresponding author.}
\affiliation[cicese]{organization={Centro de Investigaci\'on Cient\'ifica
  y de Educaci\'on Superior de Ensenada (CICESE)},
  city={Ensenada},
  state={Baja California},
  country={Mexico}}

\begin{abstract}
We study a pursuit--evasion game in which two pursuers, each carrying a
sensing disk of radius \(R\), try to keep a faster evader detected, while
the evader tries to leave the union of the two disks as early as
possible. All players have simple motion, and the evader is \(\mu>1\)
times faster than each pursuer. We show that the value of the game is a
geometric escape time: expand each sensing disk by the distance its
pursuer can travel, and take the first time at which the evader's
reachable disk strictly meets the region outside both expanded disks. We
give an exact formula and a finite algorithm for this time. Regular
optimal plays are straight lines in the inertial frame, given in closed
form: a radial race against one pursuer, or a simultaneous exit through
a corner of the sensing union, possibly completed by an instantaneous
evader turn at contact. For pursuer half-separations
\(a\le R\sqrt{1-\mu^{-2}}\) the strategies of both sides are explicit. For
arbitrary separations we partition the state space completely, including
a fold, its cusp curve, and states at which the value jumps, and a
viability argument proves that the escape time is the value for every
initial state, whether the pursuers react to the evader or the evader to
the pursuers.
\end{abstract}

\begin{keyword}
Pursuit--evasion games \sep differential games \sep multiple pursuers
\sep sensing \sep method of characteristics \sep viability theory \sep
discriminating domains
\end{keyword}

\end{frontmatter}

\section{Introduction}
\label{sec:intro}
Teams of mobile sensors are routinely asked to keep an intruder under
observation. Detection is lost only when the intruder is outside the
sensing range of \emph{every} team member, so the members can cooperate:
one may hand detection to another, or several may close in so that the
intruder's best exit moves away. When the intruder is faster than each
sensor platform, it eventually escapes, and the natural question is how
long the team can delay the escape and how both sides should move.

This paper answers that question exactly for the smallest nontrivial
team. Two pursuers move at unit speed, each carrying a sensing disk of
radius \(R\), and we write \(a\) for half the distance between them.
The evader moves at speed \(\mu>1\) and wins once it is
outside both disks; the escape time is the payoff, which the evader
minimizes and the pursuers maximize.

\begin{figure}[htbp]\centering
\includegraphics[width=\linewidth]{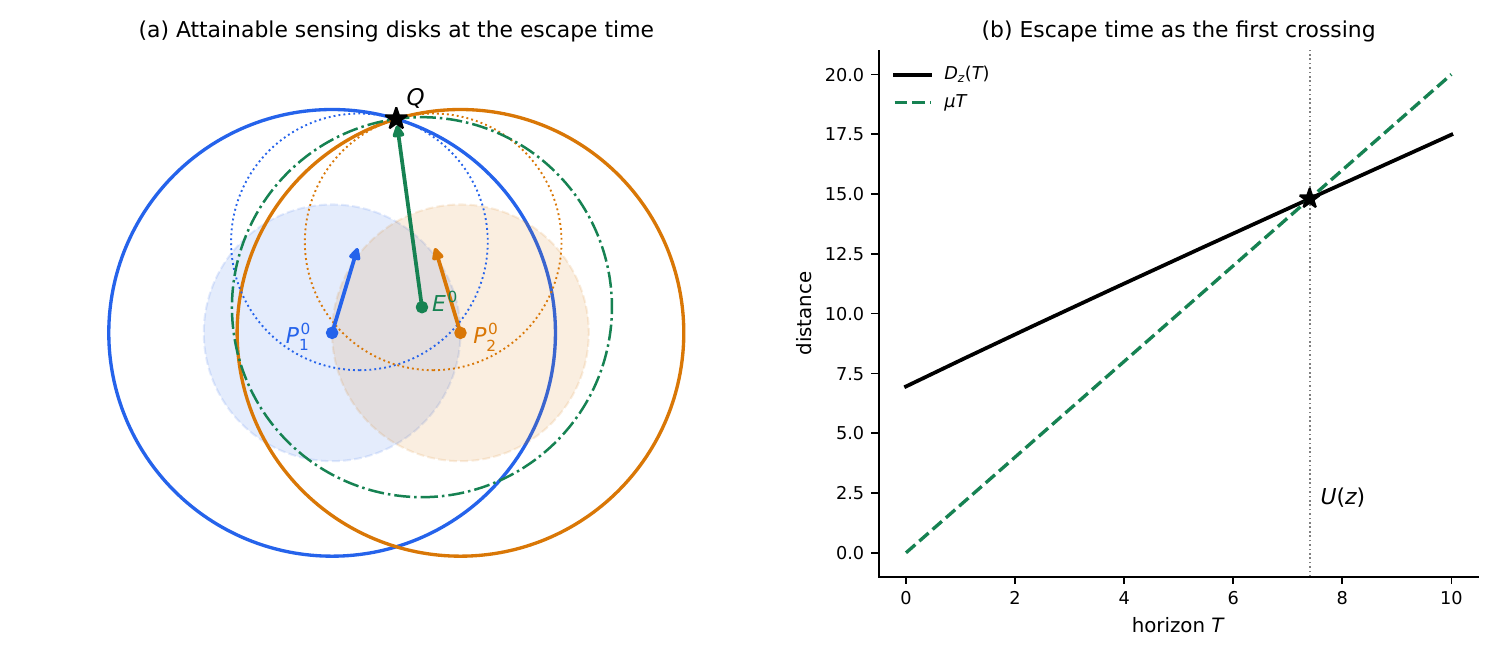}
\caption{The attainable-disk construction for \(R=10\), \(\mu=2\),
initial pursuer positions \(P_{1,2}^0=(\mp5,0)\), and initial evader
position \(E^0=(2,2)\). (a) Initial sensing disks
(shaded, dashed), expanded disks of radius \(R+T\) at the escape time
(solid), the evader's reach circle of radius \(\mu T\) (dash-dot), and
the optimal straight paths. The pursuers end at the dotted sensing
circles, which pass through the escape corner \(Q\). (b) The escape time
is the first horizon at which the evader's reach \(\mu T\) exceeds the
distance \(D(T)\) to the exterior of the expanded disks; here
\(U\approx7.402\) (Section~\ref{sub:distance}).}
\label{fig:geometry}
\end{figure}

\paragraph{Main result}
The answer has a short geometric form. After time \(T\), each pursuer is
within distance \(T\) of its initial position, so it cannot sense
anything outside the \emph{expanded disk} of radius \(R+T\) about that
position. Let \(D(T)\) be the distance from the evader's initial position
to the region outside both expanded disks (Figure~\ref{fig:geometry}).
The evader, moving at speed \(\mu\), can be in that region at time \(T\)
when \(D(T)<\mu T\). Define the \emph{escape time}
\[
 U=\inf\{T\ge0:\ D(T)<\mu T\}.
\]
\begin{quote}
\emph{For every speed ratio \(\mu>1\), every sensing radius, and every
initial state, \(U\) is the value of the game, both when the pursuers
react to the evader and when the evader reacts to the pursuers
(Theorems~\ref{thm:globalvalue} and~\ref{thm:reverse}).}
\end{quote}
The strict inequality in the definition matters. The difference
\(D(T)-\mu T\) can touch zero and rise again before it finally becomes
negative (Figure~\ref{fig:residual}(b)), and such a contact gives the
evader no escape. Much of the paper is about the states at which this
happens.

\paragraph{Related work}
Differential games of simple motion go back to Isaacs~\cite{isaacs},
whose method of characteristics constructs candidate optimal plays
backward from the terminal surface; standard references
are~\cite{basar,bardi}. Games with several pursuers have been studied
extensively, mostly with capture rather than loss of detection as the
terminal event. Isaacs' two cutters and fugitive ship problem was solved
in~\cite{pachter2019}, the multi-pursuer single-evader game was analyzed
geometrically in~\cite{vonmoll}, and closely related target-guarding and
active-defense games were solved in~\cite{guarding,garcia};
\cite{weintraub} surveys this line of work, and \cite{garciampme} treats
teams of several pursuers and several evaders. Cooperative encirclement
strategies for capturing or containing a faster evader have also been
studied~\cite{fang,garciabopardikar}; in the present game the faster
evader always escapes, and the question is when. For larger teams,
reach-avoid formulations combine Hamilton--Jacobi
reachability~\cite{mitchell} with pairwise decompositions~\cite{chen},
and reachable-set methods have been applied to multiple pursuers in
flowfields~\cite{sun}. Time-optimal capture strategies are also known for
nonholonomic pursuers~\cite{ruiz2013}. Games whose payoff concerns
sensing rather than capture form a smaller literature. In the
surveillance--evasion game of Lewin and Breakwell~\cite{lewin}, a single
pursuer tries to keep the evader inside its detection circle, and
visibility-based games ask a pursuer to keep line of sight among
obstacles~\cite{bhattacharya}. Recent work treats single-pursuer
surveillance with nonholonomic players and other sensor footprints: a
differential drive robot that keeps an omnidirectional agent within its
sensing range, including when the agent is
faster~\cite{ruiz2022,saavedra}; two differential drive robots with a
limited field of view~\cite{ruiz2024}; a differential drive robot
escaping a circular detection region~\cite{ruiz2025}; and an evader with
bounded turning rate escaping the surveillance disk of an agile
pursuer~\cite{braun}. All of these have a single sensing platform. The
closest work to ours is~\cite{braun2p}, which extends the game
of~\cite{braun} to two pursuers carrying sensing disks. For an evader
with bounded turning rate it gives partial results, for stationary
pursuers and for pursuers at most half as fast as the evader, and leaves
the complete solution open. Here all three players have simple motion,
and we determine the value for every speed ratio \(\mu>1\) and every
initial state, including the fold states at which the earliest contact
precedes the escape. In the
present game the detection region
is the union of two pursuers' sensing disks, so the pursuers can hand
detection to one another, and the evader may have to leave both disks at
once.

Turning candidate plays into a proven value requires a precise notion
of strategy, such as the nonanticipative strategies of Elliott and
Kalton~\cite{elliottkalton}, and an argument that one side can enforce
the candidate time. Viability theory~\cite{aubin} and the theory of
discriminating domains~\cite{cqs,cardaliaguet96,cardaliaguetwp} supply such arguments:
a closed set is a discriminating domain if and only if the player who
wants to remain in it has a nonanticipative strategy to do so, provided
that player's velocity sets are convex.

The two-disk game has four features that make a direct application of
these tools delicate. First, the target is the complement of a
\emph{union}, so the evader may have to leave both disks at the same
instant, at a corner of the sensing region. Second, in the natural
reduced coordinates the dynamics are nonlinear because the pursuer
baseline rotates. Third, the escape time is not always the first root
of the scalar equation \(D(T)=\mu T\): some roots are isolated tangencies
that the evader cannot exploit, and the value jumps across the
corresponding states (Figure~\ref{fig:residual}(b)). Fourth, the fixed-speed control sets are circles, which are not
convex, so standard viability theorems do not apply directly.

\paragraph{Contributions}
\begin{enumerate}
\item \emph{The geometric escape time and its optimal plays.} We give the
distance \(D\) in closed form, so that \(U\) is a computable scalar
quantity (Section~\ref{sec:distance}). On regular optimal arcs the
active players move along straight lines in the inertial frame, and the
plays form two families, a radial race against one pursuer and a
simultaneous exit through a corner, with states and costates in closed
form (Sections~\ref{sec:hamiltonian} and~\ref{sec:families}). At a
corner the incoming heading may point back into one disk; a single
scalar test, \(\Gamma\ge0\), decides whether a turn to the bisector at
contact still completes the escape (Proposition~\ref{prop:completion}
and Figure~\ref{fig:corner}).
\item \emph{A complete partition of the state space.} A finite algorithm
computes \(U\) exactly, and explicit surfaces partition the state space
into regions of radial, corner, and turning escapes. They include a
fold, at which \(D(T)-\mu T\) touches zero before the escape time, its
cusp curve, and states at which the escape time jumps
(Section~\ref{sec:partition}).
\item \emph{Verification under both information conventions.} For
half-separations \(a\le R\sqrt{1-\mu^{-2}}\), the escape time is the
value, with an explicit optimal evader policy and an explicit
nonanticipative pursuer strategy (Theorem~\ref{thm:value}), and the
explicit feedback extends to part of the larger separations
(Theorem~\ref{thm:extension}). For every initial state, a time-budget
viability argument proves \(U=V_P\), the value when the pursuers react
to the evader (Theorem~\ref{thm:globalvalue}), and an approximate
viability theorem with a short opening phase proves \(U=V_E\), the value
when the evader reacts to the pursuers (Theorem~\ref{thm:reverse}).
Beyond Theorems~\ref{thm:value} and~\ref{thm:extension}, the pursuer
strategy is obtained from an existence theorem, and in some corner
regions the construction supplies \(\varepsilon\)-optimal evader policies.
\end{enumerate}
Table~\ref{tab:results} summarizes the four value results and how
explicit each one is.

\begin{table}[htbp]\centering\small
\begin{tabular}{@{}>{\raggedright\arraybackslash}p{.17\linewidth}>{\raggedright\arraybackslash}p{.27\linewidth}>{\raggedright\arraybackslash}p{.48\linewidth}@{}}
\toprule
Result & Domain & Content\\
\midrule
Theorem~\ref{thm:value}
 & \(0<a_0\le a_{\rm c}=R\sqrt{1-\mu^{-2}}\)
 & The value is the unique root of \(D(T)=\mu T\); explicit pursuer
   feedback and an explicit evader policy, possibly with a terminal
   turn.\\
Theorem~\ref{thm:globalvalue}
 & Every state
 & \(V_P=U\), including folds, cusps, and limiting cells; a
   nonanticipative pursuer strategy exists; evader policies are exact or
   \(\varepsilon\)-optimal depending on the cell.\\
Theorem~\ref{thm:reverse}
 & Every state
 & \(V_E=V_P=U\): for every nonanticipative evader strategy and every
   \(T<U\), some pursuer control keeps detection until \(T\).\\
Theorem~\ref{thm:extension} (Appendix~\ref{app:extension})
 & \(a_0>a_{\rm c}\), \(|x_0|>S(a_0)\)
 & The explicit feedback of Theorem~\ref{thm:value} still realizes the
   value on an invariant region beyond \(a_{\rm c}\).\\
\bottomrule
\end{tabular}
\caption{Value results of this paper. Here \(a_0\) and \(x_0\) are the
initial half-separation and the initial evader coordinate along the
pursuer baseline; \(D(T)\) is the evader's distance to the region outside
both expanded disks, eq.~\eqref{eq:distance}; \(U\) is the escape time
\eqref{eq:strictroot}; \(V_P\) is the value \eqref{eq:strategyvalue};
\(S\) is the threshold \eqref{eq:extensionS}; and \(V_E\) is the value
under the reverse ordering.}
\label{tab:results}
\end{table}

\paragraph{How the proofs fit together}
The upper bound \(V_P\le U\) is elementary
(Proposition~\ref{prop:bounds}) and holds at every state. The main
remaining difficulty is the matching lower bound \(V_P\ge U\), a
guarantee for the pursuers against every evader control.
\begin{itemize}
\item Sections~\ref{sec:distance}--\ref{sec:families} give the distance
in closed form, the regular plays, and the exact evader policies. They
settle the case in which one pursuer alone is decisive
(Proposition~\ref{prop:single}), but do not yet give a cooperative
pursuer guarantee.
\item Theorem~\ref{thm:value} proves the lower bound for
\(a_0\le a_{\rm c}\) with an explicit feedback. This feedback is
discontinuous on the symmetry line, and Lemma~\ref{lem:realization}
shows that it still defines a causal motion.
Theorem~\ref{thm:extension} keeps it on a larger invariant region.
\item Theorem~\ref{thm:globalvalue} proves the lower bound at every
state. The escape time jumps at folds, so the pursuers preserve instead
a time budget at all horizons (Lemma~\ref{lem:auxviability}, whose cases
follow the classification of Section~\ref{sec:partition}).
Lemma~\ref{lem:saturation} then restores unit speed.
\item Theorem~\ref{thm:reverse} treats the reverse ordering. The
viability argument gives the bound only up to the first contact, which
at a fold comes before the escape time; Lemma~\ref{lem:foldopening}
closes this gap.
\end{itemize}

\paragraph{Organization}
Section~\ref{sec:formulation} states the game, the strategy convention,
and the attainable-disk bound on which everything else rests.
Section~\ref{sec:distance} computes the distance \(D\) in closed form, so
that the escape time can be evaluated from the start.
Section~\ref{sec:hamiltonian} shows that, on regular optimal arcs, the
active players move along straight lines, and gives the Hamiltonian
structure in reduced coordinates.
Section~\ref{sec:families} constructs the two families of optimal plays,
the terminal-completion test, and the matching of an initial state, and
Section~\ref{sec:regime} proves the value theorem for moderate
separations. Section~\ref{sec:partition} gives a finite algorithm for
the escape time and partitions the full state space, and
Section~\ref{sec:global} proves the global value theorem and treats the
reverse ordering. Section~\ref{sec:examples} presents examples and
numerical validation, and Section~\ref{sec:conclusion} concludes with
open problems. Appendix~\ref{app:extension} extends the explicit feedback
beyond \(a_{\rm c}\); Appendices~\ref{app:realization},
\ref{app:auxviability}, and~\ref{app:reverse} contain the most technical
proofs. The configuration of Figure~\ref{fig:geometry} is followed as a
running example through Sections~\ref{sec:distance}--\ref{sec:families}.
A reader interested mainly in the results can read
Sections~\ref{sec:formulation}--\ref{sec:distance}, \ref{sec:families},
and~\ref{sec:partition}, the statements of
Theorems~\ref{thm:value}--\ref{thm:reverse}, and
Section~\ref{sec:examples}.

\section{Problem formulation}
\label{sec:formulation}

\subsection{Dynamics and payoff}
Let \(P_1,P_2\in\R^2\) be the pursuer positions and \(E\in\R^2\) the
evader position, and let \(v_P\) and \(v_E\) be the physical speeds of
each pursuer and of the evader. With normalized time \(t=v_Pt_{\rm phys}\),
the dynamics are
\begin{equation}\label{eq:physical}
 \dot P_i=u_i,\quad \norm{u_i}=1,\qquad
 \dot E=\mu e,\quad \norm e=1,\qquad \mu=v_E/v_P>1.
\end{equation}
The normalized time \(t\) is the distance a pursuer can travel, so it
has units of length unless a length scale is also fixed; physical times
are recovered by dividing by \(v_P\). We write
\(P_i^0\) and \(E^0\) for the initial positions.

The game is invariant under translations and rotations of the plane, so
it depends on the positions only through three numbers: half the
distance \(a\) between the pursuers, and the coordinates \((x,y)\) of the
evader relative to the pursuer midpoint, with the first axis pointing
from \(P_1\) to \(P_2\) (any direction will do if the pursuers
coincide). Throughout the paper the \emph{state} is this
triple,
\[
 z=(a,x,y),
\]
and Section~\ref{sec:hamiltonian} gives its dynamics. Positions in the
plane are always written \(P_i\) and \(E\); the letter \(z\) is never used
for them. The players have
complete state information, and there are no obstacles and no bound on
turning rate. The Hamiltonian is the same for the closed unit-ball
control sets, and all regular controls constructed below are saturated.

Each pursuer senses the evader within distance \(R>0\). The sensing
region, called the \emph{sensing union} below, is the union of the two
disks: the evader remains detected
while at least one distance is at most \(R\). The payoff is the escape
time
\begin{equation}\label{eq:exit}
 t_f=\inf\{t\ge0:\norm{E(t)-P_1(t)}>R,\ \norm{E(t)-P_2(t)}>R\},
\end{equation}
which the evader minimizes and the pursuer team maximizes. Leaving the
overlap of the two disks does not end the game. At a transversal escape,
the infimum in \eqref{eq:exit} equals the time of boundary contact, and
the strict inequalities hold immediately afterward.

\subsection{Strategies and value}
Let \(\mathcal E\) and \(\mathcal U\) denote the measurable evader controls
and pursuer-team controls. A pursuer strategy
\(\mathcal B:\mathcal E\to\mathcal U\) is \emph{nonanticipative} if evader
controls that agree almost everywhere on \([0,t]\) produce pursuer
controls that agree almost everywhere on \([0,t]\). Such a strategy may
respond to the evader's current heading but not to its future headings.
This is an idealization: the pursuers know the state exactly and observe
the evader's heading without delay.
Controls and strategy outputs are identified when they agree almost
everywhere.
Writing \(J_z(e,u)\) for the escape time from state \(z\), we study
\begin{equation}\label{eq:strategyvalue}
 V_P(z)=\sup_{\mathcal B\ {\rm nonanticipative}}\
 \inf_{e\in\mathcal E}J_z(e,\mathcal B[e]),
\end{equation}
and refer to it as the value. This ordering models pursuers that observe
the evader through their sensors and respond continuously, and several
of the strategies below use the evader's current heading explicitly.
The reverse ordering,
\[
 V_E(z)=\inf_{\mathcal A\ {\rm nonanticipative}}\
 \sup_{u\in\mathcal U}J_z(\mathcal A[u],u),
 \qquad \mathcal A:\mathcal U\to\mathcal E,
\]
gives the nonanticipative strategy to the evader. The Hamiltonian \(H\) of the game, derived in
Section~\ref{sec:hamiltonian} (eq.~\eqref{eq:H}), satisfies the
pointwise Isaacs identity \(\min_e\max_u H=\max_u\min_e H\). Equality of
\(V_E\) and \(V_P\) would nevertheless require a value theorem whose
regularity hypotheses on the terminal set hold for the open escape
target in \eqref{eq:exit}. For instance, \cite[Theorem~8.1]{elliottkalton}
extends the equality of upper and lower values from games of prescribed
duration to games of survival whose terminal boundary is regular, under
the Isaacs condition and a Lipschitz condition. We have not verified
that regularity for the open escape target, and we do not use such a
theorem. Instead, Theorem~\ref{thm:reverse} proves \(V_E=V_P\) at every
state directly. The two orderings thus use different information
conventions but have the same value in this game; the equality does not
follow merely from interchanging the infimum and the supremum. Since the game is zero-sum, a Nash equilibrium is a saddle
point; Remark~\ref{rem:saddle} states what the results give in those
terms.

\subsection{Attainable sensing disks}
\label{sub:attainable}
\begin{figure}[H]\centering
\includegraphics[width=.9\linewidth]{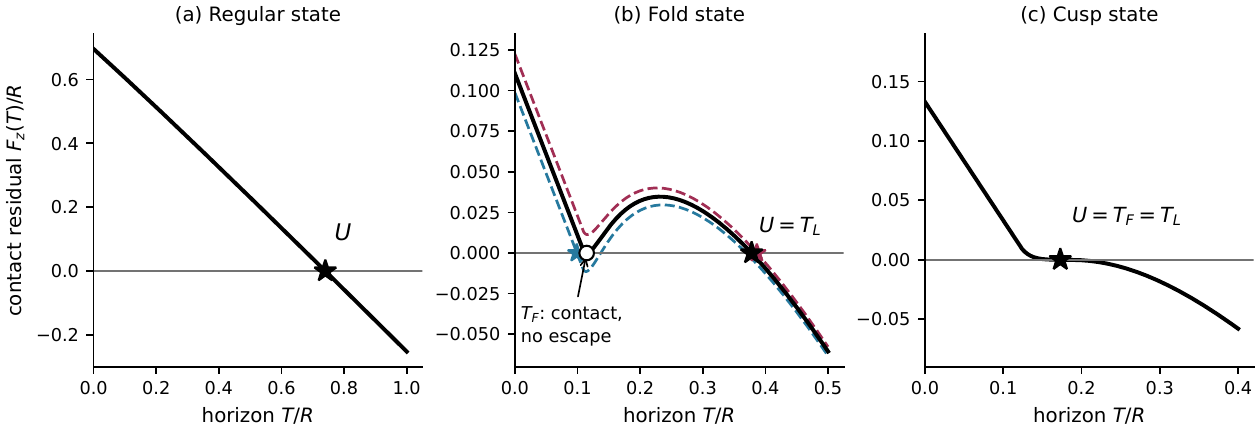}
\caption{The three shapes of the contact residual \(F_z(T)=D_z(T)-\mu T\),
for \(R=1\), \(\mu=2\). (a) A regular state: \(F_z\) crosses zero once, and
the crossing is the escape time \(U\). (b) A fold state (solid): \(F_z\)
touches zero at \(T_F\) and rises again, so reaching the exterior of the
expanded disks at time \(T_F\) gives no escape, and \(U\) is the later
crossing \(T_L\). The dashed curves are two nearby states, one on each
side of the fold: for one the escape time is close to \(T_F\), for the
other close to \(T_L\) (coloured stars), so the escape time jumps across
the fold. (c) A cusp state: the tangency and the crossing merge.}
\label{fig:residual}
\end{figure}

The analysis rests on one geometric observation. After time \(T\),
pursuer \(i\) lies in the disk of radius \(T\) about its initial position
\(P_i^0\). The set it can possibly sense at time \(T\) is therefore
contained in the \emph{expanded disk} of radius \(R+T\) about \(P_i^0\).
If the evader can reach a point outside both expanded disks within time
\(T\), it has escaped by time \(T\) whatever the pursuers do. To make
this precise, let
\begin{equation}\label{eq:distance}
 C_T=\{Q:\norm{Q-P_1^0}\ge R+T,\ \norm{Q-P_2^0}\ge R+T\},
 \qquad D_z(T)=\min_{Q\in C_T}\norm{Q-E^0},
\end{equation}
be the exterior of the expanded disks and the evader's distance to it.
When the state is clear from the context we write \(D(T)\) for
\(D_z(T)\). Define
\begin{equation}\label{eq:strictroot}
 F_z(T)=D_z(T)-\mu T,\qquad
 U(z)=\inf\{T\ge0:F_z(T)<0\}
\end{equation}
for states in the closed sensing union, and \(U(z)=0\) for states
strictly outside both disks. We call \(F_z\) the \emph{contact residual}
and \(U\) the \emph{escape time}. Figure~\ref{fig:geometry} in the
Introduction illustrates the construction, and
Figure~\ref{fig:residual} shows the shapes that \(F_z\) can take. The
strict inequality in \eqref{eq:strictroot} matters only when \(F_z\)
touches zero without becoming negative, as in
Figure~\ref{fig:residual}(b); Section~\ref{sub:fold} shows that this
does occur, and gives an example in which the evader reaches the
closing gap between the disks at the instant it closes.

Write \(d=\min_i\norm{E^0-P_i^0}\) for the distance to the nearer pursuer
and \(B=R/(\mu-1)\). A single pursuer that heads straight at the evader
keeps it detected for the time \((R-d)/(\mu-1)\), because their distance
grows at rate at most \(\mu-1\); \(B\) is this time for an evader that
starts at the pursuer's position. The next proposition places the value
between the single-pursuer time and \(U\), and shows that \(U\) never
exceeds \(B\).

\begin{proposition}[Elementary bounds]\label{prop:bounds}
For every state in the sensing union,
\begin{equation}\label{eq:globalbounds}
 \frac{(R-d)_+}{\mu-1}\ \le\ V_P(z)\ \le\ U(z)\ \le\ B.
\end{equation}
For every \(\varepsilon>0\) the evader has a constant heading that escapes
by \(U(z)+\varepsilon\) against all pursuer controls.
\end{proposition}
\begin{proof}
\emph{Lower bound.} Let the nearer pursuer head straight at the evader
at all times (pure pursuit).
Their distance then grows at rate at most \(\mu-1\), so it stays at most
\(R\) until time \((R-d)/(\mu-1)\).

\emph{The bound \(V_P\le U\).} Let \(F_z(t)<0\). By the definition
\eqref{eq:distance} of \(D_z(t)\), the point \(Q\) of \(C_t\) closest to
\(E^0\) satisfies \(\norm{Q-E^0}<\mu t\). The evader runs straight to \(Q\)
and arrives at a time \(t'<t\). Whatever the pursuers do, each has moved
at most \(t'\) by then, so \(\norm{Q-P_i(t')}\ge(R+t)-t'>R\) for both
\(i\), and \(t_f\le t'<t\). Such times \(t\) exist arbitrarily close to
\(U(z)\), which proves \(V_P\le U\) and the last claim.

\emph{The bound \(U\le B\).} Two closed half-planes through the origin
always share a direction, so there is a unit vector \(e\) with
\((E^0-P_i^0)\cdot e\ge0\) for both \(i\). For every \(T\ge0\), the point
\(E^0+(R+T)e\) is at distance at least \(R+T\) from both \(P_i^0\), so it
lies in \(C_T\) and \(D_z(T)\le R+T\). Hence \(F_z(T)\le R-(\mu-1)T\),
which is negative for \(T>B\).
\end{proof}
In the middle step, \(t\) is a test time, not the arrival time: the
evader reaches \(Q\) at \(t'=D_z(t)/\mu<t\), and the margin
\(t-t'=-F_z(t)/\mu\) is what keeps \(Q\) outside both sensing disks at
that instant. The upper bound \(V_P\le U\) is now settled at every state.
What remains, and what most of the paper is about, is the lower bound
\(V_P\ge U\): a pursuer strategy that keeps the evader detected until
\(U\) against every evader control. The pure-pursuit bound in
\eqref{eq:globalbounds} gives it only where one pursuer suffices.

\begin{remark}[Information used by the explicit strategies]\label{rem:information}
The two players use very different amounts of information.
\begin{enumerate}
\item \emph{The evader needs no online observations.} Its optimal and \(\varepsilon\)-optimal
policies (Propositions~\ref{prop:bounds}, \ref{prop:single},
and~\ref{prop:completion}) are open-loop: a constant heading, followed
at most by one turn at a prescribed time. They work against every
pursuer motion, so the upper bound \(V_P\le U\) needs no information
about the pursuers. The same policies, viewed as constant evader
strategies, also give \(V_E\le U\).
\item \emph{The explicit pursuer strategy needs little.} The feedback of
Theorem~\ref{thm:value} and Theorem~\ref{thm:extension} depends on the
current state, the evader's current heading, and one bit of memory
recording whether the handoff to pure pursuit, in which a pursuer heads
straight at the evader, has occurred. Off the
symmetry line \(y=0\) it is a pure state feedback: each pursuer heads
straight for the current escape point. The
current heading enters only on \(y=0\), through the dispersal response
\eqref{eq:tie}; the local fold response \eqref{eq:foldguard} of
Section~\ref{sub:foldguard} uses it as well. On
\(y=0\) this is unavoidable in continuous time: no fixed choice of
pursuer headings there keeps the remaining escape time from decreasing
faster than time elapses against all evader deviations
(Section~\ref{sec:regime}).
\end{enumerate}
Remark~\ref{rem:saddle}, after the general theorems, says what
information their strategies use and what the results mean in terms of
saddle points.
\end{remark}

The rest of the paper computes \(U\) in closed form, describes the plays
that achieve it, and proves the reverse inequality \(V_P\ge U\).
Table~\ref{tab:notation} collects recurring notation. Symbols that are
not listed there are local to the subsection or proof in which they are
introduced. A few letters, such as \(\tau\), \(\rho\), \(\delta\), \(h\),
and \(s\), are reused locally, with the meaning stated at each use.

\begin{table}[!htb]\centering\footnotesize
\begin{tabular}{@{}>{\raggedright\arraybackslash}p{.24\linewidth}>{\raggedright\arraybackslash}p{.68\linewidth}@{}}
\toprule
Symbol & Meaning\\
\midrule
\(R,\ \mu\) & Sensing radius; evader-to-pursuer speed ratio, \(\mu>1\).\\
\(z=(a,x,y)\) & State: half the pursuer separation, and the evader
 position relative to the pursuer midpoint, with the \(x\)-axis along the
 baseline, Section~\ref{sec:formulation}.\\
\(P_i^0,\ E^0\) & Initial positions in the plane.\\
\(D_z,\ F_z,\ U\) & Exterior distance, contact residual
 \(F_z=D_z-\mu T\), and escape time,
 eqs.~\eqref{eq:distance}--\eqref{eq:strictroot}.\\
\(V_P,\ V_E\) & Value under pursuer- and evader-nonanticipative
 strategies, Section~\ref{sec:formulation}.\\
\(U_0\) & Earliest zero of the contact residual \(F_z\),
 eq.~\eqref{eq:earliestzero}.\\
\(B=R/(\mu-1)\) & Universal upper bound on the escape time.\\
\(X,\ Y,\ d,\ m\) & \(|x|\), \(|y|\), distance to the nearer pursuer, and
 \(\mu^2-1\), Section~\ref{sub:distance}.\\
\(r,\ h\) & Expanded radius \(R+T\) and corner height \(\sqrt{r^2-a^2}\)
 at the horizon \(T\) in question.\\
\(r_m,\ t_m\) & Expanded radius and horizon at which the radial and corner
 pieces of \(D_z\) join, eq.~\eqref{eq:rm}.\\
\(\beta,\ c,\ L\) & Corner weight, \(c=\norm{\beta n_1+(1-\beta)n_2}\),
 and \(L=1/(\mu c-1)\), eqs.~\eqref{eq:cornerdefs}
 and~\eqref{eq:cornercontrols}.\\
\(q_i,\ \Gamma\) & Outward rate of the incoming heading at constraint
 \(i\); outward rate after a turn to the bisector,
 eqs.~\eqref{eq:rates} and~\eqref{eq:completioncriterion}.\\
\(T_F,\ T_L\) & At a fold state: the time at which \(F_z\) touches zero,
 and its later crossing, Section~\ref{sub:fold}.\\
\(\kappa=\sqrt{1-\mu^{-2}}\) & So that \(a_{\rm c}=\kappa R\),
 eq.~\eqref{eq:safe}.\\
\(a_{\rm tr}<a_{\rm c}<a_*<a_{\rm one}\) & Separation thresholds
 \(R\sqrt{(\mu-1)/(2\mu)}\), \(\kappa R\), \(\mu^2R/(\mu^2-1)\),
 \(\mu R/(\mu-1)\), eqs.~\eqref{eq:transversalitythreshold},
 \eqref{eq:safe}, \eqref{eq:foldrange}, and Section~\ref{sub:regimes};
 at \(\mu=2\) they are \(0.5R\), \(0.866R\), \(1.333R\), \(2R\).\\
\bottomrule
\end{tabular}
\caption{Recurring notation and where each symbol is defined.}
\label{tab:notation}
\end{table}

\FloatBarrier
\section{The exact distance function}
\label{sec:distance}
The escape time \eqref{eq:strictroot} is defined through the distance
\(D_z(T)\) from the evader to the exterior \(C_T\) of the expanded disks.
This section describes the closest point of \(C_T\), gives \(D_z(T)\) in
closed form, and shows that its two pieces, one for each family of
Section~\ref{sec:families}, join smoothly. Everything later in the paper
is computed from this formula.

\subsection{Closest points and ties}
\label{sub:ties}
The nearest point of \(C_T\) is either a radial point on one expanded
circle that is \emph{exposed}, meaning that it does not lie inside the
other expanded disk, or one of the two intersections of the expanded
circles, the \emph{corners}. Section~\ref{sec:families} therefore
enumerates all regular candidates.
\begin{lemma}[Closest-point ties and continuity]\label{lem:ties}
Let \(a_0>0\), \(r=R+T>a_0\), and let the evader lie strictly inside the
expanded union. The only tie between distinct, nondegenerate closest
points is between the upper and lower corners, and it occurs at
\(y_0=0\). An exposed point can merge with a corner at a change
of active set, but cannot tie with a distinct one. The distance \(D(T)\)
is continuous, including at such transitions.
\end{lemma}
\begin{proof}
Put \(d_i=\norm{E^0-P_i^0}\). An exposed minimum on circle \(i\) has
\(0<d_i<r\), radial point \(Q_i=P_i^0+r(E^0-P_i^0)/d_i\), and distance
\(r-d_i\). Two such distances tie only if \(d_1=d_2=d\), hence
\(x_0=0\); but then \(\norm{Q_i-P_j^0}^2-r^2=4a_0^2(1-r/d)<0\), so
neither radial point is exposed. For any \(Q\) on circle \(i\), with
\(\alpha\) the angle between \(Q-P_i^0\) and \(E^0-P_i^0\),
\[
 \norm{Q-E^0}^2-(r-d_i)^2=2rd_i(1-\cos\alpha),
\]
which is positive unless \(Q=Q_i\), so an exposed point cannot tie with
a distinct corner. For the corners \(Q_\pm=(0,\pm h)\), the
squared-distance difference is \(-4hy_0\).

The radial and corner locations depend continuously on \(r\). A radial
candidate enters or leaves its exposed arc by meeting a corner, where
the two distance expressions agree, so the minimum of these matched
pieces is continuous. In the degenerate case \(d_i=0\) the evader is at
pursuer \(i\)'s position, every point of circle \(i\) is at distance \(r\),
and \(D(T)=r\) remains continuous.
\end{proof}

\subsection{The closed form}
\label{sub:distance}
By reflection and interchange of the pursuers, put
\[
 X=|x|,\quad Y=|y|,\quad d=\sqrt{(a-X)^2+Y^2},\quad
 m=\mu^2-1,\quad r=R+T,
\]
so \(d\) is the distance to the nearer pursuer. Assume \(a>0\) and
\(0<d<R\). For \(X<a\) define
\begin{equation}\label{eq:rm}
 r_m=\frac{ad}{a-X},\qquad t_m=r_m-R,
\end{equation}
and set \(r_m=+\infty\) for \(X\ge a\). Here \(r_m\) is the expanded
radius at which the radial point of the nearer pursuer's circle reaches
the other expanded circle, and \(t_m\) is the corresponding horizon:
up to \(t_m\) the closest exterior point is radial, and after it the
closest point is a corner. Then
\begin{equation}\label{eq:Dexplicit}
 D_z(T)=
 \begin{cases}
 r-d,&r\le r_m,\\[2pt]
 \sqrt{X^2+\big(\sqrt{r^2-a^2}-Y\big)^2},&r>r_m.
 \end{cases}
\end{equation}
In the second case write \(h=\sqrt{r^2-a^2}\); it is real because
\(r_m\ge a\). The first line is the single-active family of
Section~\ref{sub:single}, and the second is the corner family of
Section~\ref{sub:corner}.

To see this, take \(x\ge0\), so the nearer pursuer is \(P_2=(a,0)\), with
outward radial point \(Q_2=P_2+r(E-P_2)/d\) on its expanded circle.
Direct calculation gives
\[
 \norm{Q_2-P_1}^2-r^2=4a\left[a-\frac{r(a-X)}d\right],
\]
so \(Q_2\) is exposed exactly when \(r\le r_m\). Otherwise the closest
exterior point is the intersection on the side of \(y\), or either
intersection if \(y=0\). The corner is selected exactly when
\begin{equation}\label{eq:cornerselection}
 X<a,\qquad h(a-X)\ge aY,
\end{equation}
which is the positive-normal-cone condition in state coordinates.

The two pieces of \eqref{eq:Dexplicit} join smoothly. This is used in
Sections~\ref{sec:regime} and~\ref{sec:global} whenever a play or a
constraint passes from the single-active to the corner family. We call
the horizon \(t_m\) the \emph{join time}.

\begin{lemma}[Smooth merger]\label{lem:merger}
Let \(0<X<a\) and \(Y>0\). On the merger \(r=r_m\), the exposed
expression \(r-d\) and the corner expression
\(D_c=\sqrt{X^2+(h-Y)^2}\) in \eqref{eq:Dexplicit} agree in value and in
their gradients with respect to \((a,x,y,T)\). Consequently \(D_z(T)\) is
continuously differentiable in \((z,T)\) across the merger, and so is
every root of \(D_z(T)=\mu T\), as a function of \(z\), at which
\(\partial_T(D-\mu T)\ne0\).
\end{lemma}
\begin{proof}
By reflection take \(x=X\) and \(y=Y\). At \(r=r_m=ad/(a-x)\),
\[
 h=\sqrt{r^2-a^2}=\frac{ay}{a-x}=\frac{ry}{d},\qquad
 h-y=\frac{xy}{a-x},\qquad r-d=\frac{xd}{a-x}.
\]
The radial point \(Q_2=P_2+r(E-P_2)/d\) has coordinates
\(\big(a-r(a-x)/d,\ ry/d\big)=(0,h)\), so it is the corner, and both
expressions equal \(r-d\). Since \(\partial_Th=r/h\) and
\(\partial_ah=-a/h\), the corner expression has
\[
 \partial_TD_c=\frac{r(h-y)}{hD_c},\quad
 \partial_aD_c=-\frac{a(h-y)}{hD_c},\quad
 \partial_xD_c=\frac{x}{D_c},\quad
 \partial_yD_c=-\frac{h-y}{D_c}.
\]
Substituting the values above, and using \(a/r=(a-x)/d\), gives
\[
 \nabla_{(a,x,y,T)}D_c=\Big(-\frac{a-x}{d},\ \frac{a-x}{d},\
 -\frac yd,\ 1\Big),
\]
which is exactly the gradient of \(r-d\). Both expressions are \(C^1\)
near the merger, where \(d>0\), \(h>0\), and \(D_c>0\), and the merger
is the \(C^1\) hypersurface \(r=r_m(z)\). A function that equals one
\(C^1\) expression on each side of such a hypersurface, with matching
values and gradients on it, is \(C^1\). The statement about the root
follows from the implicit function theorem.
\end{proof}
At the axial endpoint \(Y=0\), \(r_m=a\), the corner gradient is singular
because \(h=0\); this is the axial kink treated separately in
Section~\ref{sub:fold}.

\paragraph{Running example}
The configuration of Figure~\ref{fig:geometry} has \(R=10\), \(\mu=2\),
and \(z=(5,2,2)\). The nearer pursuer is \(P_2\), at distance
\(d=\sqrt{13}\approx3.606\), and \eqref{eq:rm} gives
\(r_m=5\sqrt{13}/3\approx6.009<R\), so \(t_m<0\). The radial point of
\(P_2\)'s circle is covered by the other disk from the start, and
\(D_z\) is the corner piece for every \(T\ge0\):
\[
 D_z(T)=\sqrt{4+\Big(\sqrt{(10+T)^2-25}-2\Big)^2},\qquad
 D_z(0)=\sqrt{4+(5\sqrt3-2)^2}\approx6.954.
\]
The residual \(F_z(T)=D_z(T)-2T\) crosses zero once, at
\[
 U\approx7.40194,
\]
as in Figure~\ref{fig:geometry}(b); Section~\ref{sec:regime} shows that
the crossing is unique for all states with \(a\le\kappa R\), and here
\(a=5<5\sqrt3\). The escape corner is \(Q=(0,h)\) with \(r=R+U\approx17.402\)
and \(h=\sqrt{r^2-a^2}\approx16.668\). Section~\ref{sub:matching}
continues the example.

\section{Hamiltonian structure and straight-line characteristics}
\label{sec:hamiltonian}

\subsection{Straight lines in the inertial frame}
\label{sub:straight}
All players have simple motion. Let \(\Pi_1,\Pi_2,\pi_E\) be the inertial costates. The
inertial Hamiltonian
\[
 H=1+\Pi_1\cdot u_1+\Pi_2\cdot u_2+\mu\pi_E\cdot e
\]
does not depend on positions, so all costates are constant. Whenever
they are nonzero,
\begin{equation}\label{eq:straight}
 u_i=\Pi_i/\norm{\Pi_i},\quad e=-\pi_E/\norm{\pi_E},\qquad
 P_i(t)=P_i^0+t u_i,\quad E(t)=E^0+\mu t e.
\end{equation}
So on each regular characteristic arc the evader and every active
pursuer, one with nonzero costate, have constant inertial headings. An
inactive pursuer, with zero costate, does not affect the escape time to
first order and may use any admissible motion; a constant heading is a
convenient representative. Terminal completion may add an evader turn at
contact (Section~\ref{sub:completion}). Finding a regular optimal play
therefore amounts to finding its terminal configuration and its constant
headings.
Section~\ref{sec:families} does this for the two kinds of escape. The
rest of this section describes the same plays in the reduced
coordinates \((a,x,y)\), in which the families are parameterized and the
value is verified. There the dynamics are nonlinear, because the frame
attached to the pursuers rotates.

\subsection{Reduced coordinates}
The game is invariant under translations and rotations of the plane, so
three numbers describe the relative configuration: the half-separation
\(a\) of the pursuers and the evader's coordinates \((x,y)\) in a frame
attached to the pursuer baseline. Attach this frame to the pursuer
midpoint, with its horizontal axis pointing from \(P_1\) to \(P_2\), so
that
\[
 P_1=(-a,0),\qquad P_2=(a,0),\qquad E=(x,y),\qquad a>0.
\]
This is the state \(z=(a,x,y)\) of Section~\ref{sec:formulation}; we also
call it the reduced state when the contrast with positions in the plane
matters.

Two frames are used, and in both of them angles are measured
counterclockwise. The \emph{inertial frame} is fixed. A heading is a unit
vector, or equivalently its angle measured counterclockwise from the
first coordinate axis; the controls \(u_i\) and \(e\) in
\eqref{eq:physical} are inertial headings. Unless stated otherwise, the
inertial frame is chosen to coincide with the moving frame at the
initial time, so that \(P_1^0=(-a,0)\), \(P_2^0=(a,0)\), and
\(E^0=(x,y)\) for the initial state. The \emph{moving frame} has its first
axis \(b_1\) pointing from \(P_1\) to \(P_2\) and its second axis \(b_2\)
equal to \(b_1\) rotated counterclockwise by a right angle. If
\(\vartheta\) is the angle of \(b_1\) in the inertial frame, a heading with
inertial angle \(\phi\) has the angle \(\phi-\vartheta\) in the moving
frame, measured counterclockwise from \(b_1\). Its components there are
its \emph{reduced heading}:
\[
 (c_i,s_i)=(u_i\cdot b_1,\ u_i\cdot b_2),\qquad
 (e_x,e_y)=(e\cdot b_1,\ e\cdot b_2).
\]
We use the same letters \(u_i\) and \(e\) for a heading and for its
reduced components when the frame is clear. The moving frame rotates at
the rate \(\omega=\dot\vartheta=(s_2-s_1)/(2a)\), positive
counterclockwise; it does not rotate when the two pursuers have the same
velocity component normal to the baseline. In these terms,
\begin{equation}\label{eq:dynamics}
 \dot a=\frac{c_2-c_1}{2},\quad
 \dot x=\mu e_x-\frac{c_1+c_2}{2}+\omega y,\quad
 \dot y=\mu e_y-\frac{s_1+s_2}{2}-\omega x.
\end{equation}
The first equation says that the half-separation changes at half the
pursuers' relative velocity along the baseline. In the other two, the evader's
velocity is taken relative to the pursuer midpoint, and the terms in
\(\omega\) account for the rotation of the frame.
Let \(p=(p_a,p_x,p_y)\) be the costate conjugate to \(z\). On a smooth
local branch of the candidate value, introduced after \eqref{eq:normal},
the normalized costate is the gradient of that branch with respect to
\(z\). After verification it is the gradient of the value wherever the
value is differentiable. It is not a classical gradient where the value
has a kink or a jump, as on the line \(y=0\) and across the folds of
Section~\ref{sub:fold}. Put
\begin{align}
 K&=\frac{yp_x-xp_y}{a},\nonumber\\
 A_1&=\frac12(-p_a-p_x,-p_y-K),\qquad
 A_2=\frac12(p_a-p_x,-p_y+K).\label{eq:A}
\end{align}
The control Hamiltonian and its saddle value are
\begin{align}
 H&=1+\mu(p_x,p_y)\cdot e+A_1\cdot u_1+A_2\cdot u_2,\nonumber\\
 \bar H(z,p)&=1+\norm{A_1}+\norm{A_2}-\mu\sqrt{p_x^2+p_y^2},
 \label{eq:H}
\end{align}
so the regular feedback controls are
\begin{equation}\label{eq:controls}
 e^*=-\frac{(p_x,p_y)}{\sqrt{p_x^2+p_y^2}},\qquad
 u_i^*=\frac{A_i}{\norm{A_i}}.
\end{equation}
If \(A_i=0\), the control of pursuer \(i\) does not enter the local
Hamiltonian and is not determined by it. We call pursuer \(i\)
\emph{active} on an arc where \(A_i\ne0\) and \emph{inactive} where
\(A_i=0\); a pursuer whose range at the escape time exceeds \(R\) is
inactive. An arc is \emph{regular} if \((p_x,p_y)\ne0\) and each pursuer is
either active throughout or inactive throughout, so that
\eqref{eq:controls} determines the controls of the evader and of the
active pursuers.

Each player steers along its own costate. A pursuer moves along \(A_i\),
which increases the remaining time fastest, and the evader moves against
\((p_x,p_y)\). Section~\ref{sub:reconstruction} shows that \(A_i\) is pursuer
\(i\)'s inertial costate expressed in the moving frame. The coupling
term \(K\) and the frame rate \(\omega\) appear only because the frame
rotates when the pursuers have different velocity components normal to
their baseline. When those components are equal, \(\omega=0\) and the
reduced equations are linear in the headings; this is the situation in
the two-active family of Section~\ref{sub:corner}.

The forward costate equations are
\begin{equation}\label{eq:adjoint}
 \dot p_a=\frac{\omega}{a}(yp_x-xp_y),\qquad
 \dot p_x=\omega p_y,\qquad \dot p_y=-\omega p_x.
\end{equation}
Equations \eqref{eq:dynamics}, \eqref{eq:controls}, and \eqref{eq:adjoint}
are the characteristic equations; in retro-time \(\tau=t_f-t\) every
right-hand side changes sign. Along a normal characteristic with free
terminal time,
\begin{equation}\label{eq:normal}
 \bar H=0,\qquad \frac{\dd V}{\dd t}=-1,
\end{equation}
where \(V\) denotes a smooth local branch of the candidate value, the
remaining escape time along the characteristic. Before verification it
is only a candidate; Theorems~\ref{thm:value} and~\ref{thm:globalvalue}
identify it with \(V_P\) wherever it is smooth.

\subsection{From the inertial to the reduced description}
\label{sub:reconstruction}
Given the constant headings of Section~\ref{sub:straight}, the reduced
trajectory follows without integrating \eqref{eq:dynamics}:
\begin{align}
 \Delta P(t)&=P_2^0-P_1^0+t(u_2-u_1),& a(t)&=\tfrac12\norm{\Delta P(t)},\nonumber\\
 b_1(t)&=\Delta P(t)/\norm{\Delta P(t)},& b_2(t)&=(-b_{1y},b_{1x}),\nonumber\\
 M(t)&=\tfrac12(P_1(t)+P_2(t)),&
 x(t)&=(E(t)-M(t))\cdot b_1(t),\nonumber\\
 &&y(t)&=(E(t)-M(t))\cdot b_2(t).\label{eq:reconstruction}
\end{align}

The game is invariant under translations and rotations of the whole
configuration. For the inertial costates of a characteristic to
correspond to a costate \(p\) on the reduced state, they must have zero
total linear momentum and zero total angular momentum:
\begin{equation}\label{eq:zeromomenta}
 \Pi_1+\Pi_2+\pi_E=0,\qquad
 \mathcal J=P_1\mathbin\times\Pi_1+P_2\mathbin\times\Pi_2
              +E\mathbin\times\pi_E=0,
\end{equation}
where \(v\mathbin\times w=v_xw_y-v_yw_x\). On a smooth branch these
conditions say that the gradient of an invariant function vanishes along
the symmetry directions; the argument that follows does not use
differentiability. Both conditions hold for the
terminal families of Section~\ref{sec:families}. Let \(P_{if}\) and
\(E_f\) be the positions at the escape time \(t_f\). Writing
\(\Pi_i=\lambda_i n_i\) with \(\lambda_i\ge0\) and
\(n_i=(E_f-P_{if})/R\) for the active constraints, and
\(\pi_E=-\sum_i\Pi_i\), gives
\[
 \mathcal J_f=\sum_i(P_{if}-E_f)\mathbin\times\lambda_i n_i=0.
\]
The translational covector is conserved, and
\(\dot{\mathcal J}=\sum_i u_i\mathbin\times\Pi_i+
\mu e\mathbin\times\pi_E=0\) along extremals, including for an inactive
pursuer with zero costate. Hence both conditions hold along every
regular characteristic, and the reduced costates are
\begin{equation}\label{eq:preconstruction}
 p_a=(\Pi_2-\Pi_1)\cdot b_1,\qquad
 p_x=\pi_E\cdot b_1,\qquad p_y=\pi_E\cdot b_2.
\end{equation}
In the moving frame, the angular condition reads
\(a(\Pi_2-\Pi_1)\cdot b_2+xp_y-yp_x=0\), that is,
\((\Pi_2-\Pi_1)\cdot b_2=K\). The pursuer covectors expressed in the
frame are therefore exactly \(A_1,A_2\) in \eqref{eq:A}; translation
invariance alone would not give this identification.

Two consequences are worth stating. First, varying reduced headings do
not indicate curved physical motion: curvature in a reduced-coordinate
plot reflects rotation of the frame. Second, the costate
\((p_x,p_y)\) is a fixed inertial vector seen from the rotating frame,
so by \eqref{eq:adjoint} it rotates opposite to the frame, through the
angle \(-\int\omega(t)\,\dd t\), which equals \(-\omega t\) only when
\(\omega\) is constant. Wherever the derivatives exist,
\begin{equation}\label{eq:variableomega}
 \ddot p_x+\omega^2p_x=\dot\omega\,p_y,
\end{equation}
so a constant-rate description of the costates holds only along plays
with constant frame rate.
In practice, an optimal play is computed by determining its terminal
configuration and constant inertial headings and mapping the straight
lines through \eqref{eq:reconstruction}; integrating
\eqref{eq:dynamics}--\eqref{eq:adjoint} is needed only as a consistency
check (Section~\ref{sec:examples}).

\section{The two families of optimal plays}
\label{sec:families}
The nearest point of \(C_T\) to the evader is either a radial point on
one expanded circle, or one of the two intersections of the expanded
circles. Correspondingly, there are two families of regular optimal
plays: a radial race against one pursuer, and a simultaneous exit
through a corner of the sensing union.

\subsection{One active pursuer}
\label{sub:single}
In the simplest play the evader runs straight away from one pursuer,
along the ray from \(P_i\) through \(E\), and that pursuer chases along
the same ray. Suppose pursuer \(i\) is active at termination,
\(E_f-P_{if}=Rn\) with \(\norm n=1\), and the other pursuer is strictly
farther than \(R\). The normalized inertial costates are
\begin{equation}\label{eq:singlep}
 \Pi_i=\frac{n}{\mu-1},\qquad \Pi_j=0,\qquad
 \pi_E=-\frac{n}{\mu-1},
\end{equation}
where \(j\ne i\) denotes the other pursuer. Both active players move along
\(n\), and in retro-time
\begin{equation}\label{eq:singlechar}
 E(\tau)-P_i(\tau)=[R-(\mu-1)\tau]n.
\end{equation}
For given initial positions, let \(d_i=\norm{E^0-P_i^0}\in(0,R)\) and
\begin{equation}\label{eq:singleT}
 n=\frac{E^0-P_i^0}{d_i},\quad
 T_i=\frac{R-d_i}{\mu-1},\quad Q_i=E^0+\mu T_i n.
\end{equation}
The other pursuer cannot interfere when even its best motion cannot
bring \(Q_i\) back into its sensing region, that is, when
\begin{equation}\label{eq:inactive}
 \norm{Q_i-P_j^0}\ge R+T_i
\end{equation}
holds strictly.

\begin{proposition}\label{prop:single}
If \eqref{eq:inactive} holds strictly, then \(V_P=T_i\).
\end{proposition}
\begin{proof}
Let the evader follow \(n\). For any pursuer controls,
\(\norm{E(t)-P_i(t)}\ge d_i+(\mu-1)t\), and by \eqref{eq:inactive} the
other pursuer cannot sense \(Q_i\) at time \(T_i\). Strict separation
persists for a short interval after \(T_i\), so \(t_f\le T_i\). For the
lower bound, pursuer \(i\) uses pure pursuit; its range grows at rate at
most \(\mu-1\), so the evader cannot leave that disk before \(T_i\),
whatever the other pursuer does.
\end{proof}

Equality in \eqref{eq:inactive} makes \(Q_i\) a point of both expanded
circles, and Proposition~\ref{prop:single} no longer applies. This case,
a \emph{single/corner merger}, is treated at the end of
Section~\ref{sub:completion}, once the corner family has been described.
In figures, the inactive
pursuer is drawn heading toward \(Q_i\); this choice is a convenient
representative, not a uniquely optimal motion.

\subsection{Two active pursuers}
\label{sub:corner}
In the second family the evader exits where the two sensing circles
intersect, leaving both disks at the same instant, and each pursuer
moves straight toward that final corner. Use terminal coordinates
\begin{equation}\label{eq:terminal}
 P_{1f}=(-a_f,0),\quad P_{2f}=(a_f,0),\quad
 E_f=(0,b_f),\quad b_f=\sigma\sqrt{R^2-a_f^2},
\end{equation}
with \(0<a_f<R\) and \(\sigma\in\{-1,1\}\). For a weight
\(\beta\in[0,1]\), interpreted below, define
\begin{equation}\label{eq:cornerdefs}
 n_1=\frac{(a_f,b_f)}R,\quad n_2=\frac{(-a_f,b_f)}R,\quad
 \nu=2\beta-1,\quad w=\beta n_1+(1-\beta)n_2,
 \quad c=\norm w=\frac{\sqrt{\nu^2a_f^2+b_f^2}}R.
\end{equation}
When both pursuers are active, \(0<\beta<1\); the endpoints \(\beta=0\)
and \(\beta=1\), at which one costate vanishes, are the single/corner
mergers of Section~\ref{sub:completion}. The normalized inertial
costates and headings are
\begin{equation}\label{eq:cornercontrols}
 L=\frac1{\mu c-1}>0,\quad
 \Pi_1=L\beta n_1,\quad\Pi_2=L(1-\beta)n_2,\quad\pi_E=-Lw,
 \qquad u_1=n_1,\ u_2=n_2,\ e=w/c,
\end{equation}
where the factor \(L\) is fixed by \(\bar H=0\). It measures how
sensitive the escape time is to the positions: it is large when \(\mu c\)
is close to \(1\), and it diverges at the cusps of
Section~\ref{sub:fold}. The evader's heading is
a weighted bisector of the two outward normals. The weight \(\beta\)
measures how much of the terminal sensitivity is carried by pursuer~1,
with \(\beta=1/2\) in the symmetric case. The quantity \(c\le1\) also
has a direct meaning: along a corner branch the distance to the escape
point grows at rate \(D'(T)=1/c\), so \(\mu c>1\) says that the evader's
reach grows faster than its distance to the exit.

The pursuers have equal vertical and opposite horizontal velocities, so
their baseline does not rotate. Substituting the straight lines into
\eqref{eq:reconstruction} gives the closed form
\begin{equation}\label{eq:explicitfan}
 \begin{aligned}
 a(\tau)&=a_f\left(1+\frac\tau R\right),\\
 x(\tau)&=-\frac{\mu\tau\,\nu a_f}{Rc},\\
 y(\tau)&=b_f\left[1+\frac\tau R\left(1-\frac\mu c\right)\right],
 \end{aligned}
\end{equation}
with constant reduced costate
\begin{equation}\label{eq:pcorner}
 (p_a,p_x,p_y)=-\frac{L}{R}(a_f,\nu a_f,b_f).
\end{equation}
The parameters \((a_f,\beta,\tau)\) describe a three-dimensional family,
with upper and lower copies indexed by \(\sigma\). Its members are
straight lines even in the reduced state space, a property specific to
equal pursuer speeds, equal radii, and two active pursuers.

\subsection{Terminal admissibility and completion}
\label{sub:completion}
\begin{figure}[!htb]\centering
\includegraphics[width=\linewidth]{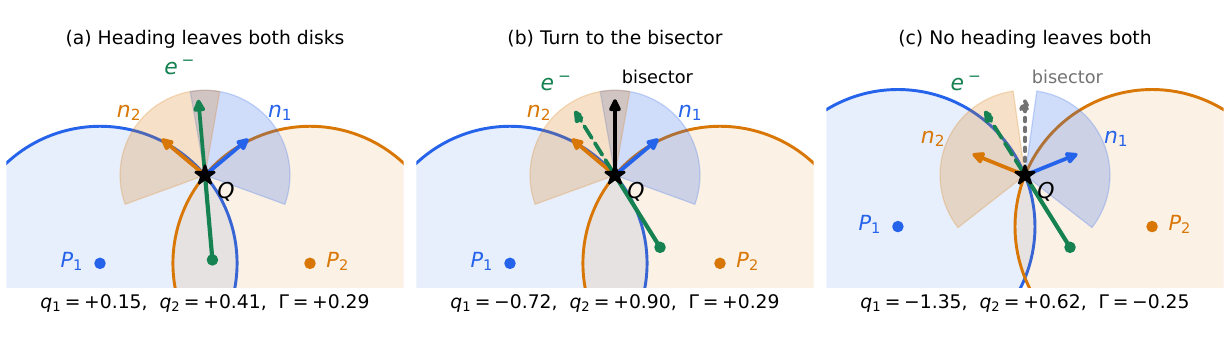}
\caption{Exit through a corner \(Q\) of the sensing union, \(\mu=2\).
The arrows \(n_1,n_2\) are the outward normals of the two disks at
\(Q\). A heading \(e\) leaves disk \(i\) faster than pursuer \(i\) can
follow when \(\mu n_i\cdot e>1\), that is, when it lies within the angle
\(\arccos(1/\mu)\) of \(n_i\) (shaded sectors, blue for disk~1 and orange
for disk~2). (a) The incoming heading \(e^-\) lies in both sectors,
\(q_1,q_2>0\): the evader keeps its heading (region
\(\mathcal C_{\rm in}\)). (b) The incoming heading lies outside the
sector of disk~1, \(q_1<0\), but the sectors overlap, \(\Gamma\ge0\): the
evader turns to the bisector at contact (region
\(\mathcal C_{\rm turn}\)). (c) The pursuers are farther apart, the
normals point farther from each other, and the
sectors do not overlap, \(\Gamma<0\): no heading leaves both disks at
once, and the straight policies we give are only \(\varepsilon\)-optimal
(region \(\mathcal C_\varepsilon\)). The three regions are defined in
Table~\ref{tab:regions}.}
\label{fig:corner}
\end{figure}
If the evader keeps its incoming heading after contact, a transversal
simultaneous exit requires
\begin{equation}\label{eq:rates}
 q_1=\mu n_1\cdot e-1>0,\qquad q_2=\mu n_2\cdot e-1>0,
\end{equation}
or equivalently \(\mu(b_f^2-|\nu|a_f^2)>R^2c\). Here \(q_i\) is the rate
at which the evader's range from pursuer \(i\) grows at contact: the
evader moves away along \(n_i\) at speed \(\mu n_i\cdot e\), and the
pursuer follows at unit speed. A positive \(q_i\) means that the incoming
heading carries the evader out of disk \(i\). Along the family the
ranges satisfy the exact identity
\begin{equation}\label{eq:ranges}
 \norm{E(\tau)-P_i(\tau)}^2
 =R^2-2R q_i\tau+\norm{\mu e-n_i}^2\tau^2,
\end{equation}
where \(\tau\) is the time remaining before contact. For \(q_i>0\) the
right-hand side is below \(R^2\) only for
\(0<\tau<2Rq_i/\norm{\mu e-n_i}^2\). So, when both rates are positive
and the play is traced backward from the corner, it lies strictly inside
both disks until
\[
 \tau_{\rm range}=\min_i\frac{2Rq_i}{\norm{\mu e-n_i}^2},
\]
the time before contact at which one range returns to \(R\). The upper and
lower families meet on \(y=0\). By \eqref{eq:explicitfan}, a play of the
upper family reaches \(y=0\) at the time before contact
\[
 \tau_{\rm sym}=\frac{Rc}{\mu-c},
\]
and beyond it the mirror-image play through the lower corner is at least
as fast, so the family is cut there.

Condition \eqref{eq:rates} is stronger than necessary. The incoming
heading is a weighted bisector, and when \(\beta\) is close to \(0\) or
\(1\) it can point slightly back into one disk. Escape from a corner only
requires \emph{some} direction whose outward rate from each circle is
nonnegative; a zero rate is enough, because the evader is faster than
the pursuers (Proposition~\ref{prop:completion}). The best such
direction is the bisector, which makes equal angles with both normals. Because simple motion places no bound on
turning rate, the evader may switch to the bisector at the instant of
contact.

To state the test, take a two-active candidate of duration \(T\) with
\(P_1^0=(-a,0)\), \(P_2^0=(a,0)\), \(Q=(0,\sigma h)\), \(r=R+T\), and
\(h=\sqrt{r^2-a^2}\). Let \(n_i=(Q-P_i^0)/r\), let
\(e^-=(Q-E^0)/(\mu T)\) be the incoming heading, and set
\begin{equation}\label{eq:completioncriterion}
 \Gamma=\frac{\mu h}{r}-1.
\end{equation}
The bisector \((0,\sigma)\) makes the same angle with both normals, with
\(n_i\cdot(0,\sigma)=h/r\). So \(\Gamma\) is the rate at which an evader
moving along the bisector leaves \emph{both} disks while each pursuer
follows at unit speed; it is the value of \(q_1=q_2\) for the bisector
heading, and no other heading has a larger \(\min_iq_i\).
Figure~\ref{fig:corner} shows the three situations: the incoming heading
already leaves both disks; it does not, but the bisector does; and not
even the bisector does.

\begin{proposition}[Terminal completion]\label{prop:completion}
If \(\Gamma\ge0\), the evader policy
\begin{equation}\label{eq:turnpolicy}
 e(t)=\begin{cases}e^-,&0\le t<T,\\(0,\sigma),&t\ge T,\end{cases}
\end{equation}
which follows the incoming heading \(e^-=(Q-E^0)/(\mu T)\) to the corner
\(Q\) and turns to the bisector \((0,\sigma)\) at contact, guarantees \(t_f\le T\) against every pursuer motion, even if one
incoming rate \(q_i\) is negative. If \(\Gamma<0\), consider the limiting
configuration at time \(T\), with the evader at \(Q\) and
\(P_i(T)=P_i^0+Tn_i\), so that both ranges equal \(R\). If from then on
the pursuers move with constant headings \(u_i=n_i\), then for every
evader control the evader is within distance \(R\) of at least one
pursuer at every time in \((T,T+s_*]\), where
\begin{equation}\label{eq:completiondelay}
 s_*=\frac{2R|\Gamma|}{(\mu+1)^2}.
\end{equation}
\end{proposition}
\begin{proof}
Throughout the proof, \(s>0\) is the time elapsed since contact. The
policy gives \(E(T+s)=Q+\mu s(0,\sigma)\), and for both
pursuers
\begin{equation}\label{eq:uniformgap}
 \norm{E(T+s)-P_i^0}^2-(R+T+s)^2
 =2s(\mu h-r)+(\mu^2-1)s^2.
\end{equation}
This is positive when \(\Gamma\ge0\), including equality, because
\(\mu>1\). Since \(\norm{P_i(T+s)-P_i^0}\le T+s\), the triangle
inequality gives \(\norm{E(T+s)-P_i(T+s)}>R\) for every pursuer motion,
so \(t_f\le T\).

For the second claim, note first that for every \(v\) with
\(\norm v\le1\),
\begin{equation}\label{eq:coneidentity}
 \min_i n_i\cdot v\le\tfrac12(n_1+n_2)\cdot v
 \le\tfrac12\norm{n_1+n_2}=\frac hr,
\end{equation}
because \(n_1+n_2=(0,2\sigma h)/r\); equality holds at
\(v=(0,\sigma)\). At the limiting configuration
\(E(T)-P_i(T)=Q-P_i^0-Tn_i=Rn_i\). Let the evader use any measurable
control on \([T,T+s]\) and let
\(\bar e=s^{-1}\int_T^{T+s}e(t)\,\dd t\), so that \(\norm{\bar e}\le1\)
and \(E(T+s)=Q+\mu s\bar e\). With \(P_i(T+s)=P_i(T)+sn_i\),
\[
 \norm{E(T+s)-P_i(T+s)}^2-R^2
 =\norm{Rn_i+s(\mu\bar e-n_i)}^2-R^2
 =2Rs(\mu n_i\cdot\bar e-1)+s^2\norm{\mu\bar e-n_i}^2.
\]
By \eqref{eq:coneidentity}, some \(i\) has
\(\mu n_i\cdot\bar e-1\le\mu h/r-1=\Gamma\), and
\(\norm{\mu\bar e-n_i}\le\mu+1\). For that \(i\),
\[
 \norm{E(T+s)-P_i(T+s)}^2-R^2\le s\big[2R\Gamma+s(\mu+1)^2\big]<0
 \qquad(0<s<s_*).
\]
At \(s=s_*\) the right-hand side is zero, so the range is at most \(R\).
Hence the evader is detected at every time in \((T,T+s_*]\).
\end{proof}

In terminal coordinates \(h/r=|b_f|/R\), so the test reads
\(\mu|b_f|\ge R\). If both incoming rates are nonnegative, the quadratic
term in \eqref{eq:uniformgap} completes the escape without a turn.

When \(\Gamma<0\), the expanded circles meet at too sharp an angle: even
the bisector cannot leave both of them faster than they advance. The
escape time is still \(U\), and by Proposition~\ref{prop:bounds} the
evader escapes by \(U+\varepsilon\) for every \(\varepsilon>0\), by aiming
at a point just outside the corner. For example, \(a=14\), \(E^0=(10,1)\),
\(R=10\), \(\mu=2\) gives \(U\approx6.07420\) and
\(\Gamma\approx-0.0172945\); at \(\varepsilon=10^{-4}\) the guaranteed
terminal range margin is about \(7.3\times10^{-5}\).

\paragraph{Single/corner mergers}
When equality holds in \eqref{eq:inactive}, the radial exit point \(Q_i\)
of Section~\ref{sub:single} lies on both expanded circles, and
Proposition~\ref{prop:single} does not apply. Two cases arise.
\begin{itemize}
\item \emph{A nondegenerate merger}, where the expanded circles cross
at \(Q_i\) (\(h>0\)). Then
\(Q_i\) is the corner of the two-active family with \(\beta\in\{0,1\}\), the
smooth matching of Lemma~\ref{lem:merger} gives \(U=T_i\), and the other
pursuer is no longer irrelevant: the radial heading coincides with one
outward normal and may point back into the other disk. If
\(\Gamma\ge0\), Proposition~\ref{prop:completion} completes the escape at
\(T_i\); if \(\Gamma<0\), Proposition~\ref{prop:bounds} gives
\(\varepsilon\)-optimal policies, as in the cell \(\mathcal C_\varepsilon\)
of Section~\ref{sub:regions}.
\item \emph{A degenerate axial contact}, where the expanded circles only
touch at \(Q_i\) (\(h=0\)). Then \(F_z\) may merely touch zero at \(T_i\),
the strict-root rule \eqref{eq:strictroot} must be applied, and \(U\) may
exceed \(T_i\). For example, with \(R=1\), \(\mu=2\), \(P_{1,2}^0=(\mp1.1,0)\),
and \(E^0=(0.2,0)\), equality holds with \(T_2=0.1\), but this is the axial
fold of Section~\ref{sub:fold} and \(U=17/30\).
\end{itemize}
For a nondegenerate merger with \(\Gamma<0\), take \(R=1\), \(\mu=2\),
\(P_{1,2}^0=(\mp1.2,0)\), and \(E^0=(0.6,\sqrt{19}/15)\): then \(d_2=2/3\),
\(U=T_2=1/3\), and \(\Gamma=\sqrt{19}/5-1\approx-0.128\). For one with
\(\Gamma>0\) in which the turn is needed, take \(R=1\), \(\mu=2\),
\(P_{1,2}^0=(\mp0.8,0)\), \(T=0.05\), \(Q=(0,h)\) with
\(h=\sqrt{1.05^2-0.8^2}\), and \(E^0=Q-2Tn_2\), where
\(n_i=(Q-P_i^0)/1.05\). Then \(T_2=U=0.05\) and \(Q\) is the corner, but
the radial heading has \(q_1=2n_1\cdot n_2-1\approx-1.322\). Against
pursuers that hold \(u_i=n_i\), the evader that keeps the radial heading
re-enters disk~1 as it leaves disk~2 and first escapes at
\(T-2Rq_1/\norm{2n_2-n_1}^2\approx0.518\). Since
\(\Gamma=2h/1.05-1\approx0.295>0\), the turn to the bisector at time
\(T\) escapes at \(0.05\).

\subsection{Matching a prescribed initial state}
\label{sub:matching}
The families are parameterized by their terminal states; in practice one
starts from an initial state. Let \(P_1^0=(-a_0,0)\), \(P_2^0=(a_0,0)\),
and \(E^0=(x_0,y_0)\). Because the pursuers move at unit speed toward the
final corner, that corner lies on both expanded circles of radius
\(R+T\), and the evader reaches it in time \(T\):
\begin{equation}\label{eq:Q}
 Q_\sigma(T)=(0,\sigma h(T)),\quad
 h(T)=\sqrt{(R+T)^2-a_0^2},\qquad
 x_0^2+[\sigma h(T)-y_0]^2=\mu^2T^2.
\end{equation}
With \(\Lambda(T)=(\mu^2-1)T^2-2RT+a_0^2-R^2-x_0^2-y_0^2\), squaring gives the
quartic
\begin{equation}\label{eq:quartic}
 \Lambda(T)^2-4y_0^2\big[(R+T)^2-a_0^2\big]=0.
\end{equation}
For a root that passes the checks listed below, the play is recovered
from
\begin{align}
 e&=\frac{Q_\sigma(T)-E^0}{\mu T},&
 u_i&=\frac{Q_\sigma(T)-P_i^0}{R+T},\nonumber\\
 \beta&=\frac12\left(1+\frac{\sigma h(T)e_x}{a_0e_y}\right),&
 a_f&=\frac{Ra_0}{R+T},\qquad b_f=\frac{R\sigma h(T)}{R+T}.
 \label{eq:match}
\end{align}
The matching procedure is:
\begin{enumerate}
\item solve \eqref{eq:quartic}, or the quadratic \(\Lambda(T)=0\) when
\(y_0=0\);
\item keep the real positive roots that satisfy the unsquared equation
\eqref{eq:Q};
\item recover the play from \eqref{eq:match} and check \(\norm e=1\),
\(\beta\in[0,1]\), and \(\mu c>1\); if an incoming rate in
\eqref{eq:rates} is negative, apply Proposition~\ref{prop:completion};
\item compare with the single-active times \eqref{eq:singleT} and select
the escape time by the rule \eqref{eq:strictroot}.
\end{enumerate}
Symmetric data can produce two equally good escape corners. The
procedure covers regular states, with \(\mu c>1\) at the selected corner.
It excludes the cusps of Section~\ref{sub:fold}, where \(L\) diverges,
and the exceptional sets of
Section~\ref{sub:regions}; for those, and in general when the escape time
is not the first root, the finite algorithm of Section~\ref{sub:algorithm}
is used.

\paragraph{Running example (continued)}
For \(z=(5,2,2)\), \(R=10\), \(\mu=2\), the quartic \eqref{eq:quartic}
is \((3T^2-20T-83)^2-16[(10+T)^2-25]=0\). Its root
\(T\approx7.40194\) on the upper corner, \(\sigma=1\), satisfies
\eqref{eq:Q} and is the escape time of Section~\ref{sub:distance}.
Formula \eqref{eq:match} gives
\[
 e\approx(-0.13510,\ 0.99083),\qquad
 u_{1,2}=n_{1,2}\approx(\pm0.28732,\ 0.95783),\qquad
 \beta\approx0.27273,
\]
so that \(\nu\approx-0.45454\), \(c\approx0.96670\), \(\mu c\approx1.9334>1\),
\(L\approx1.0714\), and \((a_f,b_f)\approx(2.8732,\ 9.5783)\). The weight
\(1-\beta\approx0.727\) of \(P_2\) is the larger one because the evader is
closer to \(P_2\). Both incoming rates \eqref{eq:rates} are positive,
\(q_1\approx0.8205\) and \(q_2\approx0.9757\), so the evader keeps its
heading through the corner and the state lies in the region
\(\mathcal C_{\rm in}\) of Table~\ref{tab:regions}; the rate after a turn
would be \(\Gamma\approx0.9157\). Since \(a=5\le a_{\rm c}=5\sqrt3\),
Theorem~\ref{thm:value} applies, and the pursuers' optimal feedback is
to head for the current escape corner. Figure~\ref{fig:trajectories}(c)
shows the play.

\section{The value for moderate separations}
\label{sec:regime}
This section proves that \(U\) is the value, with explicit strategies for
both players, when the initial half-separation satisfies
\begin{equation}\label{eq:safe}
 0<a_0\le a_{\rm c}:=R\sqrt{1-\frac1{\mu^2}}=\kappa R.
\end{equation}
Here \(\kappa=\sqrt{1-\mu^{-2}}\). The upper bound follows from
Proposition~\ref{prop:bounds} and the terminal completion. For the lower
bound we follow the escape time \(U(z(t))\) of the current state along
the play and show that, when each pursuer heads toward the current
escape corner, it decreases at most at unit rate whatever the evader
does. Escape requires \(U=0\), so it cannot occur before the initial
value \(U(z(0))\). Two points need care.
On the symmetry line \(y=0\) the upper and lower corners tie, and an
explicit dispersal response is needed. The separation condition must
also persist, which holds because the pursuers only move closer
together.

\subsection{The pursuer feedback}
\label{sub:feedback}
In the cooperative phase the pursuers move mirror-symmetrically about the
perpendicular bisector of their baseline, with common vertical velocity
\(v\) and inward horizontal speed \(\sqrt{1-v^2}\). The baseline then keeps
its direction, and the reduced dynamics \eqref{eq:dynamics} become
\begin{equation}\label{eq:symmetricdynamics}
 \dot a=-\sqrt{1-v^2},\qquad \dot x=\mu e_x,\qquad \dot y=\mu e_y-v.
\end{equation}
Throughout this section \(U=U(z)\) is the escape time of the
\emph{current} state \(z\). In the present range it is the unique root
of \(D(T)=\mu T\), as the proof of Theorem~\ref{thm:value} shows. The
closest point of \(C_T\) at \(T=U\) is
an exposed radial point or a corner, and we call the corresponding case
the \emph{selected branch}: single-active if \(R+U\le r_m\), which
includes a single/corner merger, and corner if \(R+U>r_m\), with \(r_m\)
as in \eqref{eq:rm}. In the second case we say that the corner is
\emph{strictly selected}. Put \(r=R+U\),
\(h=\sqrt{r^2-a^2}\), and \(k(z)=h/r\in(0,1)\).
While the selected branch is a corner, the pursuers use
\begin{equation}\label{eq:tie}
 v=\begin{cases}k\operatorname{sgn}(y),&y\ne0,\\
 \operatorname{clip}(\mu e_y,-k,k),&y=0,\end{cases}\qquad
 u_1=\big(\sqrt{1-v^2},v\big),\quad u_2=\big(-\sqrt{1-v^2},v\big).
\end{equation}
Here \(\operatorname{clip}(\xi,-k,k)=\max\{-k,\min\{k,\xi\}\}\) truncates
\(\xi\) to \([-k,k]\). For \(y\ne0\) this is \(u_i=n_i\): each pursuer heads
toward the current escape corner. On the symmetry line, the pursuers copy the evader's
vertical velocity as far as \(k\) allows; when \(|\mu e_y|\le k\) the
state slides along \(y=0\) while the pursuers close at
\(\dot a=-\sqrt{1-\mu^2e_y^2}\), at least as fast as on either side of the line. At
the first time the selected branch becomes single-active, the
corresponding pursuer switches permanently to pure pursuit.

The feedback \eqref{eq:tie} is discontinuous across \(y=0\) and depends
on the evader's current heading, so its closed loop needs an argument.
With \(\operatorname{Sgn}(y)=\{\operatorname{sgn}(y)\}\) for \(y\ne0\) and
\(\operatorname{Sgn}(0)=[-1,1]\), the vertical equation is the
differential inclusion \(\dot y\in\mu e_y-k\operatorname{Sgn}(y)\), and
the realized control is \(v=\mu e_y-\dot y\).

\begin{lemma}[Realization of the feedback]\label{lem:realization}
Let \(\mathcal O\) be the set of states with \(a>0\) and \(U>0\) at which
a corner is strictly selected, \(R+U(z)>r_m(z)\), with \(\mu c>1\) at the
selected corner, that is, \(\partial_TF_z(U(z))<0\), and at which the
contact residual is positive before the escape time:
\begin{equation}\label{eq:realizationdomain}
 F_z(\tau)>0\qquad(0\le\tau<U(z)).
\end{equation}
Let \(\mathcal C\subset\mathcal O\) be compact. For every measurable evader
heading \(e(\cdot)\) and every \(z_0\in\mathcal C\), there are exactly one
absolutely continuous \(z(\cdot)\) with \(z(0)=z_0\) and one measurable
\(v(\cdot)\) such that, for as long as \(z(t)\in\mathcal C\),
\eqref{eq:symmetricdynamics} holds almost everywhere and
\[
 v=k(z)\operatorname{sgn}(y)\ \text{a.e. on }\{y\ne0\},\qquad
 v=\mu e_y\ \text{and}\ |\mu e_y|\le k(z)\ \text{a.e. on }\{y=0\}.
\]
In particular \(v\) agrees almost everywhere with \eqref{eq:tie}, and
\(z\) and \(v\) on \([0,t]\) depend only on \(e\) on \([0,t]\). Hence
\eqref{eq:tie} defines a nonanticipative pursuer strategy.
\end{lemma}
Strict selection makes the corner expression of \eqref{eq:Dexplicit}
represent \(D_z\) near the selected root, condition
\eqref{eq:realizationdomain} keeps that root selected under small
perturbations of the state, and \(\partial_TF_z(U)<0\) lets the implicit
function theorem continue it. So \(\mathcal O\) is open, and
on it the selected corner root is \(C^1\) in the reflected variables
\((a,x,Y)\), with \(Y=|y|\) (Appendix~\ref{app:realization}).
Consequently \(U\) and \(k\) are locally Lipschitz in \((a,x,y)\), with
smooth branches on either side of the upper/lower tie \(y=0\); across a
tie, \(\partial_yU\) jumps from \(Lh/r\) for \(y<0\) to \(-Lh/r\) for
\(y>0\), so \(U\) is not differentiable in \(y\) there. Every interior
state of Theorem~\ref{thm:value} at which a corner is strictly selected
lies in \(\mathcal O\), because there the root of \(D(T)=\mu T\) is unique
and \(D'<\mu\); so does every such state of the invariant domain of
Theorem~\ref{thm:extension}. Nondegenerate single/corner mergers,
\(R+U=r_m\), are not in \(\mathcal O\): arbitrarily close states select
the radial branch. They lie on its boundary and are handled separately,
through Lemma~\ref{lem:merger} and the handoff to pure pursuit. The fold states of Section~\ref{sub:fold}
are excluded: there \(F_z\) touches zero before \(U\), the selected root
is still simple with \(\mu c>1\), but nearby states on one side escape
near the earlier contact time, and \(k\) jumps.

\begin{proof}[Idea of the proof]
Freeze \(k\) along a given continuous path. The vertical equation is then
a differential inclusion whose right-hand side is monotone in \(y\). It
has a solution by Filippov's theorem~\cite{filippov}, and monotonicity
makes the solution and its realized control unique. On the set where
\(y=0\), \(\dot y=0\) almost everywhere, which forces \(v=\mu e_y\)
there. A contraction argument over short time intervals then
closes the loop through \(k(z)\). Every step uses the evader's heading
only up to the current time. Appendix~\ref{app:realization} gives the
details.
\end{proof}

\subsection{The theorem}
\begin{theorem}[Moderate separations]\label{thm:value}
Assume the evader starts in the interior of the sensing union and
\eqref{eq:safe} holds. Then \(V_P\) is the unique positive solution of
\begin{equation}\label{eq:value}
 D(T)=\mu T,
\end{equation}
selected among the single-active and corner branches of
Section~\ref{sec:families}. Every selected corner contact, including a
single/corner merger, can be completed immediately, possibly by a
terminal evader turn, and the
optimal incoming plays are given by \eqref{eq:straight} and
\eqref{eq:explicitfan}. Coincident pursuers, \(a_0=0\), form a single
disk, for which the two bounds of Proposition~\ref{prop:bounds} coincide
and \(V_P=(R-d)/(\mu-1)\).
\end{theorem}

\begin{proof}
\emph{Uniqueness of the root.}
At an intersection of the expanded circles let \(r=R+T\),
\(h=\sqrt{r^2-a_0^2}\), and let \(n_1,n_2\) be the outward normals. For
\(T>0\), \eqref{eq:safe} gives \(h/r>1/\mu\). A closest exterior point
has \(e=w/c\) with \(w=\beta n_1+(1-\beta)n_2\), and
\(c\ge h/r>1/\mu\). The envelope derivative is \(D'(T)=1\) on an exposed
branch and \(D'(T)=1/c<\mu\) on a corner branch. These bounds persist
across changes of active set by one-sided differentiation, and
Lemma~\ref{lem:ties} gives continuity. Hence \(D(T)-\mu T\) is strictly
decreasing. It is positive at \(T=0\) and negative beyond \(B\) by
Proposition~\ref{prop:bounds}, so it has exactly one root.

\emph{Evader upper bound.}
Let \(Q\) be a closest point at the root and \(e=(Q-E^0)/(\mu T)\). On
a strictly exposed branch, where \eqref{eq:inactive} is strict, the
evader keeps the radial heading. At a corner, and at a merger, where
\(Q\) is also a corner, \(\Gamma>0\) for \(T>0\), so
Proposition~\ref{prop:completion} applies and the evader turns to the
bisector at contact. Either way
\(t_f\le T\) against every pursuer motion.

\emph{Pursuer lower bound away from ties.}
Regard the root \(U\) as a function of the current positions. The
envelope calculation gives
\[
 \nabla_{P_i}U=L\beta_i n_i,\qquad
 \nabla_EU=-Lw,\qquad L=(\mu c-1)^{-1},
\]
with \((\beta_1,\beta_2)=(\beta,1-\beta)\). Off the symmetry line the
feedback \eqref{eq:tie} is \(u_i=n_i\), which gives, for every evader
heading,
\[
 \frac{\dd U}{\dd t}=L-\mu Lw\cdot e\ge L-\mu Lc=-1.
\]
On a single-active branch the pure-pursuit lower bound of
Proposition~\ref{prop:bounds} applies directly. At a single/corner merger the multiplier of
the departing constraint tends to zero, and the gradients agree by
Lemma~\ref{lem:merger}. During a
two-active phase the baseline does not rotate and
\(\dot a=-a/(R+U)<0\), so \eqref{eq:safe} remains valid.

\emph{The dispersal line.}
By Lemma~\ref{lem:ties}, the remaining tie is between the upper and
lower escape points when \(y=0\). There \(|x|\le a\) and
\[
 r=R+U,\quad h=\sqrt{r^2-a^2},\quad
 \mu U=\sqrt{x^2+h^2},\quad c=\sqrt{x^2+h^2}/r.
\]
The two one-sided gradients share the components \(p_a=-La/r\) and
\(p_x=Lx/r\) and have opposite \(p_y=\mp Lh/r\). Under the response
\eqref{eq:tie} on \(y=0\), with \(s=\sqrt{1-v^2}\), the directional
derivative of the smaller branch is
\begin{equation}\label{eq:tieslack}
 \frac{\dd U}{\dd t}
 =\frac{L}{r}\{a s+\mu x e_x-h|\mu e_y-v|\}.
\end{equation}
If \(|\mu e_y|\ge h/r\), the braces equal
\(r+\mu(xe_x-h|e_y|)\ge r-\mu\sqrt{x^2+h^2}\), giving
\(\dd U/\dd t\ge-1\). If \(|\mu e_y|\le h/r\), put \(\zeta=|e_y|\); the
braces are bounded below by
\[
 f(\zeta)=a\sqrt{1-\mu^2\zeta^2}-\mu|x|\sqrt{1-\zeta^2},
\]
which is decreasing on \(0\le\zeta\le h/(\mu r)\) because \(|x|\le a\) and
\(\mu>1\). Its minimum is at the endpoint, already covered by the
previous case. A fixed upper or lower pursuer heading would not give
this bound against all deviations, which is why the response must use
the evader's current heading.

\emph{Realization of the closed loop.}
The feedback \eqref{eq:tie} is used only while a corner is strictly
selected, that is, while the state is in the open set \(\mathcal O\) of
Lemma~\ref{lem:realization}; this phase ends when the state reaches a
merger, where the corresponding pursuer commits to pure pursuit, when
the pursuers come to coincide, or when \(U\to0\). Applying the lemma on an increasing sequence of compact
subsets of \(\mathcal O\) shows that, for every measurable evader
heading, the feedback produces a unique absolutely continuous
closed-loop trajectory throughout this phase, its controls depend only on the evader's past
and present headings, and on \(\{y=0\}\) the realized vertical velocity
is \(v=\mu e_y\) almost everywhere. The estimates above therefore hold
at almost every time, even if the heading crosses \(\pm k\) infinitely
often.

On each compact set where \(U>0\), the selected root is locally
Lipschitz: it is the minimum of finitely many smooth branches, with
matching gradients at single/corner mergers
(Lemma~\ref{lem:merger}). Hence \(U(z(t))\) is
absolutely continuous, and at almost every time its derivative is
\(\min_j\nabla U_j\cdot\dot z\) over the active branches \(U_j\). The bounds
above give
\[
 U(z(t))-U(z(s))\ge-(t-s)
\]
throughout the cooperative phase, with no assumption on the number of
switches.

If the strategy commits to pure pursuit at time \(s\), the remaining
single-disk guarantee is \(T_i(z(s))=U(z(s))\), and the pure-pursuit
lower bound in Proposition~\ref{prop:bounds}, which needs no strict
exposure and so covers a merger, gives \(t_f\ge s+U(z(s))\ge U(z(0))\).
If instead the pursuers come to coincide at a time \(s\), which can
happen while the state slides on \(y=0\), then \(x(s)=0\), because
\(|x|<a\) throughout the phase, and the two disks have become one. As
\(a\to0\), \(D_z(T)\to R+T-|y|\) uniformly on \([0,B]\), so
\(U(z(t))\to U(z(s))=(R-|y(s)|)/(\mu-1)\) and the estimate above extends
to \(t=s\). Either pursuer then commits to pure pursuit, and
Proposition~\ref{prop:bounds} again gives
\(t_f\ge s+U(z(s))\ge U(z(0))\).
Otherwise let \(t\) tend to the escape time, where \(U(z(t))\to0\).
Thus one nonanticipative pursuer strategy gives
\(J_z(e,\mathcal B[e])\ge U(z)\) for every \(e\), and the completed
evader policy \(e^*\) of the upper-bound step gives
\(J_z(e^*,u)\le U(z)\) for every \(u\). Hence
\(V_P=U\).
\end{proof}

\begin{remark}[Origin of the threshold]
At an intersection of the expanded circles, \(h/r\) is the component of
each outward normal along the bisector. Moving along the bisector, the
evader gains distance from both circles at rate \(\mu h/r\) while each
circle advances at unit rate. The condition \(\mu h/r>1\) is
\(a<\kappa r\), and it holds for all horizons once it holds at
\(r=R\). Below \(a_{\rm c}\) the bisector therefore always outruns both
expanding circles, so every corner can be completed and the envelope
slope stays below \(\mu\). The theorem is a sufficient condition:
Theorem~\ref{thm:globalvalue} shows that \(V_P=U\) for every
separation, and Theorem~\ref{thm:extension} in Appendix~\ref{app:extension} shows that
the explicit feedback above remains optimal on an invariant region
beyond \(a_{\rm c}\).
\end{remark}

\subsection{When the incoming heading suffices}
A smaller threshold governs whether the evader can keep its incoming
heading through a corner. Let
\begin{equation}\label{eq:transversalitythreshold}
 a_{\rm tr}=R\sqrt{\frac{\mu-1}{2\mu}}.
\end{equation}
At a corner of expanded radius \(r=R+T\), minimizing the incoming outward
rate over both constraints and all \(\beta\in[0,1]\) gives
\begin{equation}\label{eq:worstrate}
 \min_{\beta\in[0,1]}\min_i(\mu n_i\cdot e^- -1)
 =\mu\left(1-\frac{2a_0^2}{r^2}\right)-1,
\end{equation}
attained when the incoming heading coincides with one normal and the
rate is measured at the other. Therefore \(a_0\le a_{\rm tr}\) ensures
\(q_i>0\) for every \(T>0\).

This threshold is sharp. For any \(a_0\in(a_{\rm tr},a_{\rm c})\), choose
\(\beta>0\) small and then \(T>0\) small enough that \(q_1<0<q_2\), and
set \(E^0=Q-\mu Te^-\). A sufficiently short displacement in the interior
of the corner's normal cone keeps \(Q\) as the nearest exterior point; by
\eqref{eq:ranges} the state lies inside disk~2 and outside disk~1, and
Theorem~\ref{thm:value} selects this corner. For example,
\(R=10\), \(\mu=2\), \(a_0=5.01\), \(\beta=0.001\), \(T=0.0025\) give
\[
 E^0\approx(0.002500618907,8.653031621113),\qquad
 (q_1,q_2)\approx(-0.001998310875,0.999999247497).
\]
The corner is still completed by a terminal turn, so the sharpness
concerns transversality of the incoming heading, not the value formula.
States initially inside both disks never need the turn:
\eqref{eq:ranges} forces both \(q_i>0\) when both initial ranges are
below \(R\).

\section{Complete partition of the state space}
\label{sec:partition}
For larger separations the escape time is no longer the first root of
\(D(T)=\mu T\). The simplest case shows why. Let the sensing disks be
disjoint, \(a>R\), and let the evader sit on the baseline between them,
inside the disk of \(P_2\). Its quickest option seems to be the gap at
\(x=0\). As the horizon grows, however, the expanded disks grow toward
each other and close the gap at time \(a-R\). If the evader would arrive
exactly when the gap closes, \(F_z\) touches zero and immediately becomes
positive again, as in Figure~\ref{fig:residual}(b): the evader is caught
in the closing gap, and its real escape goes around the merged disks,
through a corner, much later. This is why \eqref{eq:strictroot} uses the
first time at which \(F_z\) is \emph{strictly} negative, and why states on
either side of such a tangency can have very different escape times.

Using the closed form \eqref{eq:Dexplicit} of \(D_z\), this section turns
\eqref{eq:strictroot} into a finite algorithm and describes the regions
and surfaces that partition the state space, including the fold formed by
these tangencies. Throughout, \(X=|x|\), \(Y=|y|\), \(m=\mu^2-1\),
\(r=R+T\), \(h=\sqrt{r^2-a^2}\), and \(d\), \(r_m\), \(t_m\) are as in
Section~\ref{sub:distance}.

\subsection{A finite algorithm for the escape time}
\label{sub:algorithm}
The exposed piece of \(F_z\) is affine with slope \(1-\mu<0\). On the
corner piece,
\begin{equation}\label{eq:Dderivatives}
 D'(T)=\frac{r(h-Y)}{hD},\qquad
 D''(T)=\frac{X^2(h^3+a^2Y)-a^2(h-Y)^3}{h^3D^3}.
\end{equation}
For \(Y>0\), the ratio \((h-Y)^3/(h^3+a^2Y)\) is strictly increasing for
\(h>Y\), so \(D''\) changes sign at most once, from positive to
negative, and \(F'\) has at most two zeros on the corner piece. For
\(Y=0\) the smooth corner portion has \(D''<0\) when \(X<a\), and the
join at \(r=a\) is a possible kink. So \(F_z\) has one of the three shapes
of Figure~\ref{fig:residual}: it decreases through zero once, or it dips
to a local minimum and rises before its final crossing, or these two
features merge. These shapes make the procedure of
Algorithm~\ref{alg:escape} exact.

\begin{algorithm}[!t]
\caption{Escape time \(U(z)\).}
\label{alg:escape}
\emph{Input:} a state \(z=(a,x,y)\), the sensing radius \(R\), and the speed
ratio \(\mu>1\). \emph{Output:} the escape time \(U(z)\).
\begin{enumerate}
\item \emph{Special states.} If \(a=0\), if the evader is at a pursuer, if
it is strictly outside the sensing union or on its usable boundary, or
if it is at a nonusable double-boundary point, return the value given
for that case in Table~\ref{tab:strata} or in the paragraph before
it.
\item \emph{The two pieces of \(D_z\).} With \(X=|x|\) and \(Y=|y|\), form the
radial piece \(r-d\) and the corner piece \(\sqrt{X^2+(h-Y)^2}\) of
\eqref{eq:Dexplicit}, and the join time \(t_m=r_m-R\) of \eqref{eq:rm}.
\item \emph{Stationary horizons.} Find the roots \(h\) of the quartic
 \begin{equation}\label{eq:stationaryquartic}
  (a^2-mh^2)(h-Y)^2-\mu^2h^2X^2=0,
 \end{equation}
 and keep those with \(h>Y\), \(r\ge r_m\), \(0<T\le B\), and the
 unsquared condition \(D'(T)=\mu\) of \eqref{eq:Dderivatives}.
\item \emph{Monotone intervals.} Sort \(0\), the join time \(t_m\) if it is
positive, the stationary horizons, and a time slightly beyond \(B\). On
each interval between consecutive horizons, \(F_z\) is monotone.
\item \emph{First strict escape.} Take the first interval at whose right
end \(F_z<0\); by Proposition~\ref{prop:bounds} one exists. If \(F_z=0\) at
its left end, return the left end; this is a cusp or a crossing at a
horizon of the list. Otherwise return the unique root of \(F_z\) in the
interval, found by a bracketed scalar solve.
\end{enumerate}
\end{algorithm}

In step~5 of Algorithm~\ref{alg:escape}, every earlier horizon at which \(F_z=0\) has \(F_z\ge0\) on both
sides; these isolated noncrossing contacts, such as the fold contact of
Figure~\ref{fig:residual}(b), are skipped. Uniform sampling in \(T\), by
contrast, could miss a narrow negative interval.

The algorithm computes the value at every state. It does not by itself
produce a policy that attains it. For regular states, candidate plays
are recovered from the radial formula \eqref{eq:singleT} or from the
matching of Section~\ref{sub:matching}; their completion and attainment
are decided by Propositions~\ref{prop:bounds}--\ref{prop:completion} and
the classification of Section~\ref{sub:regions}. In
\(\mathcal C_\varepsilon\) the construction supplies
\(\varepsilon\)-optimal policies. Cusps and the other exceptional sets
are outside the scope of the regular matching procedure.

The algorithm is exact. In floating point, the tests \(F_z=0\) and
\(F_z<0\) in step~5 need a tolerance, and within it the selection near a
fold is a convention, not a certificate; Section~\ref{sub:validation}
describes the convention used and the flag that marks such states.

\subsection{Regions}
\label{sub:regions}
For a selected corner or a merger with \(U>0\), write
\[
 r=R+U,\quad h=\sqrt{r^2-a^2},\quad
 Q=(0,\operatorname{sgn}(y)h),\quad e^-=(Q-E)/(\mu U),
\]
with either sign when \(y=0\). Two quantities decide how the escape is
completed: the smaller outward rate \(q_{\min}\) of the incoming heading,
and the rate \(\Gamma\) after a turn to the bisector,
\begin{equation}\label{eq:partitiondiagnostics}
 q_{\min}=\frac{h(h-Y)-aX}{rU}-1,\qquad
 \Gamma=\frac{\mu h}{r}-1.
\end{equation}
Table~\ref{tab:regions} assigns the states with positive escape time to
regions according to how the escape is completed. It covers every such
state outside the exceptional sets of Table~\ref{tab:strata}, and also
the upper/lower ties and the nondegenerate single/corner mergers, which
appear in both tables: they lie in the corner regions, and
Table~\ref{tab:strata} only adds the statements specific to them. Write \(T_s=(R-d)/(\mu-1)\) for the
single-pursuer time. On a nondegenerate single/corner merger, one with
\(h>0\), \(U=T_s\) and \(R+U=r_m\); the value is that of the single family, but the exit point
is the corner \(Q\), so the completion is decided by \(q_{\min}\) and
\(\Gamma\) exactly as for a corner. We therefore list mergers with the
corner cells, which are defined by the completion of the escape rather
than by the family. At a degenerate axial contact, \(h=0\), the zero of
\(F_z\) at \(T_s\) may be isolated; the strict-root rule
\eqref{eq:strictroot} then skips it, \(U\ne T_s\), and the state belongs
to the axial fold of Section~\ref{sub:fold}.

\begin{table}[htbp]\centering\small
\begin{tabular}{@{}p{.12\linewidth}p{.37\linewidth}p{.43\linewidth}@{}}
\toprule
Region & Definition & Evader play\\
\midrule
\(\mathcal S_1\) & \(x<0\), \(U=T_s\), \(R+U<r_m\)
 & Radial escape from \(P_1\); \(V_P=T_s\).\\
\(\mathcal S_2\) & \(x>0\), \(U=T_s\), \(R+U<r_m\)
 & Radial escape from \(P_2\); \(V_P=T_s\).\\
\(\mathcal C_{\rm in}\) & Selected corner or merger, \(q_{\min}\ge0\)
 & Straight to the corner; the incoming heading completes the escape.\\
\(\mathcal C_{\rm turn}\) & Selected corner or merger, \(q_{\min}<0\), \(\Gamma\ge0\)
 & Straight to the corner, then a terminal turn to the bisector.\\
\(\mathcal C_{\varepsilon}\) & Selected corner or merger, \(\Gamma<0\)
 & Straight to a point just outside the corner; \(\varepsilon\)-optimal.\\
\bottomrule
\end{tabular}
\caption{Regions of the state space. On \(\mathcal S_i\) the value is
the single-pursuer time \(T_s\), and on the three corner regions it is
the corner root \(U\); Theorems~\ref{thm:globalvalue}
and~\ref{thm:reverse} show that \(U=V_P=V_E\) in all of them.
Table~\ref{tab:strata} lists the exceptional sets.}
\label{tab:regions}
\end{table}

The escape time is zero on the strict exterior and on the \emph{usable}
part of the boundary of the sensing union, from which the evader can
leave at once. Table~\ref{tab:strata} lists the remaining exceptional
sets. At the double sensing-boundary points \(x=0\),
\(Y=\sqrt{R^2-a^2}\), the evader touches both sensing circles at their
intersection. These points are usable if \(\mu Y\ge R\); otherwise
\eqref{eq:strictroot} gives
\begin{equation}\label{eq:nonusableboundaryU}
 U=\frac{2(R-\mu Y)}{\mu^2-1}>0.
\end{equation}

\subsection{Separating surfaces}
\label{sub:surfaces}
For \(0<X<a\), the merger of the single and corner families is
\begin{equation}\label{eq:singlecornerpartition}
 (\mu R-d)(a-X)=(\mu-1)ad,
 \quad\text{equivalently}\quad
 Y^2=(a-X)^2\left[\frac{\mu^2R^2}{(\mu a-X)^2}-1\right],
\end{equation}
restricted to the sensing union and nonnegative radicand. Its axial
endpoint \(Y=0\), \(r_m=a\), is exceptional: if it is an isolated
contact, the strict-root rule overrides the single-family assignment.
The boundary between \(\mathcal C_{\rm in}\) and \(\mathcal C_{\rm turn}\)
is
\begin{equation}\label{eq:incomingpartition}
 h(h-Y)-aX=rU,\qquad X^2+(h-Y)^2=\mu^2U^2,
\end{equation}
and the boundary between \(\mathcal C_{\rm turn}\) and
\(\mathcal C_\varepsilon\) has the simple form
\begin{equation}\label{eq:completionpartition}
 U=\frac a\kappa-R,\qquad
 X^2+\left(\frac a{\mu\kappa}-Y\right)^2
       =\mu^2\left(\frac a\kappa-R\right)^2,
\end{equation}
in both cases restricted to selected corners with \(U\ge0\).

\begin{table}[!htb]\centering\small
\begin{tabular}{@{}>{\raggedright\arraybackslash}p{.25\linewidth}>{\raggedright\arraybackslash}p{.27\linewidth}>{\raggedright\arraybackslash}p{.40\linewidth}@{}}
\toprule
Exceptional set & Value \(U=V_P=V_E\) & Evader policy or interpretation\\
\midrule
Coincident pursuers, \(a=0\) & \((R-\sqrt{x^2+y^2})_+/(\mu-1)\)
 & Radial escape from the common position.\\
Evader at a pursuer, \(E=P_i\) & \(B\)
 & Escape directed away from the other pursuer.\\
Upper/lower tie, \(y=0\), corner selected & Corner root, the same
 for both corners
 & Either corner; the pursuers use the dispersal response
   \eqref{eq:tie}, except at an axial fold, where
   \eqref{eq:axialfoldguard} takes precedence.\\
Nonusable double boundary & \eqref{eq:nonusableboundaryU}
 & Positive value despite initial contact with both circles.\\
Nondegenerate single/corner merger (\(h>0\)) & \(T_s=(R-d)/(\mu-1)\)
 & Radial heading to the corner; then a turn if \(\Gamma\ge0\),
   \(\varepsilon\)-optimal policies if \(\Gamma<0\).\\
Minimum fold, including the axial fold & Later root \(T_L\),
 \eqref{eq:laterfoldroot}
 & The earlier contact at \(T_F\) cannot be exploited.\\
Cusp arc, \(0<k<1/\mu\) in \eqref{eq:cuspparam}
 & Coalesced root \(T_F=T_L\)
 & Degenerate chart, \(\mu c=1\).\\
\bottomrule
\end{tabular}
\caption{Exceptional sets, with the value given by
Theorems~\ref{thm:globalvalue} and~\ref{thm:reverse}. Degenerate axial
contacts (\(h=0\)) are classified by the strict-root rule; an isolated
contact belongs to the axial-fold case. Where rows overlap, the fold
rows take precedence over the tie row. The two endpoints of the cusp arc
are the axial cusp and a usable double-boundary point, where \(U=0\).
Mergers, folds, and cusps are described in
Sections~\ref{sub:surfaces} and~\ref{sub:fold}.}
\label{tab:strata}
\end{table}

\subsection{The fold and its cusp}
\label{sub:fold}
The corner piece of \(F_z\) is first convex and then concave in \(T\),
by \eqref{eq:Dderivatives}. It can therefore dip to an interior minimum,
touch zero, and rise again before its final crossing. The states at
which the minimum touches zero form a fold
(Figure~\ref{fig:residual}(b)): on one side the evader
escapes near the earlier time \(T_F\), and on the other it must wait for
the later crossing \(T_L\). Where the minimum and the later crossing
merge, the fold ends in a cusp (Figure~\ref{fig:residual}(c)).

At a smooth double root, \(F(T_F)=F'(T_F)=0\). With \(r=R+T_F\),
\(h=\sqrt{r^2-a^2}\), and \(k=h/r\), eliminating \(X,Y\) gives the sheet
\begin{equation}\label{eq:foldparam}
 \begin{aligned}
 X_F(a,r)&=\mu(r-R)\sqrt{1-\mu^2+\frac{\mu^2a^2}{r^2}},\\
 Y_F(a,r)&=\frac{\sqrt{r^2-a^2}}r
                  [\mu^2R-(\mu^2-1)r].
 \end{aligned}
\end{equation}
At a stationary root, \eqref{eq:Dderivatives} reduces to
\[
 \operatorname{sgn} F''(T_F)
 =\operatorname{sgn}\big[\mu^2Ra^2-(\mu^2-1)r^3\big],
\]
so the minimum fold, which creates an isolated contact that
\eqref{eq:strictroot} must skip, exists precisely for
\begin{equation}\label{eq:foldrange}
 \kappa R<a<a_*:=\frac{\mu^2R}{\mu^2-1},\qquad
 \max(R,a)<r<r_c(a):=
             \left(\frac{\mu^2Ra^2}{\mu^2-1}\right)^{1/3}.
\end{equation}
The maximum-fold portion \(r>r_c\), when admissible, joins later roots
after a strict escape interval already exists, so it does not affect
\(U\). The minimum fold ends on the cusp arc \(F=F'=F''=0\),
parameterized by
\begin{equation}\label{eq:cuspparam}
 \begin{aligned}
 a_{\rm cusp}(k)&=\frac{\mu^2R}{\mu^2-1}(1-k^2)^{3/2},\\
 X_c(k)&=\frac{\mu R}{\mu^2-1}(1-\mu^2k^2)^{3/2},\\
 Y_c(k)&=\mu^2Rk^3,\qquad 0<k<1/\mu,
 \end{aligned}
\end{equation}
and reflecting \(X\) and \(Y\) gives the four copies in the full state
space. The closure of the arc has two endpoints, which are degenerate
and are not smooth triple roots. At \(k=0\) it reaches the axial point
\((a,X,Y)=(a_*,\mu R/(\mu^2-1),0)\), the \emph{axial cusp}: this is the
limit of the axial fold described below, where the value jump closes,
and \(F_z\) has a kink at \(T_F\), with one-sided slopes \(1-\mu\) and
\(0\). At \(k=1/\mu\) it reaches the double-boundary point
\((\kappa R,0,R/\mu)\), where \(T_F=0\); this point is usable, so
\(U=0\) there. On the minimum fold the later crossing has the closed form
\begin{equation}\label{eq:laterfoldroot}
 T_L=T_F+
 \frac{2}{\mu^2-1}\left[A-\mu k\sqrt{RA}\right],\qquad A=R-(\mu^2-1)T_F,
\end{equation}
with \(T_L>T_F\) on the open fold and \(T_L=T_F\) at the cusp; the
escape time on the fold is \(U=T_L\). The boundary case \(r=a\) of
\eqref{eq:foldparam}, in which \(h=0\) and \(k=0\), is the \emph{axial
fold}
\[
 Y=0,\qquad X=\mu(a-R),\qquad R<a<a_*,\qquad T_F=a-R .
\]
There the two expanded circles touch at the instant the evader reaches
the touching point, and \(F_z\) has a kink at \(T_F\) instead of a smooth
minimum.

\begin{figure}[htbp]\centering
\includegraphics[width=\linewidth]{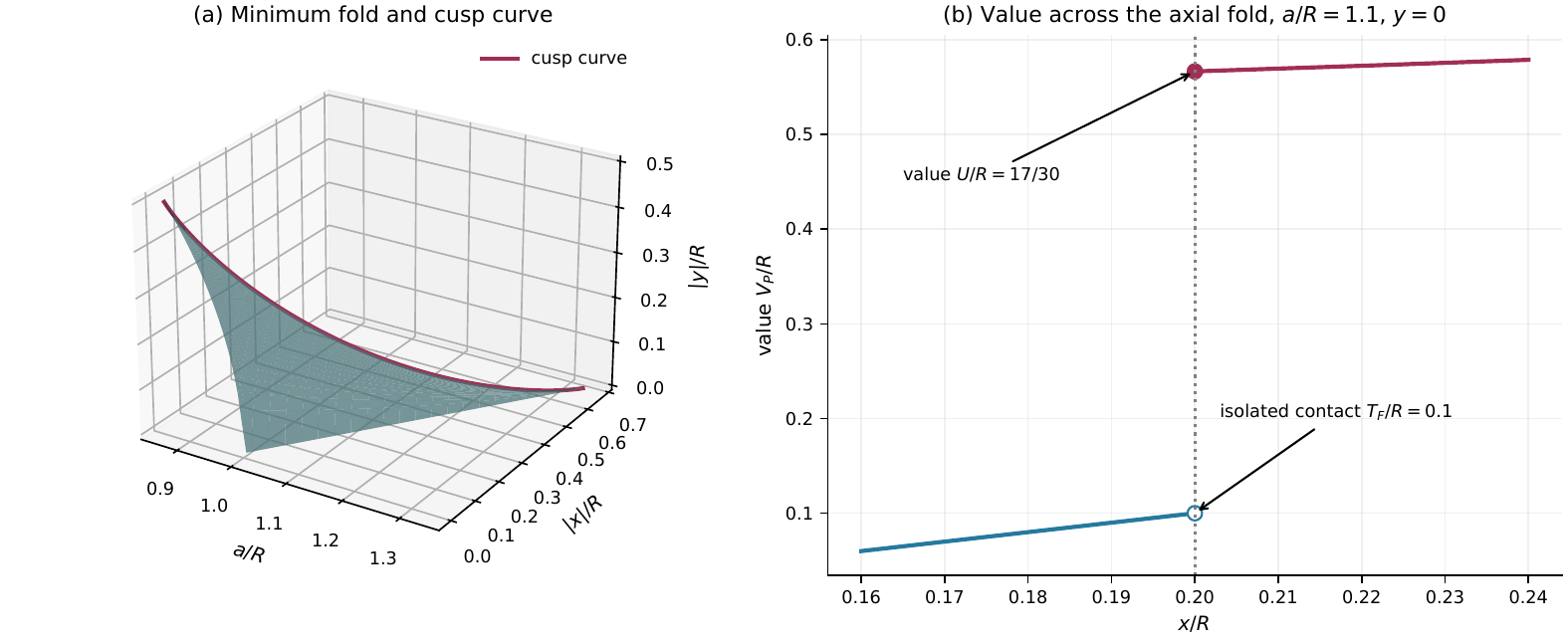}
\caption{Left: the minimum-fold sheet and its cusp curve in one symmetry
sector, \(\mu=2\). Right: escape time across the axial fold at
\(a=1.1R\), \(y=0\). The open circle is the isolated contact at
\(T_F=0.1\); the value jumps to \(T_L=17/30\).}
\label{fig:valuejump}
\end{figure}

\paragraph{A second example: a value jump}
Take \(R=1\), \(\mu=2\), \(a=1.1\), and \(y=0\). The physical disks are
disjoint, with a gap of width \(0.2\) between them on the baseline, and
the expanded circles first touch, at the origin, at \(T=0.1\). The axial
fold is at \(x=0.2\): the evader reaches the origin exactly at
\(T_F=0.1\), so \(F(T_F)=0\). At that instant, however, the gap closes,
and immediately afterwards the expanded disks overlap and cover the
origin, so \(F\) becomes positive again. This isolated contact does not
let the evader escape: it is caught in the closing gap, and its escape
goes around the merged disks through a corner, at
\[
 U=T_L=\frac{17}{30}\approx0.566667.
\]
From \(x=0.1999\) the evader reaches the gap just before it closes and
escapes at the single-pursuer time \(0.0999\); from \(x=0.2001\) it arrives
just too late, and the escape time is \(0.566695\).
Theorem~\ref{thm:globalvalue} shows that this is a jump of the value
itself (Figure~\ref{fig:valuejump}).

\subsection{Separation regimes}
\label{sub:regimes}
\begin{figure}[tbp]\centering
\includegraphics[width=\linewidth]{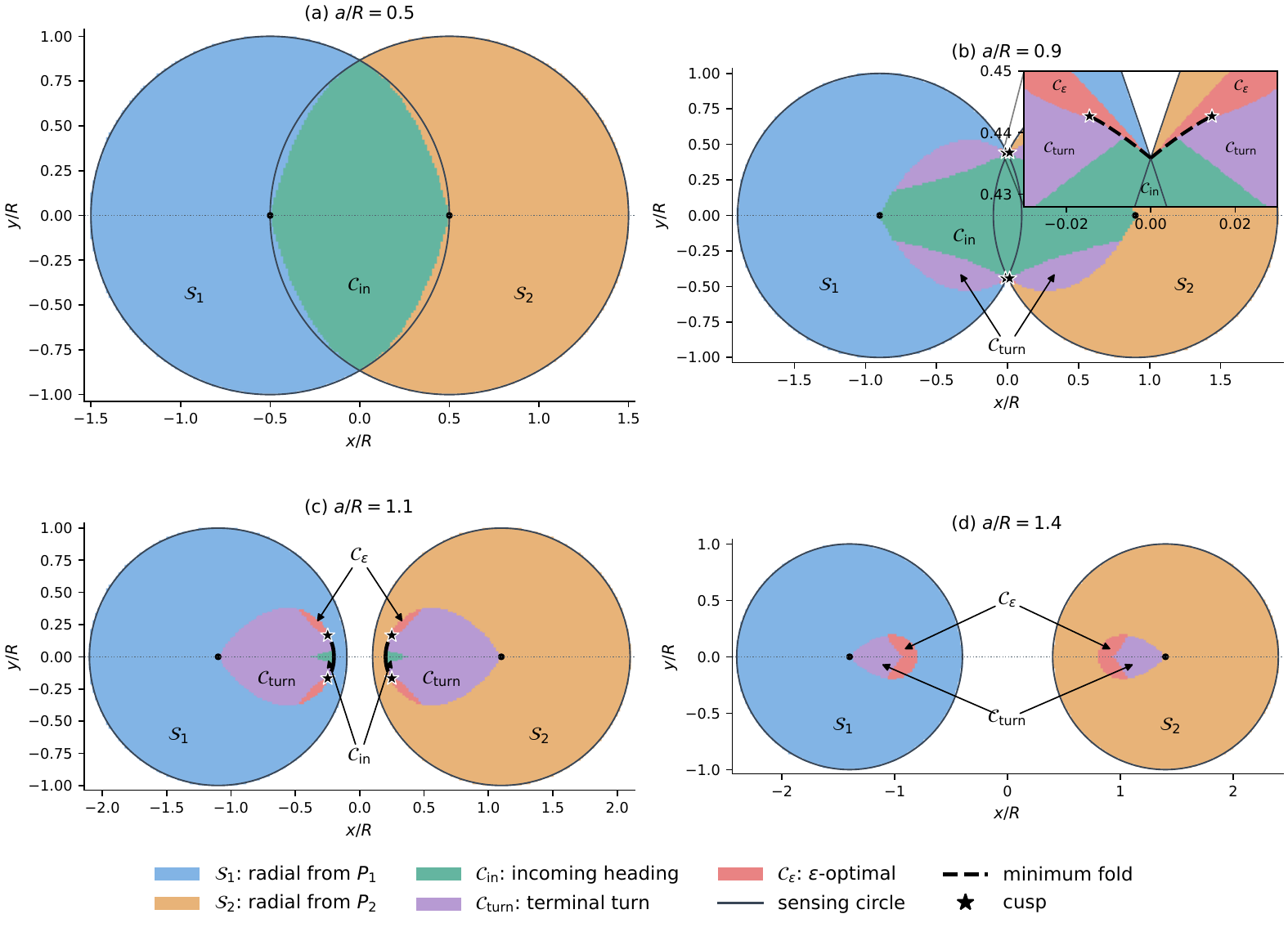}
\caption{The value partition on four separation slices, \(\mu=2\).
Colors and labels give the region of Table~\ref{tab:regions}; arrows
point to regions too small to hold their label, and each label names
both mirror-image copies. Dashed black curves are minimum folds, stars
mark cusps, and thin curves are the sensing circles. The inset in (b)
magnifies the fold and cusp near the upper double-boundary point.}
\label{fig:partition}
\end{figure}

Let \(a_{\rm one}=R+B=\mu R/(\mu-1)\). If \(a\ge a_{\rm one}\), every
interior state is single-active, because \(R+T_s\le R+B\le a\le r_m\)
whenever \(X<a\), and the half-plane \(X\ge a\) is single-active anyway.
Table~\ref{tab:thresholds} lists which structures occur in each range of
separations, and Figure~\ref{fig:partition} shows four representative
slices.

\begin{table}[htbp]\centering\small
\begin{tabular}{@{}p{.28\linewidth}p{.65\linewidth}@{}}
\toprule
Half-separation & Structure for interior states\\
\midrule
\(0<a\le a_{\rm tr}\) & Corner escapes keep the incoming heading.\\
\(a_{\rm tr}<a\le\kappa R\) & Terminal turns can occur;
Theorem~\ref{thm:value} applies.\\
\(\kappa R<a<R\) & Folds and cusps occur although the physical disks
overlap.\\
\(R\le a<a_*\) & Physical disks are tangent or disjoint. For \(R<a<a_*\)
the axial fold, at the positive time \(T_F=a-R\), separates early gap
escapes from later corner escapes; at \(a=R\) it collapses to the
double-boundary point \(x=y=0\), with \(T_F=0\).\\
\(a_*\le a<a_{\rm one}\) & No minimum fold; corner and
\(\varepsilon\)-optimal cells can still occur.\\
\(a\ge a_{\rm one}\) & All interior states are single-active, and
\(V_P=(R-d)/(\mu-1)\).\\
\bottomrule
\end{tabular}
\caption{Roles of the separation thresholds. At \(\mu=2\),
\(a_{\rm tr}/R=1/2\), \(\kappa=\sqrt3/2\), \(a_*/R=4/3\), and
\(a_{\rm one}/R=2\).}
\label{tab:thresholds}
\end{table}

\subsection{A pursuer response at the fold}
\label{sub:foldguard}
Root selection alone does not tell the pursuers how to defend a fold. A
pursuer that tracks only the later corner can let an evader deviation
cross into the early-escape side. The following response respects both
the fold and the later time front.

Consider a smooth minimum-fold point with \(X>0\), \(Y>0\), \(T_F>0\).
For \(j=F,L\), put
\[
 r_j=R+T_j,\quad h_j=\sqrt{r_j^2-a^2},\quad
 n_{1j}=(a,h_j)/r_j,\quad n_{2j}=(-a,h_j)/r_j,\quad
 e_j=(-X,h_j-Y)/(\mu T_j),
\]
all in the moving frame. Let \(\theta_j=\arctan(h_j/a)\) be the angle of
\(n_{1j}\), measured counterclockwise from the baseline direction; the
heading \(n_{2j}\) of the other pursuer is its mirror image in the
\(y\)-axis. Set \(\Delta_P=\theta_L-\theta_F\),
\(\Delta_E=\arccos(e_F\cdot e_L)\), and
\(c_j=\mu T_jh_j/[r_j(h_j-Y)]=1/D'(T_j)\), the corner quantity \(c\) of
\eqref{eq:cornerdefs} at horizon \(T_j\). At the fold \(c_F=1/\mu\), and the
double-root identities yield
\begin{equation}\label{eq:angleidentity}
 1-\cos\Delta_P=\frac{T_L}{R+T_L}
                   (1-\cos\Delta_E),\qquad
 0<\Delta_P<\Delta_E,
\end{equation}
for example by subtracting the two squared contact equations and using
\(Y=h_F[1-\mu^2T_F/r_F]\). Against the current evader heading \(e\), let
\(\delta\in[0,\pi]\) be the unsigned angle between \(e\) and \(e_F\), and
let the pursuers use the mirror-symmetric headings at angle \(\theta\):
\begin{equation}\label{eq:foldguard}
 \delta=\arccos(e_F\cdot e),\quad
 \theta=\theta_F+\min(\delta,\Delta_P),\qquad
 u_1=(\cos\theta,\sin\theta),\quad
 u_2=(-\cos\theta,\sin\theta).
\end{equation}
If \(g_F(z)=F_z(T_F(z))\) denotes the local minimum, then
\begin{equation}\label{eq:foldslack}
 \dot g_F=\mu\big[\cos(\min(\delta,\Delta_P))-\cos\delta\big]\ge0,
\end{equation}
so the response prevents first-order crossing into \(g_F<0\). For a
remaining horizon \(s\) with \(\dot s=-1\), the derivative of the
later-front constraint \(F_z(s)\) at \(s=T_L\) has the sign of
\begin{equation}\label{eq:laterslack}
 \cos\big(\Delta_P-\min(\delta,\Delta_P)\big)-1
       +\mu c_L(1-e_L\cdot e),
\end{equation}
which is also nonnegative. For \(\delta\ge\Delta_P\) the first two terms
cancel; otherwise the angle between \(e_L\) and \(e\) is at least
\(\Delta_E-\delta>\Delta_P-\delta\), and \(\mu c_L>1\).

At the axial fold \(Y=0\), \(R<a<a_*\), this response is combined with
the dispersal rule. Here \(X=\mu(a-R)\) and \(e_F=(-1,0)\). With
\(\theta_L=\arctan(h_L/a)\) and \(\delta=\arccos(-e_x)\), the unsigned
angle between \(e\) and \(e_F\), use
\begin{equation}\label{eq:axialfoldguard}
 u_1=(-e_x,e_y),\qquad u_2=(e_x,e_y)
\end{equation}
if \(\delta\le\theta_L\), and \eqref{eq:tie} with the later horizon
\(T_L\) otherwise. In the first case
\(\dd[X-\mu(a-R)]/\dd t=0\), and in the second it is positive; the
smaller upper/lower time-front derivative is nonnegative by
\eqref{eq:angleidentity} and by the dispersal inequality, respectively.
These local responses are ingredients of the global argument in the next
section; by themselves they do not yet constitute a strategy for the
whole game.

\section{The value for every initial state}
\label{sec:global}
The upper bound \(V_P\le U\) holds everywhere by
Proposition~\ref{prop:bounds}, and Theorem~\ref{thm:value} proves the
lower bound for \(a_0\le a_{\rm c}\). That theorem comes first because
it gives the strategies in closed form and because its symmetry-line
estimate is used again below. This section proves the lower bound at
every state by an existence argument that does not use the conclusion
of Theorem~\ref{thm:value}. The difficulty is that a
single pursuer strategy must keep the evader detected until time \(U\)
against every evader control, including near folds, cusps, and corners
that cannot be completed. The proof has three steps.
\begin{enumerate}
\item \emph{A viability formulation.} Attach a clock \(s\), the
remaining time budget, which decreases at unit rate. The pursuers
succeed if \((z,s)\) stays in a closed set \(\mathcal K\) of states that
are still detected and whose budget does not exceed their own escape
time.
\item \emph{A convex auxiliary game.} Restrict the pursuers to
mirror-symmetric motions with speed at most one. Their velocity sets are
then convex, and the discriminating-domain theorem
of~\cite{cardaliaguet96} reduces
the problem to a pointwise check: for every evader heading, some pursuer
velocity is tangent to \(\mathcal K\).
\item \emph{Return to unit speed.} Moving the pursuers inward faster than
the auxiliary strategy prescribes only helps them, and a single late
handoff to pure pursuit settles the remaining case.
\end{enumerate}
Restricting the pursuers in the second step is harmless for a lower
bound, since any guarantee achieved by the restricted pursuers is also
achieved by the unrestricted team.

\begin{theorem}[Global value]\label{thm:globalvalue}
For every \(R>0\), \(\mu>1\), and initial configuration, the value of
the fixed-speed game under pursuer-nonanticipative strategies is
\begin{equation}\label{eq:globalvalue}
 V_P(z)=U(z)=\inf\{T\ge0:D_z(T)<\mu T\}.
\end{equation}
The algorithm of Section~\ref{sub:algorithm} computes it, including at
folds, cusps, and nonusable double-boundary states. The pursuers have a
nonanticipative strategy guaranteeing \(t_f\ge U(z)\). The evader has
\(\varepsilon\)-optimal constant-heading policies. In
\(\mathcal S_i\), \(\mathcal C_{\rm in}\), and \(\mathcal C_{\rm turn}\),
which by Table~\ref{tab:regions} include the single/corner mergers in the
last two, the policies of Section~\ref{sec:families} attain \(U(z)\)
exactly.
\end{theorem}

\subsection{A fixed frame and a time-budget set}
Keep the initial baseline direction fixed and restrict the pursuers to
\begin{equation}\label{eq:symmetricball}
 u_1=(C,V),\quad u_2=(-C,V),\qquad C^2+V^2\le1.
\end{equation}
The midpoint then has no horizontal motion, and
\begin{equation}\label{eq:auxdynamics}
 \dot a=-C,\qquad \dot x=\mu e_x,\qquad
 \dot y=\mu e_y-V,\qquad \dot s=-1.
\end{equation}
Here \(a\) may be negative, with the disks centered at \((\pm a,0)\),
which removes the coordinate singularity at coincident pursuers. The
dynamics are bounded and globally Lipschitz, and for each evader heading
the set of pursuer-induced velocities is convex and compact; the evader
control set remains the unit circle.

Let \(\Sigma\) be the closed sensing union in these coordinates and
define the time-budget set
\begin{equation}\label{eq:budgetset}
 \mathcal K=\left\{(z,s): z\in\Sigma,\quad
 0\le s\le B,\quad F_z(\tau)\ge0\ \text{for all }0\le\tau\le s\right\}.
\end{equation}
Preserving only \(F_z(s)\ge0\) would not suffice, since an earlier
negative interval would already allow escape. To use an
infinite-horizon viability statement for a finite budget, set
\begin{equation}\label{eq:extendedbudget}
 \widehat{\mathcal K}=\{(z,s):s\le0\}\cup\mathcal K,
\end{equation}
which is closed because \(F\) is continuous. For \(s>0\), membership is
equivalent to \(s\le U(z)\), and reaching \(s=0\) discharges the
detection requirement.

\subsection{Tangency, including the nonsmooth junctions}
\begin{lemma}[Auxiliary discrimination]\label{lem:auxviability}
For every \((z,s)\in\widehat{\mathcal K}\) and every evader heading
\(e\), some \((C,V)\) in the closed unit disk makes the velocity in
\eqref{eq:auxdynamics} belong to the contingent cone of
\(\widehat{\mathcal K}\). Consequently the auxiliary pursuers have a
nonanticipative strategy that keeps the state in
\(\widehat{\mathcal K}\).
\end{lemma}

The proof is given in Appendix~\ref{app:auxviability}. It has three
parts.
\begin{enumerate}
\item \emph{Only the active horizons matter.} At a point \((z,s)\) of
\(\mathcal K\) with \(s>0\), membership requires \(F_z(\tau)\ge0\) for all
\(\tau\in[0,s]\). The constraint is felt only at the \emph{active
horizons}, where \(F_z(\tau)=0\). By the shape of \(F_z\)
(Figure~\ref{fig:residual} and Section~\ref{sub:algorithm}) the positive
active horizons are at most two: the endpoint \(\tau=s\), an interior
minimum that touches zero, or both. Moving the pursuers together strictly increases
\(F_z\) at every active horizon, so an arbitrarily small inward correction
turns a first-order condition into membership. It therefore suffices to
find a pursuer velocity for which \(F_z\) does not decrease, to first
order, at each active horizon.
\item \emph{A control for each case.} Table~\ref{tab:auxcases} lists the
velocity that achieves this. In the two basic cases the derivative of the
active constraint is \(\mu(1-e_*\cdot e)\ge0\), where \(e_*\) is the evader
heading toward the corner in question. The tie on \(y=0\) reduces to the
dispersal inequality \eqref{eq:tieslack}, and a fold with both
constraints active to \eqref{eq:foldslack} and \eqref{eq:laterslack}. No
case divides by \(\mu c-1\), so cusps are included.
\item \emph{From tangency to a strategy.} A closed set with this
tangency property is a discriminating domain, and the theorem of
Cardaliaguet~\cite{cardaliaguet96} then provides a nonanticipative
strategy that keeps the state in it. This step needs convex pursuer
velocity sets, which is why the auxiliary pursuers may move slower than
unit speed.
\end{enumerate}

\begin{table}[htbp]\centering\small
\begin{tabular}{@{}>{\raggedright\arraybackslash}p{.33\linewidth}>{\raggedright\arraybackslash}p{.59\linewidth}@{}}
\toprule
Active constraint & Auxiliary pursuer velocity \((C,V)\)\\
\midrule
Endpoint \(\tau=s\) at a corner
 & Head for the corner at horizon \(s\): \((a,h)/(R+s)\).\\
Interior minimum at \(T_F\) (fold)
 & Head for the corner at horizon \(T_F\); if the endpoint is also active,
   the fold response \eqref{eq:foldguard}.\\
Upper/lower tie, \(y=0\)
 & The dispersal response:
   \(V=\operatorname{clip}(\mu e_y,-h/r,h/r)\), \(C=\sqrt{1-V^2}\).\\
Axial fold or axial cusp
 & The axial response \eqref{eq:axialfoldguard} with the tie response;
   \((C,V)=(1,0)\) when the budget equals the kink time.\\
Nonusable double-boundary point
 & The corner control at horizon \(s\), or the tie response if the point
   lies on \(y=0\). The distance is not differentiable there, and
   horizons near zero are treated directly.\\
Outside the strip \(|x|<a\), or a single-disk budget suffices
 & Pure pursuit by the corresponding pursuer; the other mirrors its
   horizontal velocity.\\
\bottomrule
\end{tabular}
\caption{The cases in the proof of Lemma~\ref{lem:auxviability}
(Appendix~\ref{app:auxviability}). Here \(h=\sqrt{(R+s)^2-a^2}\) and
\(r=R+s\).}
\label{tab:auxcases}
\end{table}

\subsection{Returning to unit speed}
The auxiliary strategy may ask the pursuers to move slower than unit
speed, for example to hold their horizontal positions while matching the
evader vertically. The remedy is a monotonicity: for an evader between
the pursuers, bringing the two sensing disks closer together only makes
escape harder. The actual pursuers therefore copy the auxiliary vertical
motion exactly and spend all remaining speed moving inward.

\begin{lemma}[Inward saturation]\label{lem:saturation}
Let an auxiliary trajectory start with \(a_0>|x_0|\). Its bounded-speed
pursuer controls can be converted causally into unit-speed controls that
preserve its time-budget guarantee, with at most one handoff to pure
pursuit.
\end{lemma}
\begin{proof}
Run the auxiliary strategy virtually, with controls \((C(t),V(t))\), and
before the handoff use the actual controls
\begin{equation}\label{eq:saturation}
 \widehat u_1=(\sqrt{1-V^2},V),\qquad
 \widehat u_2=(-\sqrt{1-V^2},V).
\end{equation}
The virtual and actual midpoint motions coincide, so the coordinates
\(x,y\) coincide, and the half-separations satisfy
\[
 \widehat a(t)=a_0-\int_0^t\sqrt{1-V(q)^2}\,\dd q
 \ \le\ a_0-\int_0^t C(q)\,\dd q=a(t).
\]
For \(|x|\le\widehat a\le a\),
\begin{equation}\label{eq:monotonecoverage}
 D_{(\widehat a,x,y)}(\tau)\ge D_{(a,x,y)}(\tau)
 \qquad(\tau\ge0),
\end{equation}
because \(D_a=-(a-|x|)/d\le0\) on the exposed piece and
\(D_a=-a(h-|y|)/(hD)\le0\) on the corner piece, and the pieces match
continuously. The nearest range also decreases, so the actual state
stays in the sensing union, and \((\widehat z(t),s_0-t)\in
\widehat{\mathcal K}\) while \(\widehat a>|x|\).

At the first time \(t_h\) with \(\widehat a=|x|\), the nearer pursuer is
vertically aligned with the evader, so
\[
 D_{\widehat z}(\tau)=R+\tau-|y|,\qquad
 \frac{R-|y|}{\mu-1}\ge s_0-t_h,
\]
and pure pursuit by that pursuer maintains detection for the remaining
budget. If the initial state is outside the strip, the handoff occurs at
time zero. The virtual strategy, the saturation, and the handoff time
depend only on the evader's heading history up to the present, so the
composition is nonanticipative.
\end{proof}

\begin{proof}[Proof of Theorem~\ref{thm:globalvalue}]
Proposition~\ref{prop:bounds} gives \(V_P\le U\). Start the auxiliary
system with budget \(s_0=U(z_0)\); by continuity of \(F\) and the
definition of \(U\), \((z_0,s_0)\in\widehat{\mathcal K}\), including
when an earlier isolated zero is skipped.
Lemma~\ref{lem:auxviability} and Lemma~\ref{lem:saturation} give a
unit-speed nonanticipative pursuer strategy that keeps the evader in the
sensing union for all \(t<s_0\). Hence \(t_f\ge U(z_0)\) for every
evader control, and \(V_P\ge U\). The statements about evader policies
follow from Propositions~\ref{prop:bounds}, \ref{prop:single},
and~\ref{prop:completion}.
\end{proof}

\subsection{The reverse ordering}
\label{sub:reverse}
The evader policies of Theorem~\ref{thm:globalvalue} are open-loop, so they
bound the value \(V_E\) of the reverse ordering as well. The pursuer
strategy, however, may react to the evader's current heading. We now
show that reversing the order does not change the value at any state.
The statement and the proof use the \emph{earliest zero} of the contact
residual,
\begin{equation}\label{eq:earliestzero}
 U_0(z)=\inf\{T\ge0:F_z(T)\le0\}
\end{equation}
for states in the closed sensing union, and \(U_0(z)=0\) for states
strictly outside both disks. Clearly \(U_0\le U\), and \(U_0(z)<U(z)\)
exactly when \(F_z\) has a zero before \(U(z)\) at which it does not become
negative.

\begin{theorem}[Reverse ordering]\label{thm:reverse}
For every \(R>0\), \(\mu>1\), and initial state,
\begin{equation}\label{eq:reversebounds}
 V_E(z)=V_P(z)=U(z).
\end{equation}
More precisely, for every nonanticipative evader strategy \(\mathcal A\)
and every \(T<U(z)\) there is a pursuer control \(u\in\mathcal U\) with
\(J_z(\mathcal A[u],u)\ge T\). The earliest zero satisfies \(U_0(z)<U(z)\)
only on the minimum-fold sheets of Section~\ref{sub:fold}, including the
axial fold but not the cusp curve or the maximum-fold portion, and at the
nonusable double-boundary points of Section~\ref{sub:regions}.
\end{theorem}

\paragraph{Outline of the proof}
The upper bound \(V_E\le U\) is immediate from the open-loop evader
policies. The lower bound is proved in three stages.
\begin{enumerate}
\item \emph{Steps (a)--(d): \(V_E\ge U_0\) everywhere.} Against a
nonanticipative evader strategy, the viability theory gives the pursuers
a control that keeps the state within any prescribed distance of a
discriminating domain, but not inside it. We therefore apply it to the
game with a slightly smaller sensing radius \(R-\rho\), so that a small
error still leaves the evader inside the true disks, return to unit
speed by switching the pursuers' horizontal direction rapidly, and let
\(\rho\to0\). Shrinking the radius lowers the contact residual, which
turns a tangential zero into an escape interval; this is why the limit
is the earliest zero \(U_0\) and not \(U\).
\item \emph{Step (e): nonusable double-boundary points.} Here \(U_0=0<U\).
The pursuers open by heading straight for the evader's initial position
for a short time \(t_1\), a control fixed in advance. A scaling argument
shows that afterwards the residual is positive up to a horizon close to
\(U\), whatever the evader did, and steps~(a)--(d) apply from the new
state.
\item \emph{Step (f): minimum-fold states.} Here \(0<U_0=T_F<U=T_L\). The
pursuers open by heading for the contact corner
(Lemma~\ref{lem:foldopening}). If the evader ever leaves the heading
toward that corner, the residual becomes strictly positive up to a
horizon close to \(T_L\), and steps~(a)--(d) apply. If it never does, it
arrives at a nonusable double-boundary point, and step~(e) applies.
\end{enumerate}
In stages~2 and~3 the opening costs an amount proportional to its
length, which can be taken arbitrarily small.

\subsection{Opening at a minimum fold}
\label{sub:foldopening}
Step~(f) needs a preparatory estimate. In the notation of
Section~\ref{sub:fold}, a minimum-fold state with \(x,y\ge0\), contact time
\(T_F\), \(r_F=R+T_F\), and \(k=h_F/r_F\) is
\begin{equation}\label{eq:foldkT}
 a=r_F\sqrt{1-k^2},\qquad x=\mu T_F\sqrt{c_k},\qquad y=kA,\qquad
 c_k=1-\mu^2k^2,\quad A=R-mT_F,
\end{equation}
with \(m=\mu^2-1\); this is \eqref{eq:foldparam} written in \((k,T_F)\).
The range \eqref{eq:foldrange}, together with the axial fold \(k=0\), is
\begin{equation}\label{eq:foldkTrange}
 0\le k<\frac1\mu,\qquad 0<T_F<\frac{Rc_k}{m},
 \qquad\text{equivalently}\qquad
 \Delta:=R-\frac{mT_F}{c_k}>0,
\end{equation}
because \(mr_F^3<\mu^2Ra^2\) reads \(mr_F<\mu^2R(1-k^2)\). The quantity
\(\Delta\) measures how far the state is from the cusp, where
\(\Delta=0\). The heading from
the evader to the contact corner \(Q_F=(0,h_F)\) is
\(e_F=(-\sqrt{c_k},\mu k)\), and the headings from the pursuers to that
corner are
\(n_{1F}=(\sqrt{1-k^2},k)\) and \(n_{2F}=(-\sqrt{1-k^2},k)\). On the fold,
\(F_z\ge0\) on \([0,T_L]\), with zeros exactly at \(T_F\) and \(T_L\): the
exposed piece is strictly decreasing, and the corner piece is first
convex and then concave, so it has at most one interior minimum
(Section~\ref{sub:algorithm}); for \(k=0\) the zero at \(T_F\) is the join
kink. At horizon \(T_F\) the corner \(Q_F\) is the
unique point of the exterior of the expanded union closest to the evader
(Section~\ref{sub:distance}); for \(k=0\) it is the point at which the two
expanded circles touch.

\begin{lemma}[Opening at a minimum fold]\label{lem:foldopening}
Let \(z_0\) be a minimum-fold state \eqref{eq:foldkT}--\eqref{eq:foldkTrange},
and let the pursuers hold the constant headings \(n_{1F}\), \(n_{2F}\) on
\([0,t_1]\), where \(0<t_1<\min\{T_F,\Delta\}\). Put
\begin{equation}\label{eq:shiftedlaterroot}
 \widetilde T_L(t_1)=T_F+\frac2m\Big[A-t_1-\mu k\sqrt{(R-t_1)(A-t_1)}\Big],
 \qquad T_L-\frac{2t_1}{m}\ \le\ \widetilde T_L(t_1)\ \le\ T_L=\widetilde T_L(0).
\end{equation}
Then, for every evader control:
\begin{enumerate}
\item[\textup{(i)}] the evader is strictly inside the sensing union on
\([0,t_1]\);
\item[\textup{(ii)}] \(F_{z(t_1)}(\tau)\ge0\) for \(0\le\tau\le \widetilde T_L(t_1)-t_1\);
\item[\textup{(iii)}] the inequality in \textup{(ii)} is strict for
\(0\le\tau<\widetilde T_L(t_1)-t_1\), unless \(e(t)=e_F\) for almost every
\(t\in[0,t_1]\). In that case \(z(t_1)\) is again a minimum-fold state,
with the same \(k\), contact time \(T_F-t_1\), a larger \(\Delta\), and a
later root \(T_L'\) with \(t_1+T_L'\ge T_L\).
\end{enumerate}
\end{lemma}
In words, \(\widetilde T_L(t_1)\) is the time, counted from the start, up to
which the contact residual is guaranteed to stay nonnegative after the
opening, and by \eqref{eq:shiftedlaterroot} the opening loses at most
\(2t_1/m\) of the escape time \(T_L\). If the evader heads for the contact
corner throughout, there is no loss, and the state stays on the fold.
The proof, in Appendix~\ref{app:reverse}, is a comparison. Heading for
the contact corner brings the pursuers to the positions of a reference
fold state, with the same \(k\) and \(T_F\), of the game with the smaller
sensing radius \(R-t_1\). Seen from the state after the opening, the
expanded disks at horizon \(\tau\) are those of the reference state at
horizon \(\tau+t_1\), since \(R+\tau=(R-t_1)+(\tau+t_1)\). The evader is within \(\mu t_1\) of the reference evader
position, and distance to a closed set is \(1\)-Lipschitz, so the contact
residual after the opening is at least the residual of the reference
state at the shifted horizon, which is nonnegative up to its later root
\(\widetilde T_L(t_1)\). Equality at the contact horizon forces the evader to
have moved exactly toward the contact corner throughout, which is
case~(iii).

The proof of Theorem~\ref{thm:reverse} follows the outline above and is
given in Appendix~\ref{app:reverse}. The next remark explains why the
fold states need the separate opening of step~(f).

\begin{remark}[Why the fold needs its own step]\label{rem:folds}
Steps~(a)--(d) alone give only \(U_0\) at a fold: by
\eqref{eq:radiusshift} the smaller radius lowers the contact residual by
\(\mu\rho\) near the tangency and turns it into a strict escape interval.
Nor can a mirror-symmetric pursuer velocity chosen before the evader's
keep both active constraints from decreasing to first order. On a minimum fold with full
budget \(s=U(z)=T_L\), both the minimum at \(T_F\) and the endpoint at
\(T_L\) are active. For mirror-symmetric unit-speed pursuer headings at
angle \(\theta\), \eqref{eq:foldslack} and \eqref{eq:laterslack} show that the
two linearized derivatives are positive multiples of
\[
 \cos(\theta-\theta_F)-e_F\cdot e
 \qquad\text{and}\qquad
 \cos(\theta_L-\theta)-1+\mu c_L(1-e_L\cdot e).
\]
If \(\theta\ne\theta_F\), the evader heading \(e=e_F\) makes the first
negative; if \(\theta=\theta_F\), the heading \(e=e_L\) makes the second
equal to \(\cos\Delta_P-1<0\). Step~(f) accepts this first-order loss but
confines it to one short interval. The heading \(\theta=\theta_F\), which is
the opening control \(n_{iF}\), protects the minimum exactly; against it
the evader loses nothing only by following \(e_F\), which keeps it on the
fold and leads to the double-boundary corner; and as soon as it has left
\(e_F\), the contact residual is strictly positive and steps~(a)--(d)
apply. Lemma~\ref{lem:foldopening} bounds the loss over an opening of
length \(t_1\) by \(2t_1/(\mu^2-1)\). The pursuer control built in
step~(f) therefore depends on the evader strategy, through the time at
which its response to the opening first leaves \(e_F\). This is what the
definition of \(V_E\) allows. The proof does not provide a pursuer rule
with a fixed reaction delay that secures \(U\) against all evader
strategies, and in our simulations heading-blind sample-and-hold
pursuers indeed lose, at the two exact fold states tested, an amount
that roughly halves when their time step is halved
(Table~\ref{tab:heuristics}).
\end{remark}

\begin{remark}[Saddle points, Nash equilibria, and information]\label{rem:saddle}
The game is zero-sum between the evader and the pursuer team, so a Nash
equilibrium is a saddle point: a pair of strategies from which neither
side can gain by deviating alone. In these terms the results read as
follows.
\begin{enumerate}
\item \emph{The game has a value.} By Theorems~\ref{thm:globalvalue}
and~\ref{thm:reverse}, the two values in nonanticipative strategies
coincide, \(V_E=V_P=U\), at every state. The pursuer strategy of
Theorem~\ref{thm:globalvalue} comes from an existence theorem for
nonanticipative strategies and may use the evader's current heading,
which is why we state the main value results for \(V_P\).
\item \emph{An exact saddle point where the evader attains \(U\).} In the
regions \(\mathcal S_i\), \(\mathcal C_{\rm in}\), and
\(\mathcal C_{\rm turn}\), the evader has an open-loop policy \(e^*\) that
escapes by \(U(z)\) against every pursuer motion, and the pursuers have a
nonanticipative strategy \(\mathcal B^*\) that keeps detection until
\(U(z)\) against every evader control. Hence, for every evader control
\(e\) and every nonanticipative pursuer strategy \(\mathcal B\),
\[
 J_z(e^*,\mathcal B[e^*])\ \le\ J_z(e^*,\mathcal B^*[e^*])=U(z)
 \ \le\ J_z(e,\mathcal B^*[e]) ,
\]
so \((e^*,\mathcal B^*)\) is a saddle point of the game in which the evader
chooses a control and the pursuers a nonanticipative strategy. For
the separations of Theorems~\ref{thm:value} and~\ref{thm:extension}
both are explicit.
\item \emph{Approximate saddle points otherwise.} In
\(\mathcal C_\varepsilon\) the evader's policies are
\(\varepsilon\)-optimal, which gives an \(\varepsilon\)-saddle point for
every \(\varepsilon>0\); whether an exact one exists is the attainment
question of Section~\ref{sec:conclusion}.
\item \emph{The reverse ordering.} The saddle statements above concern
the game in which the evader chooses a control \(e\) and the pursuers a
nonanticipative strategy \(\mathcal B\). Under the reverse ordering,
Theorem~\ref{thm:reverse} supplies, for every fixed evader strategy
\(\mathcal A\) and every \(\varepsilon>0\), a pursuer control
\(u_{\mathcal A,\varepsilon}\) with
\(J_z(\mathcal A[u_{\mathcal A,\varepsilon}],u_{\mathcal A,\varepsilon})
\ge U(z)-\varepsilon\). This is what the definition of \(V_E\) requires.
It is a response to \(\mathcal A\): it provides neither a single pursuer
policy, independent of \(\mathcal A\), that ignores the evader's current
heading, nor a fixed saddle pair for the game in which the evader
chooses \(\mathcal A\) and the pursuers a control \(u\). Such a pair need
not exist. Take \(R=1\),
\(\mu=2\), and let both pursuers and the evader start at the same point,
so that \(U=1\). Against any fixed pursuer control and any \(T>1/2\), some
constant evader heading escapes by time \(T\): the evader can reach every
point of a circle of radius \(2T>1\), and each sensing disk covers less
than half of that circle. Hence
\(\sup_u\inf_{\mathcal A}J_z(\mathcal A[u],u)\le1/2\), while
\(\inf_{\mathcal A}\sup_uJ_z(\mathcal A[u],u)=V_E=1\).
\item \emph{What is not claimed.} We do not construct an equilibrium in
which both sides use state feedback. The value jumps across the folds and
the pursuer feedback \eqref{eq:tie} is discontinuous on \(y=0\), so the
classical verification of a feedback saddle point from a smooth solution
of the Isaacs equation is not available. Positional strategies in the
sense of Krasovskii and Subbotin~\cite{krasovskii}, that is, state
feedback applied over short time steps without knowledge of the
opponent's current control, are the natural setting for such a
statement. A positional pursuer strategy would also give a policy for
the reverse ordering that does not depend on \(\mathcal A\). Establishing
a positional guarantee would require an additional construction, which
we do not attempt; it is the first open problem of
Section~\ref{sec:conclusion}.
\end{enumerate}
\end{remark}

\subsection{Interpretation}
Tables~\ref{tab:regions} and~\ref{tab:strata} give the value and the
evader policies, exact or \(\varepsilon\)-optimal, on each part of the
state space. For example, the state
\((a,x,y)=(14,10,1)\) with \(R=10\), \(\mu=2\) has value
\(6.07420\) although its limiting corner has \(\Gamma<0\). At
\(R=1\), \(\mu=2\), \(a=1.1\), \(y=0\), \(x=0.2\), the value is
\(17/30\), not the isolated contact time \(0.1\), so the jump in
Figure~\ref{fig:valuejump} is a jump of \(V_P\). On the double boundary
with \(a=0.9\), \(y=\sqrt{0.19}\), the value is
\(2(1-2\sqrt{0.19})/3\approx0.085480\), positive although the evader
starts on the boundary of both disks.

The global pursuer strategy is defined through the auxiliary viability
problem and is not given in closed form. Where
Theorems~\ref{thm:value} and~\ref{thm:extension} apply, their explicit
feedback is available instead: head toward the current escape corner,
use the dispersal rule on \(y=0\), and hand off to pure pursuit when a
single pursuer becomes decisive. Near folds, the response
\eqref{eq:foldguard} shows what the global strategy must do.

\section{Examples and numerical validation}
\label{sec:examples}

\subsection{Benchmark configurations}
As a benchmark geometry we use \(R=10\), \(\mu=2\), and
\(P_1^0=(-5,0)\), \(P_2^0=(5,0)\). The three evader positions in
Table~\ref{tab:benchmarks} illustrate the three regular situations. The
evader at \((7,5)\) is close to \(P_2\) and far from \(P_1\), so it runs
radially away from \(P_2\). The centered evader at \((0,0)\) faces two
equally strong constraints and exits through a corner with
\(\beta=1/2\); the lower corner is equally good. The evader at \((2,2)\)
is the running example of Sections~\ref{sub:distance}
and~\ref{sub:matching}; it exits through a corner, and \(P_2\), the
closer pursuer, carries the larger weight \(1-\beta\approx0.727\). All three satisfy \eqref{eq:safe}, with
\(a_{\rm c}=5\sqrt3\approx8.66\), so Theorem~\ref{thm:value} applies.
They lie on the transversality threshold \(a_{\rm tr}=5\), but
\eqref{eq:worstrate} is positive for \(T>0\), so no terminal turn is
needed.

\begin{table}[htbp]
\centering\small
\begin{tabular}{@{}llll@{}}
\toprule
Initial evader & Optimal family & Escape time & \(\beta\)\\
\midrule
\((7,5)\) & Radial escape from \(P_2\) & \(4.61484\) & \(0\)\\
\((0,0)\) & Symmetric corner & \(9.34259\) & \(0.5\)\\
\((2,2)\) & Asymmetric corner & \(7.40194\) & \(0.272730\)\\
\bottomrule
\end{tabular}
\caption{Escape times for the benchmark configurations (normalized time).
For \((0,0)\) the upper escape is shown; its reflection is equally
optimal.}
\label{tab:benchmarks}
\end{table}

For the first configuration,
\[
 T=10-\sqrt{29},\qquad e=u_2=(2,5)/\sqrt{29},\qquad
 Q\approx(10.4278,\ 13.5695),
\]
and the inactive pursuer, drawn with heading
\(u_1\approx(0.750881,0.660437)\), ends at distance about
\(15.93>R\). For the centered configuration the scalar equation reduces
to \(4T^2=(10+T)^2-25\), so \(T=(10+\sqrt{325})/3\), with \(e=(0,1)\),
\(u_{1,2}\approx(\pm0.258497,\ 0.966012)\), and terminal reduced state
\((a_f,x_f,y_f)\approx(2.58497,\ 0,\ 9.66012)\).

\begin{figure}[!htb]\centering
\includegraphics[width=0.94\linewidth]{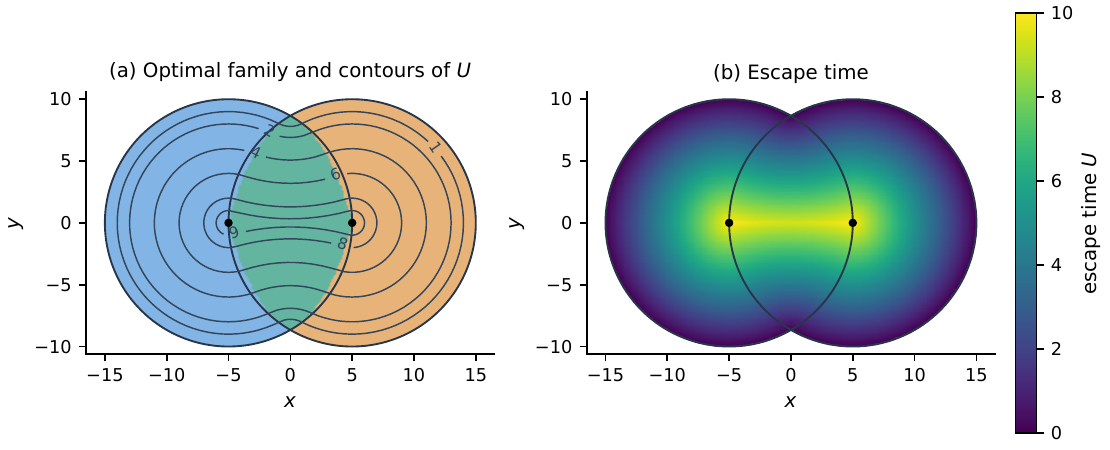}
\caption{Escape time on the slice \(a_0=5\), \(R=10\), \(\mu=2\). Left:
optimal family (blue: radial from \(P_1\); orange: radial from \(P_2\);
green: corner) with contours of the escape time. Right: the escape time.
Each point is an independently solved initial state.}
\label{fig:slice}
\end{figure}

\begin{figure}[!tbp]\centering
\includegraphics[width=\linewidth]{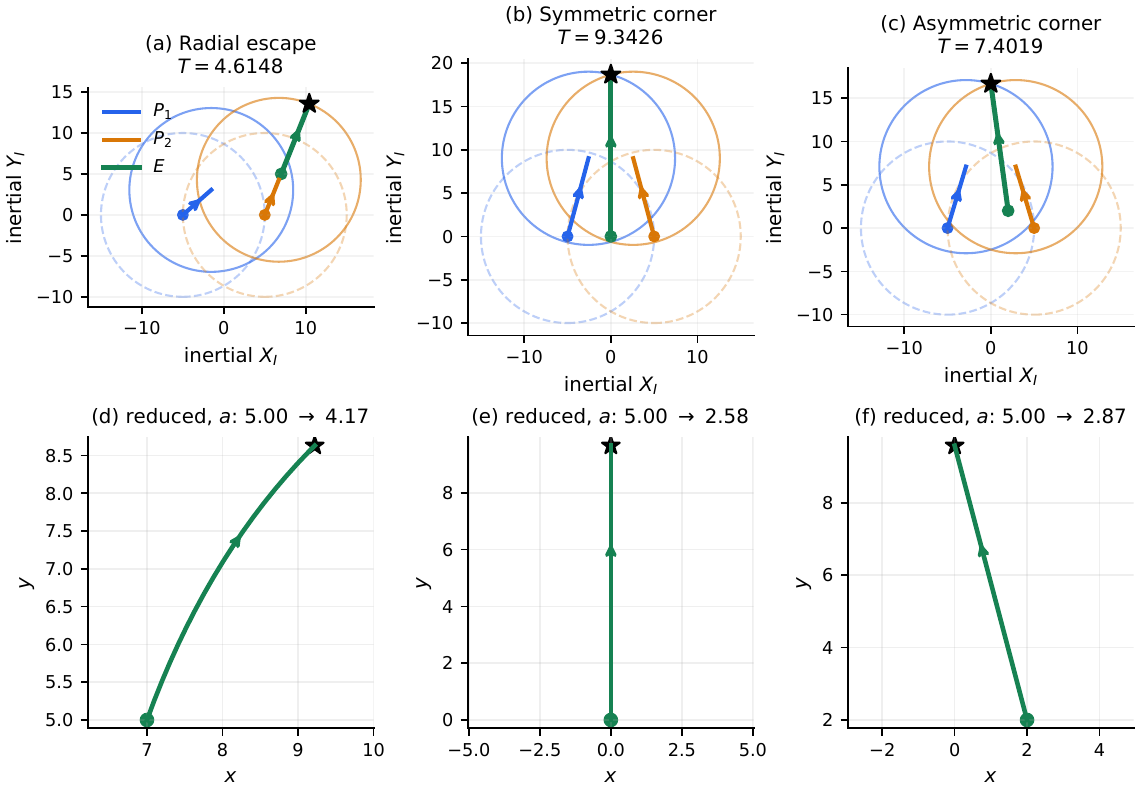}
\caption{Optimal plays for the benchmarks of Table~\ref{tab:benchmarks}.
Top: inertial trajectories with initial (dashed) and terminal (solid)
sensing disks. Bottom: projections onto the reduced \((x,y)\) plane;
the half-separation \(a\) changes along each play. Arrows point forward
in time. In the single-active example the inactive pursuer's motion
rotates the frame, which curves the reduced projection although all
physical paths are straight.}
\label{fig:trajectories}
\end{figure}

\begin{figure}[p]\centering
\includegraphics[width=0.92\linewidth]{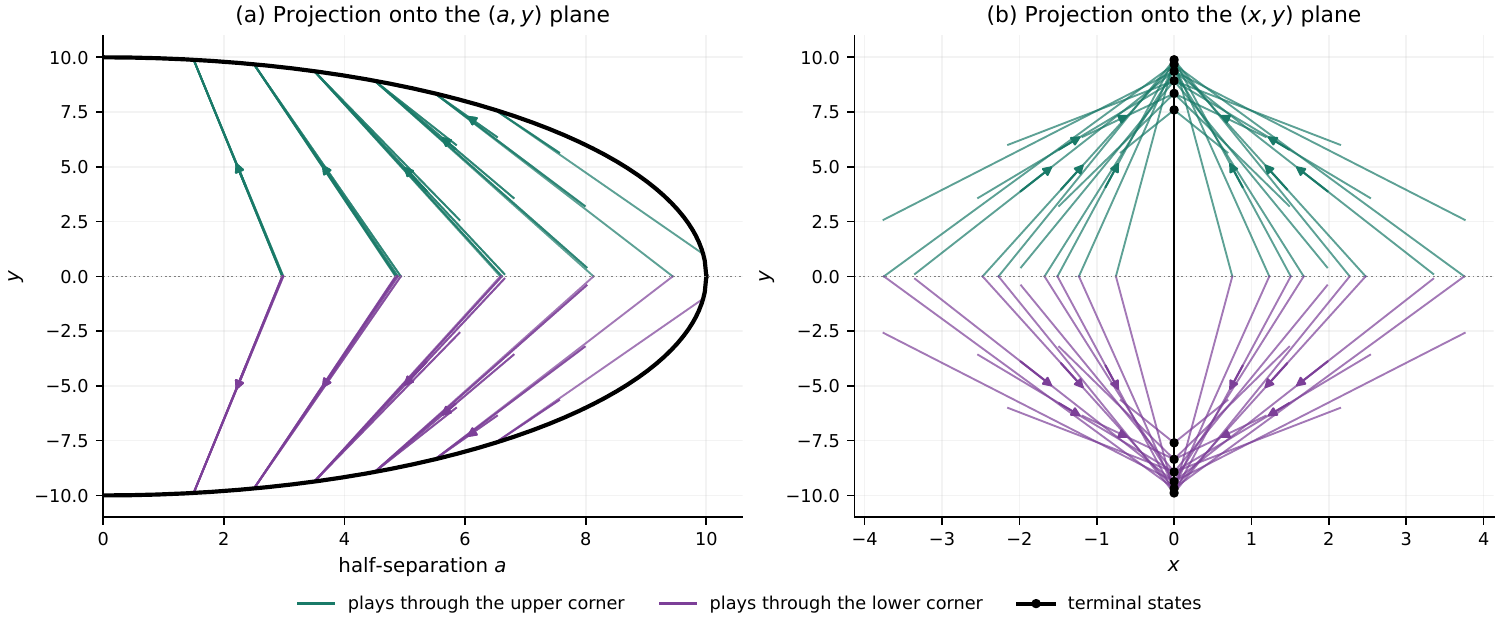}
\caption{The two-active families \eqref{eq:explicitfan} in the reduced
state space \((a,x,y)\), shown in two projections, for \(R=10\), \(\mu=2\),
terminal half-separations \(a_f\in\{1.5,2.5,\dots,6.5\}\), and weights
\(\beta\in[0.12,0.88]\). Arrows indicate forward time. Each line is one
play, generated backward from its
terminal state \((a_f,0,\pm\sqrt{R^2-a_f^2})\) on the black curve; the
plays are straight lines in the reduced state space, so both projections
are straight. (a) The half-separation grows backward in time and \(|y|\)
decreases. (b) Plays with different weights \(\beta\) leave the same
terminal state in different directions. Only the portions inside both
sensing disks and before the symmetry line \(y=0\) are shown, that is,
up to \(\min\{\tau_{\rm range},\tau_{\rm sym}\}\) of
Section~\ref{sub:completion}.}
\label{fig:fans}
\end{figure}

Figure~\ref{fig:slice} shows the escape time and the optimal family
over the whole slice of initial states with this pursuer separation,
which contains the three benchmarks. Figure~\ref{fig:trajectories}
shows the three optimal plays in inertial and reduced coordinates, and
Figure~\ref{fig:fans} the two-active families \eqref{eq:explicitfan}.

\newpage
\subsection{Numerical validation}
\label{sub:validation}
\begin{figure}[!t]\centering
\includegraphics[width=\linewidth]{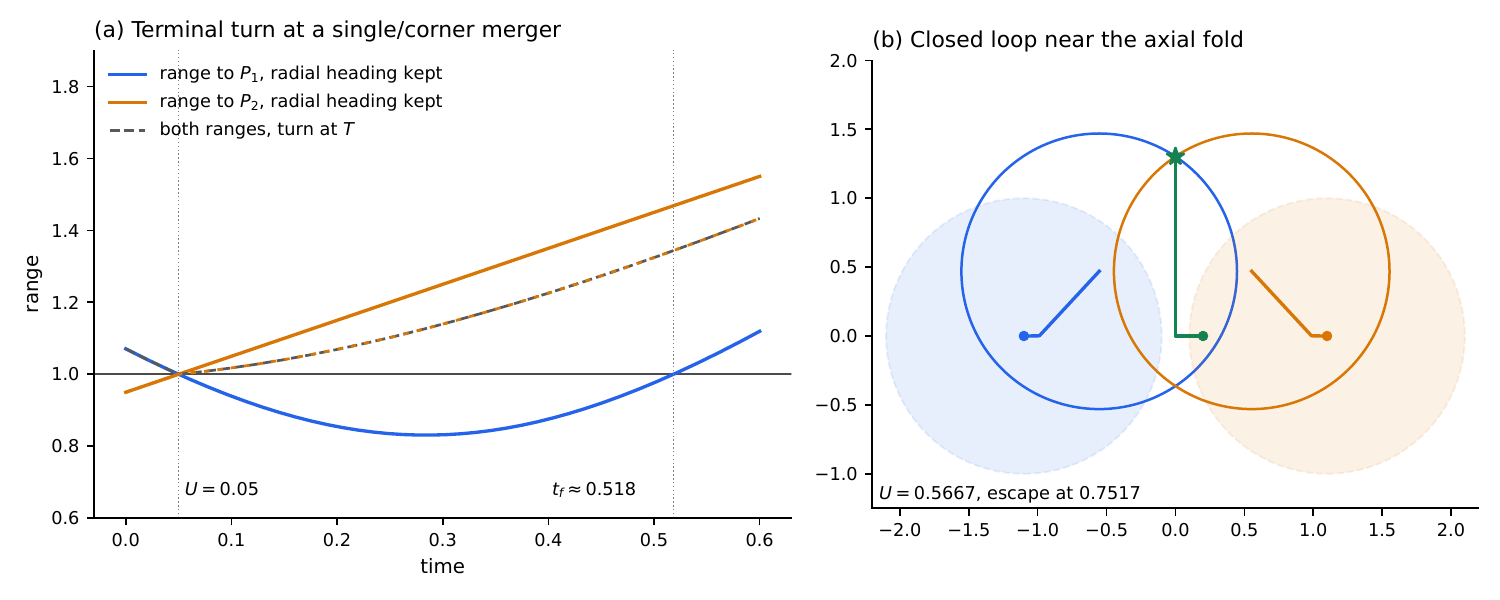}
\caption{(a) Ranges at the single/corner merger state of
Section~\ref{sub:completion} (\(R=1\), \(\mu=2\), \(a=0.8\)) against pursuers
that hold the contact headings \(n_i\). Keeping the radial heading (solid)
re-enters disk~1 and delays escape to \(t_f\approx0.518\); turning to the
bisector at \(T=U=0.05\) (dashed, both ranges equal) escapes at once.
(b) A closed-loop game from \((1.1,0.2001,0)\), \(R=1\), \(\mu=2\), just
past the axial fold: pursuers following the budget-preserving rule of
Section~\ref{sub:validation} against an evader that rushes into the closing
gap and then heads up. Blue and orange paths: \(P_1\) and \(P_2\); green
path: the evader. Shaded:
initial sensing disks; solid circles: sensing disks at the escape time.}
\label{fig:turnloop}
\end{figure}

Tables~\ref{tab:validation} and~\ref{tab:games} list independent
numerical checks. The pointwise checks of Table~\ref{tab:validation} test
the formulas and the inequalities used in the proofs. The simulated
games of Table~\ref{tab:games} play the explicit feedback of
Theorems~\ref{thm:value} and~\ref{thm:extension} and the explicit evader
policies against a range of opponents, including a greedy evader that,
at every step, picks on a grid of headings the one minimizing the escape
time of the state reached after one step of the pursuers' feedback. The checks support the analysis; they do not replace the
proofs. We also simulated two heuristic pursuer rules suggested by
Theorems~\ref{thm:globalvalue} and~\ref{thm:reverse}, which are not the
strategies of the proofs. The first, the \emph{budget-preserving} rule,
sees the evader's current heading and chooses at each step, among
mirror-symmetric unit-speed motions on a grid, the one that maximizes
the minimum of the contact residual over the remaining time budget. The
second is \emph{heading-blind}: it makes the same choice against the
worst evader heading, without seeing it.
Table~\ref{tab:heuristics} gives their worst outcomes, and
Figure~\ref{fig:turnloop}(b) shows one run of the budget-preserving
rule. The code and its documentation are available on request (see Data and code
availability).

\begin{table}[p]\centering\small
\renewcommand{\arraystretch}{1.3}
\begin{tabular}{@{}>{\raggedright\arraybackslash}p{.40\linewidth}>{\raggedright\arraybackslash}p{.24\linewidth}>{\raggedright\arraybackslash}p{.28\linewidth}@{}}
\toprule
Check & Scope & Result\\
\midrule
\multicolumn{3}{@{}l}{\emph{Formulas}}\\
Closed-form characteristics vs.\ integration of \eqref{eq:dynamics},
\eqref{eq:adjoint}
 & 3 benchmarks
 & Difference \(<1.3\times10^{-14}\); \(|\bar H|<9\times10^{-16}\)\\
HJI residual of \(U\), finite differences
 & 160 states
 & \(\le1.9\times10^{-9}\)\\
Distance formula \eqref{eq:Dexplicit} vs.\ direct enumeration
 & 1,552 states
 & \(\le4.5\times10^{-16}\)\\
Fold and cusp times vs.\ \eqref{eq:foldparam}--\eqref{eq:laterfoldroot}
 & 375 states
 & \(\le3.5\times10^{-12}\)\\
\midrule
\multicolumn{3}{@{}l}{\emph{Inequalities used in the proofs} (minimum slack)}\\
Dispersal bound \eqref{eq:tieslack}
 & 9,747 pairs
 & \(-4.4\times10^{-16}\)\\
Fold response \eqref{eq:foldslack}, \eqref{eq:laterslack}
 & \(2.7\times10^{6}\) pairs
 & \(-2.2\times10^{-15}\)\\
Axial fold response \eqref{eq:axialfoldguard}
 & 216,150 headings
 & \(\ge0\)\\
Monotonicity \eqref{eq:monotonecoverage}; junction bound
\eqref{eq:terminaljunctionslack}
 & 11,818; 276,165 cases
 & \(-3.3\times10^{-16}\)\\
Boundary opening \eqref{eq:scaling}, \eqref{eq:openingphase}
 & 336 openings
 & \(-1.0\times10^{-14}R\)\\
Fold opening, Lemma~\ref{lem:foldopening}
 & 1,680 openings
 & \(-2.8\times10^{-15}R\)\\
\bottomrule
\end{tabular}
\caption{Pointwise numerical checks. Minimum slacks are rounded to two
digits; slacks at the level of \(10^{-15}\) are roundoff. Parameters: the
characteristics, HJI and dispersal rows use \(R=10\), \(\mu=2\); the
distance row uses \(R=1\), \(\mu=2\); the fold and cusp row, the two
fold-response rows, and the monotonicity and junction row use \(R=1\) and
\(\mu\in\{1.1,1.5,2,3,10\}\); the two opening rows use 84 double-boundary
and 420 fold states, \(R\in\{1,10\}\), and
\(\mu\in\{1.05,1.2,1.5,2,3,6\}\), with \(7.8\times10^{7}\) and
\(3.9\times10^{8}\) endpoint--horizon pairs. The documentation of the
code lists the states and the exact counts.}
\label{tab:validation}
\end{table}
\clearpage

\begin{table}[!t]\centering\small
\renewcommand{\arraystretch}{1.3}
\begin{tabular}{@{}>{\raggedright\arraybackslash}p{.34\linewidth}>{\raggedright\arraybackslash}p{.22\linewidth}>{\raggedright\arraybackslash}p{.36\linewidth}@{}}
\toprule
Test & States, games & \(t_f-U\)\\
\midrule
Explicit feedback (Theorem~\ref{thm:value}), \(a_0\le a_{\rm tr}\)
 & 60 states, 370 games
 & Pursuers: \(\ge3.5\times10^{-7}\). Evader: \(\le10^{-11}\).\\
Explicit feedback (Theorems~\ref{thm:value} and~\ref{thm:extension}),
\(a_0>a_{\rm tr}\)
 & 180 states, 1,380 games
 & Pursuers: \(\ge4.9\times10^{-5}\). Evader: \(\le10^{-11}\); without the
   turn, \(0.12\)--\(6.5\) late.\\
\bottomrule
\end{tabular}
\caption{Simulated games of the explicit feedback and the explicit evader
policies, \(R=10\), \(\mu=2\), step \(0.002\). The pursuer feedback is
played against four evader types in the first row (constant heading,
chattering, oscillating, greedy) and six in the second (the same, an
evader heading for the centre, and the straight heading without the
turn); the evader policies are played against three pursuer types
(constant random headings, pure pursuit, the feedback). In the second
row 45 states need a terminal turn; 40 of them were also run at step
\(0.001\) (360 further games, not counted here), with the same bounds.
Runs censored at a time cap above \(U\) count as lower-bound successes;
no evader-policy run was censored, and no run was invalid.}
\label{tab:games}
\end{table}

\paragraph{Conventions}
In the simulated games every control is held constant over a time step
and is a unit vector, so each trajectory is an admissible play of the
fixed-speed game. Within each step, candidate exit times are obtained
from the quadratic range equations, and the intervals on which the
evader is outside both disks are identified in floating point, with a
clearance tolerance of \(10^{-10}\) on the ranges; terminal turns are
inserted at their prescribed times. These computations are numerical
checks, not certified first-exit calculations. A run that reaches its
time cap without escaping is recorded as censored, never as an escape
time.

Escape times are computed with the algorithm of
Section~\ref{sub:algorithm} in floating point. A stationary minimum of
the dimensionless contact residual \(F_z/R\) within \(2\times10^{-12}\)
of zero is treated as a
contact, not a crossing, so that the exact fold states, generated from
\eqref{eq:foldparam} and \eqref{eq:cuspparam}, use the later root. This
is a convention: it does not certify the classification of a state
within that distance of a fold, and the implementation flags every such
state. The documentation of the code gives an example, the remaining
conventions and tolerances, and the test states.

\paragraph{Reading Table~\ref{tab:games}}
Each entry is the realized escape time minus the predicted value,
\(t_f-U\). A pursuer guarantee requires \(t_f-U\ge0\) against every evader
tested, and an exact evader policy requires \(t_f-U\le0\) against every
pursuer tested. The pursuer inequality holds in every run. The evader
inequality holds to roundoff under the exit convention above: the largest value of \(t_f-U\) for an
evader policy is \(3.4\times10^{-12}\), at \(R=10\). Without the terminal turn, the
evader policies arrive late, as Proposition~\ref{prop:completion}
predicts.

\paragraph{Heuristic rules}
Table~\ref{tab:heuristics} reports the two heuristic rules at \(R=1\),
\(\mu=2\). A control is a mirror-symmetric pair of unit pursuer
velocities \((\pm\sqrt{1-V^2},\,V)\), whose horizontal components point
toward the other pursuer, with \(V\) on a uniform grid over \([-1,1]\)
that includes both endpoints. Each candidate control is scored at the
state predicted after one time step. There the rules evaluate a signed
extension \(\widetilde D\) of the exterior distance: on the radial piece
of \eqref{eq:Dexplicit} it keeps \(R+\tau-d\) even when this is negative,
that is, when the predicted state lies outside the union of the expanded
disks of radius \(R+\tau\). The
score is the minimum of \(\widetilde F=\widetilde D-\mu\tau\) over the
remaining horizons. This extension is used only to select the control;
escape is detected by the range tests described above, and \(D\) and
\(F\) always denote the actual distance and residual.

\begin{table}[!t]\centering\small
\begin{tabular}{@{}lccc@{}}
\toprule
\emph{Heading-blind rule}, 121 controls
 & \(\Delta t=0.002\) & \(0.001\) & \(0.0005\)\\
\midrule
Axial fold, \(z=(1.1,\,0.2,\,0)\)
 & \(-9.4\times10^{-4}\) & \(-4.9\times10^{-4}\) & \(-2.6\times10^{-4}\)\\
Minimum fold, \(z\approx(1.1,\,0.2312,\,0.1396)\)
 & \(-4.5\times10^{-5}\) & \(-2.4\times10^{-5}\) & \(-1.3\times10^{-5}\)\\
State of \(\mathcal C_\varepsilon\), \(z=(1.4,\,1,\,0.1)\)
 & \(-2.8\times10^{-5}\) & \(-2.8\times10^{-5}\) & \(-2.8\times10^{-5}\)\\
\midrule
\emph{Budget-preserving rule}, \(\Delta t=0.001\)
 & 61 controls & 121 & 241\\
\midrule
Axial fold, \(z=(1.1,\,0.2,\,0)\)
 & \(-1.9\times10^{-4}\) & \(-3.0\times10^{-5}\) & \(-1.5\times10^{-5}\)\\
State of \(\mathcal C_\varepsilon\), \(z=(1.4,\,1,\,0.1)\)
 & \(-4.9\times10^{-5}\) & \(-2.8\times10^{-5}\) & \(-6.4\times10^{-8}\)\\
Worst of the 14 states tested
 & \(-1.9\times10^{-4}\) & \(-3.4\times10^{-5}\) & \(-1.5\times10^{-5}\)\\
\bottomrule
\end{tabular}
\caption{Heuristic pursuer rules, \(R=1\), \(\mu=2\): smallest realized
escape time minus \(U\) over the evaders tested, rounded to two digits;
\(\Delta t\) is the time step. The minimum-fold state is generated from \eqref{eq:foldparam}
with \(a=1.1\) and \(r=1.13\), that is, \(T_F=0.13\); its displayed
coordinates are rounded, and the computations use the generated state.
The other states are exact as printed.}
\label{tab:heuristics}
\end{table}

\newpage
Each entry is the
smallest \(t_f-U\) over
the evaders tested, so a negative entry is a loss for the pursuers. There are four evaders
for the heading-blind rule, one of which sees the pursuers' control for
the current step, and seven for the budget-preserving rule, including
the greedy evader. In every row but the last, the minimum is attained by
the evader that heads straight to the predicted contact point; at an
exact fold state this is the later contact, at time \(T_L=U\), and not
the tangency at \(T_F\). The loss
of the heading-blind rule roughly halves when the step is halved at the
two exact fold states and does not depend on the step at the state of
\(\mathcal C_\varepsilon\), which is not on a fold. The loss of the
budget-preserving rule does not depend on the step: with 121 controls
its worst value over the 14 states is \(-3.4\times10^{-5}\) at steps
\(0.002\), \(0.001\), and \(0.0005\). It decreases, or stays the same,
as the control grid is refined. These are observations at the listed
states and resolutions, not convergence results.

Figure~\ref{fig:turnloop} illustrates the terminal turn and the value
jump. At the merger
state of Section~\ref{sub:completion}, the radial heading alone re-enters
disk~1, while the turn to the bisector completes the escape at \(U\). From
\((a,x,y)=(1.1,0.2001,0)\), just past the axial fold, an evader that rushes
into the closing gap until time \(0.1\), the contact time of the
neighbouring fold state, and then heads
for the upper corner is caught in the gap, as the value jump predicts,
and escapes only at \(0.7517\), after \(U=0.5667\).

\FloatBarrier
\section{Conclusions and open problems}
\label{sec:conclusion}
The escape time of the two-disk detection game has a short description:
expand each sensing disk by the distance its pursuer can travel, and
take the first time at which the evader's reachable disk strictly meets
the region outside both expanded disks. This time is given by an exact
distance formula and a finite scalar algorithm, and it is the value of
the fixed-speed game under pursuer-nonanticipative strategies for every
initial state. Under the reverse ordering, in which the evader reacts to
the pursuers, it is again the value at every state. On regular optimal arcs the active players
move along straight lines in the inertial frame, and the plays come in
two families, a radial race against one pursuer and a
simultaneous exit through a corner of the sensing union. Three features
have no counterpart in the corresponding single-pursuer game with simple
motion: the terminal turn at a
corner, corners at which the straight evader policies are only
\(\varepsilon\)-optimal, and folds of the contact residual across which
the value jumps.

Several questions remain open.
\begin{enumerate}
\item \emph{A uniform guarantee under the reverse ordering.}
Theorem~\ref{thm:reverse} gives, for each nonanticipative evader
strategy, a pursuer control that depends on that strategy. Whether a
single pursuer rule with a fixed reaction delay, or a positional strategy
in the sense of~\cite{krasovskii}, secures \(U-\varepsilon\) against all
evader strategies is open. At a minimum fold every mirror-symmetric
pursuer velocity chosen before the evader's loses to first order against
some evader heading (Remark~\ref{rem:folds}), so such a rule would have
to control
the loss accumulated over its delays. At the two tested fold states,
the heading-blind losses in Table~\ref{tab:heuristics} roughly halve
when the time step is halved.
\item \emph{Attainment in limiting cells.} In \(\mathcal C_\varepsilon\)
the evader reaches the limiting corner at time \(U\), but no turn
completes the escape there, and the straight policies we give are only
\(\varepsilon\)-optimal. We do not prove that the value is unattainable;
whether some measurable evader policy attains it is open.
\item \emph{An explicit global pursuer feedback.} The global strategy is
obtained through a viability argument. In our simulations, the
budget-preserving pursuer rule of Table~\ref{tab:heuristics} performed
as predicted, up to an error that shrinks as its control grid is refined. A proof that it
realizes the value, and a discrete-time version with error bounds, would
make Theorem~\ref{thm:globalvalue} constructive.
\item \emph{Bounded turning rate.} The terminal turn and the dispersal
response both use instantaneous heading changes, which simple motion
allows. With a bound on turning rate, cells of type
\(\mathcal C_{\rm turn}\) would generally have larger escape times and
the partition would change.
\item \emph{More general teams.} With unequal pursuer speeds or sensing radii the
two-active baseline rotates and the closed forms of
Section~\ref{sub:corner} are lost. The attainable-disk definition of
\(U\) extends directly, but the lower-bound argument would need a
replacement for the mirror-symmetric auxiliary system and for the
monotonicity \eqref{eq:monotonecoverage}. Teams of more than two
pursuers raise the same issues.
\end{enumerate}

%% Declarations required by the Journal of the Franklin Institute.
\section*{CRediT authorship contribution statement}
\textbf{Johnny Rubio:} Conceptualization, Methodology, Formal analysis,
Software, Writing -- original draft.
\textbf{Ubaldo Ruiz:} Conceptualization, Methodology, Formal analysis,
Supervision, Writing -- review \& editing.

\section*{Declaration of competing interest}
The authors declare that they have no known competing financial
interests or personal relationships that could have appeared to
influence the work reported in this paper.

\section*{Declaration of generative AI and AI-assisted technologies in
the manuscript preparation process}
During the preparation of this work the authors used Claude (Anthropic)
to check derivations, to write and run the code for the numerical checks
and the figures, and to assist in writing and revising the text. After
using this
tool, the authors reviewed and edited the content as needed and take
full responsibility for the content of the published article.

\section*{Data and code availability}
No experimental data were used. The scripts that reproduce the figures
and every entry of
Tables~\ref{tab:validation}--\ref{tab:heuristics} are available from
the corresponding author upon request, together with a README that
documents the simulation conventions and tolerances, the test states,
and the remaining results of the heuristic simulations.

\section*{Funding}
This research did not receive any specific grant from funding agencies
in the public, commercial, or not-for-profit sectors.

\appendix
%% Keep the heading "Appendix A." while cross-references print "A".
\makeatletter
\renewcommand\thesection{\Alph{section}}
\def\@seccntformat#1{\ifnum\pdfstrcmp{#1}{section}=0 Appendix~\fi
  \csname the#1\endcsname\@seccntDot\hskip 0.5em}
\makeatother
\section{An explicit invariant extension beyond \texorpdfstring{\(a_{\rm c}\)}{a\_c}}
\label{app:extension}
Theorem~\ref{thm:value} establishes the value together with an explicit
pursuer feedback when \(a_0\le a_{\rm c}=R\sqrt{1-\mu^{-2}}\).
Theorem~\ref{thm:globalvalue} establishes the value for every state, but
its pursuer strategy is not given in closed form. This appendix shows
that the same explicit feedback still realizes the value on an invariant
region beyond \(a_{\rm c}\).

The obstacle above \(a_{\rm c}\) is that a corner might fail the
completion test \(\Gamma\ge0\), and the root of \(D(T)=\mu T\) might
not be unique. Both problems disappear when the escape horizon is long
compared with the separation: once \(R+T>a/\kappa\), the expanded
circles meet at a wide enough angle that \(\mu h/r>1\). An evader far
from the symmetry axis cannot reach a corner quickly, because every
corner lies on \(x=0\), so \(\mu T\ge|x|\). Large \(|x|\) therefore
forces long horizons. The extension therefore asks for \(|x|\) to be large
compared with a threshold \(S(a)\), and shows that this inequality
survives the play. The pursuers close at a rate proportional to \(a\),
which lowers the threshold \(S(a)\) faster than the evader, at speed
\(\mu\), can reduce \(|x|\). The quantity \(G=|x|-S(a)\) in the proof
records this race. Set
\begin{equation}\label{eq:extensionS}
 \begin{gathered}
 \kappa=\sqrt{1-\mu^{-2}},\qquad B=\frac{R}{\mu-1},\\
 S(a)=\mu(R+B)\log\frac{a}{R\kappa}
       +\mu\left(\frac{a}{\kappa}-R\right),\qquad a>R\kappa.
 \end{gathered}
\end{equation}
\begin{theorem}[Invariant extension]\label{thm:extension}
Suppose the initial state is strictly inside the sensing union and either
\(a_0\le R\kappa\), or
\begin{equation}\label{eq:extensiondomain}
 a_0>R\kappa,\qquad |x_0|>S(a_0).
\end{equation}
Then the pursuer-nonanticipative value \(V_P\), defined in
\eqref{eq:strategyvalue}, is the unique positive root of
\(D(T)=\mu T\). Selected corner contacts, including single/corner
mergers, admit the completion of
Proposition~\ref{prop:completion}. The explicit feedback of
Theorem~\ref{thm:value} (\(u_i=n_i\), the dispersal response
\eqref{eq:tie} on \(y=0\), and pure pursuit once a single pursuer is
decisive) realizes this value.
\end{theorem}
\begin{proof}
The case \(a\le R\kappa\) is Theorem~\ref{thm:value}. Otherwise, every
corner root satisfies \(\mu T\ge|x|\), since its inertial escape
point has horizontal coordinate zero. Thus
\[
 |x|>S(a)>\mu(a/\kappa-R)
 \ \Longrightarrow\ R+T>a/\kappa
 \ \Longrightarrow\ h/(R+T)>1/\mu.
\]
Every possible corner zero is immediately usable and transverse
in the scalar root equation: its envelope slope is \(1/c<\mu\).
Exposed-arc zeros have slope \(1-\mu<0\) for \(D(T)-\mu T\).
This continuous function starts positive and is negative beyond \(B\)
by Proposition~\ref{prop:bounds}. Since every
zero crosses downward, there is exactly one. The envelope gradients and
smooth-cell lower strategy apply without requiring positive incoming
radial rates.

During a two-active phase use \(u_i=n_i\). The baseline does not rotate,
and
\[
 \dot a=-a/(R+U),\qquad \dot x=\mu e_x,\qquad 0<U\le B,
\]
where \(U=U(z(t))\) is the escape time of the current state, as in
Section~\ref{sec:regime}. At an upper/lower tie, response \eqref{eq:tie} has
\(\dot a\le-a/(R+U)\) and the same \(\dot x\). Define \(G=|x|-S(a)\).
While \(a>R\kappa\), \(G>0\) implies \(x\ne0\), and
\begin{align*}
 \dot G
 &\ge-\mu+
 \left[\frac{\mu(R+B)}a+\frac\mu\kappa\right]\frac{a}{R+U}\\
 &=\mu\left[\frac{R+B}{R+U}-1+\frac{a}{\kappa(R+U)}\right]>0.
\end{align*}
Hence the extension condition is preserved until entry into
\(a\le R\kappa\). If a single-active branch is selected first, switch
to its direct pure-pursuit lower strategy, which is independent of later
separation. Tie handling and active-set matching are as in
Theorem~\ref{thm:value}.

Use the absolutely continuous chain-rule argument of
Theorem~\ref{thm:value} to integrate \(\dd U/\dd t\ge-1\) throughout
the cooperative phase, followed if needed by the single-active lower bound.
The terminal boundary
is immediately usable in the visited domain, so \(U\) tends to zero at
exit. This gives \(t_f\ge U(z(0))\). The completed evader policy supplies
the matching upper bound.
\end{proof}

\section{Proof of Lemma~\ref{lem:realization}}
\label{app:realization}
We use the notation of Section~\ref{sub:feedback}: \(U\) is the escape
time of the current state, \(k(z)=h/r\) with \(r=R+U\) and
\(h=\sqrt{r^2-a^2}\), and \(\operatorname{Sgn}\) is the set-valued sign.
\begin{proof}
\emph{Regularity of \(k\).} Let \(z_0\in\mathcal O\) and
\(\bar U=U(z_0)\). Since \(R+\bar U>r_m(z_0)\) and \(r_m\) is continuous,
near \((z_0,\bar U)\) the contact residual is given by the
corner expression
\(F_z(T)=\sqrt{x^2+(h-|y|)^2}-\mu T\), with \(h=\sqrt{(R+T)^2-a^2}\),
whose \(T\)-derivative is \(1/c-\mu<0\). By the implicit function theorem
there are \(\delta\in(0,\bar U)\), a neighborhood \(N\) of \(z_0\), and a
\(C^1\) function \(T^*\) of \((a,x,|y|)\) on \(N\) such that, for \(z\in N\),
\(F_z\) is strictly decreasing on \([\bar U-\delta,\bar U+\delta]\) and vanishes
there only at \(T^*(z)\). Condition \eqref{eq:realizationdomain} makes
this root the selected one. Indeed, \(F_{z_0}\ge m>0\) on the compact
interval \([0,\bar U-\delta]\), so, since \(F\) is continuous in
\((z,\tau)\), \(F_z>0\) there for \(z\) near \(z_0\); and \(F_z>0\) on
\([\bar U-\delta,T^*(z))\) by monotonicity. Hence \(U=T^*\) near \(z_0\), and
these states satisfy \eqref{eq:realizationdomain} and, by continuity of
\(T^*\) and \(r_m\), the strict selection \(R+U>r_m\); so \(\mathcal O\) is
open and \(U\) is \(C^1\) in \((a,x,|y|)\) on it, hence locally Lipschitz
in \((a,x,y)\). Without
\eqref{eq:realizationdomain} the conclusion fails. For \(R=1\), \(\mu=2\),
take \(T_F=0.1\) and \(a=1.1\sqrt{0.91}=11\sqrt{91}/100\), so that \(k=0.3\) at
\(r=R+T_F\). By \eqref{eq:foldparam} the state
\((a,x,y)=(11\sqrt{91}/100,\ 0.16,\ 0.21)\) is a minimum-fold state. Its selected root
\(T_L\approx0.2320\) is simple, with \(\mu c_L\approx1.116\), and
\(k\approx0.524\) there, but the states \((a,0.16,0.21+\epsilon)\),
\(\epsilon>0\), escape near \(T_F=0.1\), and their values of \(k\) tend to
\(0.3\) as \(\epsilon\downarrow0\). On \(\mathcal O\), therefore, \(k\) is
Lipschitz, with constant \(\ell\), on a compact neighborhood
\(\mathcal C_\rho\subset\mathcal O\) of \(\mathcal C\), and
\(0<k_{\min}\le k\le\bar k<1\) there.

\emph{The vertical inclusion.} Let \(\tilde k:[0,t_0]\to[k_{\min},\bar k]\)
be continuous. The inclusion
\(\dot y\in\mu e_y(t)-\tilde k(t)\operatorname{Sgn}(y)\), \(y(0)=y_0\),
has bounded, compact, interval-valued right-hand side that is measurable
in \(t\) and upper semicontinuous in \(y\), so it has an absolutely
continuous solution~\cite{filippov}. Put \(v=\mu e_y-\dot y\), so
\(v=\tilde k w\) with \(w(t)\in\operatorname{Sgn}(y(t))\). Since \(\dot y=0\)
almost everywhere on the zero set of an absolutely continuous function,
\(v=\mu e_y\) and \(|\mu e_y|\le\tilde k\) almost everywhere on \(\{y=0\}\).

\emph{A comparison estimate.} Let \((y_j,v_j,w_j)\), \(j=1,2\), solve the
inclusion with the same \(y_0\) and \(e\) and with \(\tilde k_1,\tilde
k_2\). Where \(y_1\ne y_2\), put \(\chi=\operatorname{sgn}(y_1-y_2)\).
Monotonicity of \(\operatorname{Sgn}\) gives \(\chi(w_1-w_2)=|w_1-w_2|\),
and from \(v_1-v_2=\tilde k_1(w_1-w_2)+w_2(\tilde k_1-\tilde k_2)\),
\[
 |v_1-v_2|\le\tilde k_1|w_1-w_2|+|\tilde k_1-\tilde k_2|
 \le\chi(v_1-v_2)+2|\tilde k_1-\tilde k_2|.
\]
Because \(\dot y_1-\dot y_2=-(v_1-v_2)\), the term \(\chi(v_1-v_2)\)
equals \(-\frac{\dd}{\dd t}|y_1-y_2|\). Where \(y_1=y_2\), almost
everywhere \(\dot y_1=\dot y_2\), hence \(v_1=v_2\) and
\(\frac{\dd}{\dd t}|y_1-y_2|=0\). Therefore, almost everywhere,
\begin{equation}\label{eq:realizationcomparison}
 |v_1-v_2|\le-\frac{\dd}{\dd t}|y_1-y_2|+2|\tilde k_1-\tilde k_2|,
 \qquad
 \frac{\dd}{\dd t}|y_1-y_2|\le|\tilde k_1-\tilde k_2|.
\end{equation}
Integrating from \(0\), where \(y_1=y_2\),
\[
 |y_1-y_2|(t)\le\int_0^t|\tilde k_1-\tilde k_2|,\qquad
 \int_0^t|v_1-v_2|\le2\int_0^t|\tilde k_1-\tilde k_2|.
\]
With \(\tilde k_1=\tilde k_2\) this shows that the inclusion has a unique
solution, and that its realized \(v\) is unique almost everywhere.

\emph{The closed loop.} For a continuous path \(\tilde z\) in
\(\mathcal C_\rho\) on \([0,t_0]\), solve the vertical inclusion with
\(\tilde k=k(\tilde z)\), and set
\(a(t)=a_0-\int_0^t\sqrt{1-v^2}\) and \(x(t)=x_0+\mu\int_0^te_x\). Call
the resulting path \(\Phi(\tilde z)\). On \(|v|\le\bar k\) the map
\(v\mapsto\sqrt{1-v^2}\) is Lipschitz with constant
\(\bar C=\bar k/\sqrt{1-\bar k^2}\), so the estimates above give
\[
 \sup_{[0,t_0]}\big|\Phi(\tilde z_1)-\Phi(\tilde z_2)\big|
 \le(1+2\bar C)\ell t_0\sup_{[0,t_0]}|\tilde z_1-\tilde z_2|.
\]
All speeds are bounded by \(\mu+1\). Hence, for
\(t_0<\min\{\rho/(\mu+1),\,1/((1+2\bar C)\ell)\}\), which is independent
of \(z_0\in\mathcal C\), \(\Phi\) maps paths that start at \(z_0\) and
stay within distance \(\rho\) of it into themselves and is a
contraction. Its unique fixed point is the closed-loop trajectory on
\([0,t_0]\). Conversely, every pair \((z,v)\) as in the statement is a
fixed point of \(\Phi\), by uniqueness for the vertical inclusion.
Repeating the construction from \(z(t_0)\) extends the solution for as
long as it remains in \(\mathcal C\). Finally, \(\Phi\) restricted to
\([0,t]\) uses \(e\) only on \([0,t]\), so the fixed point on \([0,t]\)
is determined by \(e\) on \([0,t]\).
\end{proof}

\section{Proof of Lemma~\ref{lem:auxviability}}
\label{app:auxviability}
We use the notation of Section~\ref{sec:global}: the auxiliary dynamics
\eqref{eq:auxdynamics}, the budget set \(\mathcal K\) of
\eqref{eq:budgetset}, and its extension \(\widehat{\mathcal K}\) of
\eqref{eq:extendedbudget}. The five numbered cases below are those of
Table~\ref{tab:auxcases}.
\begin{proof}
By reflection, the cooperative part reduces to \(a>0\), \(|x|<a\),
\(y\ge0\).

\emph{Reduction to linearized inequalities.} Write \(Z=(z,s)\) for the
augmented state; it is unrelated to the costate \(p\) of
Section~\ref{sec:hamiltonian}. Fix \(Z_0=(z_0,s_0)\in\mathcal K\) with \(s_0>0\) and \(z_0\) in the interior of
the sensing union and of the cooperative strip, and suppose that no
active horizon (defined below) lies at the axial kink; the axial and
terminal degeneracies are treated in cases~4 and~5 below. With
\(\tau=\alpha s\), \(\alpha\in[0,1]\), membership in \(\mathcal K\) near
\(Z_0\) is the condition \(\Phi(Z,\alpha):=F_z(\alpha s)\ge0\) for all
\(\alpha\), since the constraints \(z\in\Sigma\) and \(s\le B\) are
inactive or, at \(s=B\), met because \(\dot s=-1\). Let
\(A_0=\{\alpha\in[0,1]:\Phi(Z_0,\alpha)=0\}\) be the set of active
horizons. It is compact and excludes \(\alpha=0\), because
\(F_{z_0}(0)>0\).

Smoothness is needed only near \(A_0\). Each \(\alpha_0\in A_0\) has an
open neighborhood \(N_{\alpha_0}\) such that, for \(Z\) near \(Z_0\) and
\(\alpha\in N_{\alpha_0}\), \(\Phi(Z,\alpha)=\min_b\varphi_b(Z,\alpha)\) is
the minimum of finitely many functions
\(\varphi_b(z,s,\alpha)=\psi_b(z,\alpha s)\) with \(\psi_b\) of class
\(C^1\) in \((z,\tau)\): at \(y=0\) the upper and lower corner branches,
and elsewhere the selected branch, which is \(C^1\) across a
single/corner merger by Lemma~\ref{lem:merger}. Let \(N\) be the union of
finitely many such neighborhoods covering \(A_0\). On the compact set
\([0,1]\setminus N\), \(\Phi(Z_0,\cdot)\ge m_0>0\), and since \(F\) is
jointly continuous, \(\Phi(Z,\alpha)\ge m_0/2\) there for all \(Z\) near
\(Z_0\). Kinks of \(F_z\) at inactive horizons therefore play no role.
For example, at \((a,X,Y)=(1.1,0.3,0)\) with \(R=1\), \(\mu=2\), the join
horizon \(\tau=a-R=0.1\) is a kink of \(F_z\) with one-sided derivatives
\(-1\) and \(1.1/0.3-2\); when it is inactive it lies in
\([0,1]\setminus N\). Let
\(\mathcal I=\{(b,\alpha):\alpha\in A_0,\ \varphi_b(Z_0,\alpha)=0\}\) be the
active index set; it is compact.

At a positive active horizon \(\tau\), a smooth corner has
\(h=\sqrt{(R+\tau)^2-a^2}>y\) and
\[
 \partial_a F_z(\tau)=-\frac{a(h-y)}{hD_z(\tau)}<0,
\]
and an exposed branch has \(\partial_a F=-(a-|x|)/d<0\). With
\(\mathbf d=(-1,0,0,0)\), the direction that decreases \(a\), continuity
and compactness give \(\delta_0>0\) with
\(\nabla_Z\varphi_b(Z_0,\alpha)\cdot\mathbf d\ge\delta_0\) on \(\mathcal I\).

Let \(\zeta\) be a velocity of \eqref{eq:auxdynamics} satisfying the
linearized inequalities
\begin{equation}\label{eq:linearizedtangency}
 \nabla_Z\varphi_b(Z_0,\alpha)\cdot\zeta\ge0\qquad\text{for all }(b,\alpha)\in\mathcal I.
\end{equation}
We show that \(\zeta\) lies in the contingent cone of \(\mathcal K\) at
\(Z_0\). Fix \(\varepsilon>0\) and put \(\zeta_\varepsilon=\zeta+
\varepsilon\mathbf d\). By continuity there is a neighborhood \(N'\) of
\(\mathcal I\) in the index set on which
\(\nabla_Z\varphi_b(Z_0,\alpha)\cdot\zeta_\varepsilon\ge\varepsilon\delta_0/2\),
and by compactness \(\varphi_b(Z_0,\alpha)\ge m_1>0\) for the remaining
indices \((b,\alpha)\) with \(\alpha\) in the closure of \(N\). Since the
\(\varphi_b\) are \(C^1\) in \(Z\) uniformly in \(\alpha\), for all
sufficiently small \(t>0\) we have
\(\varphi_b(Z_0+t\zeta_\varepsilon,\alpha)\ge t\varepsilon\delta_0/4>0\) on
\(N'\) (using \(\varphi_b(Z_0,\alpha)\ge0\)),
\(\varphi_b(Z_0+t\zeta_\varepsilon,\alpha)\ge m_1/2\) at the other indices
over \(N\), and \(\Phi(Z_0+t\zeta_\varepsilon,\alpha)\ge m_0/2\) for
\(\alpha\notin N\). Hence
\(Z_0+t\zeta_\varepsilon\in\mathcal K\) for small \(t>0\), so
\(\zeta_\varepsilon\) lies in the contingent cone. The contingent cone is
closed, and letting \(\varepsilon\to0\) gives the claim. It therefore
suffices, at such points, to choose \((C,V)\) so that
\eqref{eq:linearizedtangency} holds. For an endpoint index \(\alpha=1\)
the left side of \eqref{eq:linearizedtangency} is
\(\frac{\dd}{\dd t}F_{z(t)}(s(t))\), and for an interior minimum, where
\(\partial_\tau F=0\), it is \(\dot g_F\).

By \eqref{eq:Dderivatives}, the positive active horizons are the endpoint
\(s\), a single interior local minimum, or both; an isolated maximum
cannot be active with nonnegative values on both sides, and at a cusp the
minimum and endpoint coalesce. The following controls cover every case
without dividing by \(\mu c-1\).
\begin{enumerate}
\item \emph{Smooth endpoint only.} Use \((C,V)=(a,h)/(R+s)\), with the
vertical sign reflected when necessary. At \(F_z(s)=0\),
\[
 \frac{\dd}{\dd t}F_{z(t)}(s(t))=\mu(1-e_*\cdot e)\ge0,
\]
where \(e_*\) is the incoming heading at the selected corner. At a
single/corner merger the exposed and corner gradients and controls
agree, by Lemma~\ref{lem:merger}; a single-active branch can also be handed to pure pursuit.

\item \emph{Smooth minimum.} Use \((C,V)=(a,h_F)/(R+T_F)\), for which
\(\dot g_F=\mu(1-e_F\cdot e)\ge0\). If the minimum and the later
endpoint are both active, use \eqref{eq:foldguard}; then
\eqref{eq:foldslack} and \eqref{eq:laterslack} hold together. At a cusp
there is one active constraint with \(\partial_\tau F_z=0\), and the inward argument
applies with the control \((a,h_F)/(R+T_F)\) at the coalesced horizon.

\item \emph{Upper/lower tie.} At \(y=0\), an active endpoint has
\(D=\sqrt{x^2+h^2}=\mu s\). With \(r=R+s\), set
\[
 V=\operatorname{clip}(\mu e_y,-h/r,h/r),\qquad C=\sqrt{1-V^2}.
\]
The smaller of the two active derivatives has numerator
\[
 aC+\mu x e_x-h|\mu e_y-V|-r+\mu D\ge0,
\]
which is the estimate from the dispersal step of
Theorem~\ref{thm:value}, before division by a normalizing factor. It
therefore remains valid when \(\mu c=1\).

\item \emph{Axial folds and the axial cusp.} At an axial minimum with
\(R<a<a_*\) and \(X=\mu(a-R)>0\), the local minimum is continuously
differentiable across \(y=0\), with
\begin{equation}\label{eq:axialenvelope}
 g_F=X-\mu(a-R)-\frac{\mu y^2}{2(a-\mu X)}+O(y^4),
 \qquad a-\mu X>0.
\end{equation}
Its linearized condition is \(\mu(\operatorname{sgn}(x)e_x+C)\ge0\).
In the reduction above, the constraint attached to this horizon is the
envelope \(g_F\), not the residual at a fixed horizon, which has a kink
there; the inward correction still applies, because
\(\partial_ag_F=-\mu<0\).
When the later endpoint is also active, \eqref{eq:axialfoldguard} and
the tie response give both inequalities. If the budget equals the kink
time \(s=a-R\), use \(C=1\), \(V=0\); then
\(\frac{\dd}{\dd t}[X-\mu(a-R)]=\mu(1+\operatorname{sgn}(x)e_x)\ge0\)
and \(\frac{\dd}{\dd t}[s-(a-R)]=0\). Strict inequality enters the
feasible side, and equality forces \(e=(-\operatorname{sgn}(x),0)\),
along which the state stays on the feasible axial contact. Feasibility
can also be checked exactly. Take \(x>0\) and put \(d_0=a-X>0\). For a
fixed heading \(e\) and small \(t>0\) the state is \(a_t=a-t\),
\(x_t=X+\mu e_xt\), \(y_t=\mu e_yt\), with budget \(s_t=s-t\). Every
horizon \(\tau\le s_t\) has \(R+\tau\le a_t\), so the expanded disks do
not overlap and \(D_{z_t}(\tau)=R+\tau-d_t\), where the nearer range
\(d_t\) satisfies
\([d_0+(\mu-1)t]^2-d_t^2=2\mu(1+e_x)t(d_0-t)\ge0\). Since
\(R-d_0=(\mu-1)s\), this gives
\(F_{z_t}(\tau)\ge(\mu-1)(s_t-\tau)\ge0\) for \(0\le\tau\le s_t\). This
construction also covers the axial cusp \(a=a_*\), where
\eqref{eq:axialenvelope} does not apply. For a budget strictly between
the axial contact and the later root, only the minimum is active.

\item \emph{The double sensing boundary.} A positive budget there
requires \(x=0\), \(b=|y|=\sqrt{R^2-a^2}\), and \(\mu b<R\). Take \(y\ge0\).
Then \(F_{z_0}(\tau)=\sqrt{(R+\tau)^2-a^2}-b-\mu\tau\) is concave, vanishes
at \(\tau=0\), and has no interior minimum, so the active horizons are
\(\tau=0\) and, if \(s=U(z_0)\), the endpoint \(\tau=s\). The horizon
\(\tau=0\) is degenerate: there \(r_m=R\), so the exposed and corner
pieces of \eqref{eq:Dexplicit} meet at the active horizon itself.

For \(b>0\) use \((C,V)=(a,h_s)/(R+s)\) with \(h_s=\sqrt{(R+s)^2-a^2}\),
the corner control at horizon \(s\); for \(b=0\) use the tie response of
case~3, which has \(C\ge a/(R+s)>0\). With
\(\eta=(a\dot a+b\dot y)/R\) and \(h_s=b+\mu s+F_{z_0}(s)\), every evader
heading gives
\begin{equation}\label{eq:terminaljunctionslack}
 \eta\le-\frac{R-\mu b}{R+s}
          -\frac{bF_{z_0}(s)}{R(R+s)}<0.
\end{equation}
For \(b>0\) an active endpoint is treated as in case~1, since the
control is the corner control at horizon \(s\) and \(\partial_aF<0\) there;
for \(b=0\) the endpoint lies on \(y=0\), where the two corner branches are
individually smooth and their tie is handled by case~3. As in the
reduction step it suffices to show that \(Z_0+t\zeta_\varepsilon\in\mathcal
K\) for small \(t>0\), where \(\zeta_\varepsilon=\zeta+\varepsilon\mathbf d\).
This makes the endpoint inequality strict, and it replaces \(\eta\) by
\(\eta_\varepsilon=\eta-\varepsilon a/R<0\). Write
\(z(t)=z_0+t\zeta_\varepsilon\). Horizons bounded away from zero, up to and
including an active endpoint, are covered by the reduction step: for
\(b>0\), \(F_{z_0}\) is smooth there; for \(b=0\) the corner branches are
individually smooth, and their tie is handled by the finite-branch
reduction. It remains to treat horizons near zero, which
we split into three ranges.

\emph{(i) Scaled horizons \(\tau=t\rho\), \(0\le\rho\le\rho_1\), when
\(b>0\).} Put \(k=b/R\in(0,1/\mu)\) and \(\xi=\dot x\). The distance in
\eqref{eq:Dexplicit} is not differentiable at \((z_0,0)\): the corner
expression \(D_c=\norm{(x,\,h-y)}\) vanishes there, and the exposed
expression involves the nearer-pursuer choice through \(|x|\). We
therefore expand the smooth quantities inside these expressions and use
Lipschitz bounds for the rest. Near \((z_0,0)\) we have \(h\ge b/2>0\), so
\(h\) is smooth there. Along \(z(t)\), and for \(\tau=t\rho\) with \(\rho\)
bounded,
\[
 x(t)=t\xi,\qquad h(t,t\rho)-y(t)=t\,\frac{\rho-\eta_\varepsilon}{k}+O(t^2),
\]
and since the norm is \(1\)-Lipschitz,
\(D_c=t\sqrt{\xi^2+((\rho-\eta_\varepsilon)/k)^2}+O(t^2)\). In the exposed
expression \(r-d\), the nearer range \(d\) is the smaller of two ranges
that are smooth near \(z_0\) and equal to \(R\) there, so
\(d(t)=R+t\dot d+O(t^2)\) with \(\dot d=\eta_\varepsilon-a|\xi|/R\), the
smaller of their two rates. Both expressions are locally Lipschitz and
agree on the merger, whose scaled location \(\rho_m\) is determined to
first order in the same way. Hence, uniformly for \(\rho\in[0,\rho_1]\),
\[
 F_{z(t)}(t\rho)=t\,f(\rho)+O(t^2),\qquad
 f(\rho)=\begin{cases}
 -\dot d-(\mu-1)\rho,&\rho\le\rho_m,\\
 \sqrt{\xi^2+\big((\rho-\eta_\varepsilon)/k\big)^2}-\mu\rho,&\rho\ge\rho_m,
 \end{cases}
\]
where the radial interval is absent if \(\rho_m\le0\). Because \(\mu k<1\)
and \(\eta_\varepsilon<0\), the corner coefficient satisfies
\[
 \sqrt{\xi^2+\big((\rho-\eta_\varepsilon)/k\big)^2}-\mu\rho
 \ \ge\ \frac{\rho-\eta_\varepsilon}{k}-\mu\rho
 \ =\ -\frac{\eta_\varepsilon}{k}+\Big(\frac1k-\mu\Big)\rho
 \ \ge\ -\frac{\eta_\varepsilon}{k}>0.
\]
The radial coefficient is affine and decreasing on \([0,\rho_m]\) and
equals the corner coefficient at \(\rho_m\), so it is at least as large.
Hence \(f\ge f_*:=-\eta_\varepsilon/k>0\), and
\(F_{z(t)}(t\rho)\ge tf_*/2\) for \(\rho\in[0,\rho_1]\) and all sufficiently
small \(t\).

\emph{(ii) Intermediate horizons \(t\rho_1\le\tau\le\tau_1\), when
\(b>0\).} Since \(F_{z_0}\) is concave with \(F_{z_0}(0)=0\) and
\(F_{z_0}'(0)=R/b-\mu>0\), there are \(\tau_1\in(0,U(z_0))\) and \(c_1>0\)
with \(F_{z_0}(\tau)\ge c_1\tau\) on \([0,\tau_1]\). For
\(\tau\in[0,\tau_1]\) and \(z\) near \(z_0\), each expression in
\eqref{eq:Dexplicit} is Lipschitz in \(z\), uniformly in \(\tau\): the corner
expression is a norm of smooth quantities, because \(h\ge b/2\), and the
exposed expression involves \(r\), \(|x|\), and ranges bounded away from
zero. The selected expressions agree where the selection changes, so
\(D_z(\tau)\) is Lipschitz in \(z\) there, uniformly in \(\tau\). Hence
\(|F_{z(t)}(\tau)-F_{z_0}(\tau)|\le c_2t\) on \([0,\tau_1]\) for some \(c_2\).
Choosing \(\rho_1=2c_2/c_1\) before applying (i) gives
\(F_{z(t)}(\tau)\ge c_1\tau-c_2t\ge c_2t>0\) in this range.

\emph{(iii) Small horizons when \(b=0\).} Then \(a_0=R\), \(x_0=y_0=0\),
and \(F_{z_0}(\tau)=\sqrt{2R\tau+\tau^2}-\mu\tau\). The pursuers'
midpoint moves only vertically, at speed at most one, so the evader is
within \((\mu+1)t\) of it, and \(a(t)\le R-ct\) with \(c=a/(R+s)>0\). The
distance from the midpoint to the exterior of the expanded union at
horizon \(\tau\) is the corner height
\(h_t(\tau)=\sqrt{(R+\tau)^2-a(t)^2}\), which is at least
\(\max\{\sqrt{Rct},\sqrt{2R\tau}\}\) for small \(t\). Hence
\(F_{z(t)}(\tau)\ge\max\{\sqrt{Rct},\sqrt{2R\tau}\}-(\mu+1)t-\mu\tau\). For
\(\tau\le t^{3/2}\) the first term dominates, and for
\(t^{3/2}\le\tau\le\tau_1\) the second does, so \(F_{z(t)}(\tau)>0\) on
\([0,\tau_1]\) for small \(t\).

A usable double boundary has \(U=0\) and contributes no positive-budget
face.
\end{enumerate}

Outside the cooperative strip, or whenever a single-disk budget suffices,
the corresponding pursuer uses pure pursuit and the other mirrors its
horizontal velocity as in \eqref{eq:symmetricball}. The bound
\(\dot d_i\le\mu-1\) preserves \(d_i+(\mu-1)s\le R\), which defines a
subset of \(\widehat{\mathcal K}\). Coincident pursuers and zero range
are covered by the same construction; at zero range the active pursuer
takes heading \(e\), and the set of zero-range times has measure zero
when \(\mu>1\). At \(s=0\) the clock enters the unconstrained half-space,
and at \(s=B\) it points inward.

These cases verify the contingent condition at every point of
\(\widehat{\mathcal K}\), including the nonsmooth junctions. Since the
contingent cone is contained in the polar of every proximal normal cone,
\(\widehat{\mathcal K}\) satisfies the proximal-normal characterization
of a discriminating domain~\cite[Proposition~2.2]{cqs}: for every
\(Z\in\widehat{\mathcal K}\), every proximal normal \(\nu\) at \(Z\), and
every evader heading \(e\), some pursuer control gives
\(\langle f(Z,e,(C,V)),\nu\rangle\le0\). We now apply the
characterization of discriminating domains by nonanticipative strategies
of Cardaliaguet~\cite{cardaliaguet96}, stated as Theorem~2 in the
preprint version~\cite{cardaliaguetwp}. In its notation the state is
\(Z=(a,x,y,s)\in\R^4\); the player who wants to stay in the closed set
(there called Victor) is the pursuer team, with control set the closed
unit disk of \((C,V)\); the other player (Ursula) is the evader, with
control set the unit circle; and the dynamics are
\(f(Z,e,(C,V))=(-C,\mu e_x,\mu e_y-V,-1)\). The hypotheses hold: both
control sets are compact metric spaces; the pursuers' control set is
convex and \(f\) is affine in it; \(f\) is continuous, and Lipschitz in
\(Z\) (it does not depend on \(Z\)); and Isaacs' condition holds because
\(f\) is the sum of an evader term and a pursuer term. The theorem then
provides, for every initial point of \(\widehat{\mathcal K}\), a single
nonanticipative strategy of the auxiliary pursuers that keeps the state
in \(\widehat{\mathcal K}\) on \([0,\infty)\) against every measurable
evader heading.
\end{proof}

\section{Proofs for the reverse ordering}
\label{app:reverse}
This appendix proves Lemma~\ref{lem:foldopening} and
Theorem~\ref{thm:reverse}, in the notation of Sections~\ref{sec:global}
and~\ref{sub:foldopening}.

\subsection{Proof of Lemma~\ref{lem:foldopening}}
\begin{proof}
After the opening, the half-separation is
\(a-t_1\sqrt{1-k^2}=(R-t_1+T_F)\sqrt{1-k^2}\), and, relative to the moving
midpoint, the evader lies within \(\mu t_1\) of the reference point
\(\bar E=(x,y-kt_1)=(\mu T_F\sqrt{c_k},\,k(A-t_1))\). By \eqref{eq:foldkT}, the
reference state \(\bar z=(a-t_1\sqrt{1-k^2},\bar E)\) is the minimum-fold state
with the same \(k\) and \(T_F\) in the game with sensing radius \(R-t_1\),
where \(A\) becomes \(A-t_1\). It lies on the open minimum fold of that
game because \((R-t_1)c_k>mT_F\) is \(t_1<\Delta\), and by
\eqref{eq:laterfoldroot} its later root is \(\widetilde T_L(t_1)\). Write
\(F^{\rho}\) and \(D^{\rho}\) for the contact residual and the distance
function of the game with sensing radius \(R-\rho\). Then \(F^{t_1}_{\bar z}(T)\ge0\) for \(0\le T\le\widetilde T_L(t_1)\), with
zeros exactly at \(T=T_F\) and \(T=\widetilde T_L(t_1)\).

From \(z(t_1)\), the expanded radius at horizon \(\tau\) is
\(R+\tau=(R-t_1)+(\tau+t_1)\). The exterior of the expanded union is
therefore the same set for \(z(t_1)\) at horizon \(\tau\) and for \(\bar z\),
with sensing radius \(R-t_1\), at horizon \(\tau+t_1\). The distance to a
closed set is \(1\)-Lipschitz, so
\begin{equation}\label{eq:foldcomparison}
 F_{z(t_1)}(\tau)\ \ge\ D^{t_1}_{\bar z}(\tau+t_1)-\mu t_1-\mu\tau
 =F^{t_1}_{\bar z}(\tau+t_1),\qquad\tau\ge0.
\end{equation}
This gives (ii). The same comparison with any \(t\in(0,t_1]\) in place of
\(t_1\), with its own reference state \(\bar z_t\), gives at \(\tau=0\) that
\(F_{z(t)}(0)\ge F^{t}_{\bar z_t}(t)>0\), since \(0<t<T_F\). So the distance
from the evader to the exterior of the physical union is positive, which
is (i); at \(t=0\) it holds because \(F_{z_0}(0)>0\).

For the bounds in \eqref{eq:shiftedlaterroot}, put
\(v=\sqrt{(A-t_1)/(R-t_1)}\). Then \(\mu k<v<1\), the first inequality being
\(t_1<\Delta\), and
\[
 \widetilde T_L'(t_1)=\frac2m\Big[-1+\frac{\mu k}{2}\Big(v+\frac1v\Big)\Big],
 \qquad \frac{\mu k}{2}\Big(v+\frac1v\Big)<\frac{1+\mu^2k^2}{2}<1,
\]
because \(v+1/v\) is decreasing on \((0,1)\). Hence \(-2/m\le\widetilde T_L'<0\).

For (iii), the right-hand side of \eqref{eq:foldcomparison} is positive
for \(0\le\tau<\widetilde T_L(t_1)-t_1\) except at \(\tau=T_F-t_1\). At that horizon the
exterior set \(\mathcal X\) of the expanded union of \(\bar z\) satisfies
\(\operatorname{dist}(\bar E,\mathcal X)=\mu T_F\), attained only at the
contact corner \(\bar Q\) of \(\bar z\), and \(\bar Q-\bar E=\mu T_Fe_F\). If
\(F_{z(t_1)}(T_F-t_1)=0\), then
\(\operatorname{dist}(E(t_1),\mathcal X)=\mu T_F-\mu t_1\), so a point of
\(\mathcal X\) closest to \(E(t_1)\) lies within \(\mu T_F\) of \(\bar E\); it
is therefore \(\bar Q\), and equality in the triangle inequality gives
\(E(t_1)=\bar E+\mu t_1e_F\). In the fixed frame this reads
\(\int_0^{t_1}e(t)\,\dd t=t_1e_F\), and
\(\int_0^{t_1}(1-e_F\cdot e)\,\dd t=0\) forces \(e=e_F\) almost everywhere.
Conversely, if \(e=e_F\) almost everywhere on \([0,t_1]\), then
\[
 z(t_1)=\big((r_F-t_1)\sqrt{1-k^2},\ \mu(T_F-t_1)\sqrt{c_k},\
 k(A+mt_1)\big),
\]
which is \eqref{eq:foldkT} with \(T_F-t_1\) in place of \(T_F\), and
\eqref{eq:foldkTrange} holds with \(\Delta+mt_1/c_k\) in place of \(\Delta\).
Its later root is
\(T_L'=T_F-t_1+\frac2m\big[A+mt_1-\mu k\sqrt{R(A+mt_1)}\big]\), and
\[
 \frac{\dd}{\dd t_1}\big(t_1+T_L'\big)
 =2-\mu k\sqrt{\frac{R}{A+mt_1}}>0,
\]
because \(A+mt_1\ge A>\mu^2k^2R\) by \(\Delta>0\). Hence
\(t_1+T_L'\ge T_L\).
\end{proof}

\subsection{Proof of Theorem~\ref{thm:reverse}}
\label{sub:reverseproof}
\begin{proof}
\emph{Upper bound.} By Proposition~\ref{prop:bounds}, for every
\(\varepsilon>0\) a constant heading escapes by \(U(z)+\varepsilon\) against
all pursuer controls. A constant heading is a nonanticipative evader
strategy, so \(V_E\le U\).

\emph{Where \(U_0<U\).} On \([0,U(z)]\) we have \(F_z\ge0\). Suppose
\(F_z(\tau_0)=0\) for some \(\tau_0<U(z)\). If \(\tau_0=0\), the evader
starts on the sensing boundary with a positive escape time, which by
Section~\ref{sub:regions} happens only at nonusable double-boundary
points. If \(\tau_0>0\), then \(\tau_0\) is an interior minimum of \(F_z\)
with value zero. By Section~\ref{sub:algorithm} it is either a smooth
double root or, when \(Y=0\), the join kink, and these are the
minimum-fold states of Section~\ref{sub:fold}. At a cusp the zero
coalesces with the crossing, so \(U_0=U\) there.

\emph{Lower bound.} Steps~(a)--(d) prove \(V_E\ge U_0\), which is the claim
wherever \(U_0=U\). Steps~(e) and~(f) treat the nonusable double-boundary
points and the minimum-fold sheets. For the first four steps, if
\(U_0(z_0)=0\) there is nothing to prove, so assume \(U_0(z_0)>0\). Then
\(F_{z_0}(0)>0\), and the evader starts strictly inside the sensing union.

\emph{(a) A smaller sensing radius.} For small \(\rho>0\), write
\(U^\rho\), \(F^\rho\), and \(\widehat{\mathcal K}^\rho\) for the escape
time, the scalar function \eqref{eq:strictroot}, and the extended budget
set \eqref{eq:extendedbudget} of the game with sensing radius \(R-\rho\).
Lemma~\ref{lem:auxviability} holds for every sensing radius, and the
auxiliary dynamics \eqref{eq:auxdynamics} do not involve \(R\), so each
\(\widehat{\mathcal K}^\rho\) is a discriminating domain for
\eqref{eq:auxdynamics}. Since the expanded radius at horizon \(T\) is
\(R-\rho+T\),
\begin{equation}\label{eq:radiusshift}
 F^\rho_z(T)=F_z(T-\rho)-\mu\rho,
\end{equation}
where \(F_z(\tau)\) is extended to small negative \(\tau\) by the same
definition with expanded radius \(R+\tau\). The extension is continuous by
Lemma~\ref{lem:ties}.

\emph{(b) The approximate strategy theorem.} Let \(\mathcal A\) be a
nonanticipative evader strategy of the original game, and fix \(\delta>0\)
and a switching time step \(\Delta t>0\) with
\(\sqrt3\,\delta+2\Delta t<\rho\). For an auxiliary control
\(v=(C,V)(\cdot)\), let \(a(t)=a_0-\int_0^tC\). Define unit-speed pursuer
controls \(S_{\Delta t}[v]\) by
\[
 u_1=\big(\sigma\sqrt{1-V^2},\,V\big),\qquad
 u_2=\big(-\sigma\sqrt{1-V^2},\,V\big),
\]
with \(\sigma\in\{-1,1\}\) constant on each interval \([j\Delta t,(j+1)\Delta t)\), equal
to \(1\) if \(\widehat a(j\Delta t)\ge a(j\Delta t)\) and to \(-1\) otherwise, where
\(\widehat a(t)=a_0-\int_0^t\sigma\sqrt{1-V^2}\). Since
\(|C|\le\sqrt{1-V^2}\), the actual horizontal rate brackets the auxiliary
one, and induction over the intervals gives \(|\widehat a-a|\le2\Delta t\). The map
\(S_{\Delta t}\) is nonanticipative and does not involve the evader, so
\(\alpha=\mathcal A\circ S_{\Delta t}\) is a nonanticipative evader strategy of the
auxiliary game. Put \(s_0=U^\rho(z_0)\), so that
\((z_0,s_0)\in\widehat{\mathcal K}^\rho\). When the opponent of the player
who wants to remain in a discriminating domain plays a nonanticipative
strategy, that player can still remain within any distance \(\delta\) of
the domain on any finite interval; this is Theorem~3
of~\cite{cardaliaguetwp}, under the hypotheses checked in the proof of
Lemma~\ref{lem:auxviability}. Hence there is an auxiliary control \(v\)
whose trajectory against \(\alpha\) satisfies
\(\operatorname{dist}\big((z(t),s_0-t),\widehat{\mathcal K}^\rho\big)\le\delta\)
on \([0,s_0]\).

\emph{(c) Detection.} Let the pursuers play \(u=S_{\Delta t}[v]\) against
\(\mathcal A\). The evader control is \(\mathcal A[u]=\alpha[v]\), so the
evader path coincides with the auxiliary one. The pursuers share the
auxiliary midpoint and vertical motion, and their half-separation differs
from the auxiliary one by at most \(2\Delta t\). Let \(t<s_0-\delta\). Some
\((z',s')\in\widehat{\mathcal K}^\rho\) lies within \(\delta\) of
\((z(t),s_0-t)\); then \(s'>0\), so \(z'\) lies in the sensing union of
radius \(R-\rho\). The ranges \(\sqrt{(x\pm a)^2+y^2}\) change by at most
\(|\Delta a|+|\Delta x|+|\Delta y|\le\sqrt3\,|\Delta z|\), so one actual
range is at most \(R-\rho+\sqrt3\,\delta+2\Delta t<R\). The evader is therefore
detected on \([0,s_0-\delta)\), and
\(J_{z_0}(\mathcal A[u],u)\ge U^\rho(z_0)-\delta\). Since \(\mathcal A\) and
\(\delta\) are arbitrary, \(V_E(z_0)\ge U^\rho(z_0)\).

\emph{(d) The limit \(\rho\to0\).} Suppose that
\(\liminf_{\rho\to0}U^\rho(z_0)<T_*<U_0(z_0)\). Then there are
\(\rho_k\to0\) and \(T_k\in[0,T_*]\) with \(F^{\rho_k}_{z_0}(T_k)<0\), that is,
\(F_{z_0}(T_k-\rho_k)<\mu\rho_k\) by \eqref{eq:radiusshift}. A limit point
\(\bar T\in[0,T_*]\) of \(T_k-\rho_k\) satisfies \(F_{z_0}(\bar T)\le0\) by
continuity, so \(U_0(z_0)\le T_*\), a contradiction. Hence
\(V_E(z_0)\ge\liminf_{\rho\to0}U^\rho(z_0)\ge U_0(z_0)\).

\emph{(e) Nonusable double-boundary points.} Let \(z_0=(a_0,0,b)\) with
\(b=\sqrt{R^2-a_0^2}\ge0\) and \(\mu b<R\), so that \(U_0(z_0)=0\) and
\(s_0:=U(z_0)=2(R-\mu b)/(\mu^2-1)>0\). The evader is on the symmetry
axis below the upper corner, so
\(F_{z_0}(T)=\sqrt{(R+T)^2-a_0^2}-b-\mu T\). This function is strictly
concave, vanishes at \(T=0\) and \(T=s_0\), and is positive in between.

On a short interval \([0,t_1]\) both pursuers head for the evader's
initial position, which is at range \(R\) from each:
\[
 u_1=(a_0,b)/R,\qquad u_2=(-a_0,b)/R,\qquad
 0<t_1<\frac{Rs_0}{R+s_0}.
\]
This control is fixed in advance. Put \(\gamma=1-t/R\). At time \(t\) the
pursuers have covered the fraction \(t/R\) of the way, so their
half-separation is \(\gamma a_0\), and, relative to their midpoint, the
evader lies within \(\mu t\) of the reference point \((0,\gamma b)\). The
reference state \(\gamma z_0=(\gamma a_0,0,\gamma b)\) is \(z_0\) scaled by
\(\gamma\); it is a double-boundary point of the game with sensing radius
\(\gamma R=R-t\). Scaling all lengths and times by \(\gamma\) gives, in the
notation of step~(a),
\begin{equation}\label{eq:scaling}
 F^{t}_{\gamma z_0}(T)=\gamma\,F_{z_0}(T/\gamma),\qquad T\ge0.
\end{equation}
As in the proof of Lemma~\ref{lem:foldopening}, the expanded radius at
horizon \(\tau\) from \(z(t)\) is \(R+\tau=(R-t)+(\tau+t)\), and the distance
to a closed set is \(1\)-Lipschitz, so for every evader control
\begin{equation}\label{eq:openingphase}
 F_{z(t)}(\tau)\ \ge\ F^{t}_{\gamma z_0}(\tau+t)
 =\gamma\,F_{z_0}\Big(\frac{\tau+t}{\gamma}\Big),\qquad\tau\ge0.
\end{equation}
The right-hand side is positive when \(0<\tau+t<\gamma s_0\). For
\(0<t\le t_1\) we have \(t<\gamma s_0\), by the choice of \(t_1\). Taking
\(\tau=0\) shows that the evader is strictly inside the sensing union on
\((0,t_1]\), and taking \(t=t_1\) gives
\[
 U_0(z(t_1))\ \ge\ \Big(1-\frac{t_1}{R}\Big)s_0-t_1 .
\]
The case \(b=0\), in which the contact residual is not differentiable in
the state, needs no separate treatment.

Now let \(\mathcal A\) be a nonanticipative evader strategy and \(u^1\)
the control above. Since \(u^1\) is fixed in advance, the evader's control
on \([0,t_1]\) is \(\mathcal A[u^1]\) there, and \(z(t_1)\) is determined.
The map \(u'\mapsto\mathcal A[u^1\oplus u'](t_1+\cdot)\), where
\(u^1\oplus u'\) follows \(u^1\) on \([0,t_1]\) and \(u'(\cdot-t_1)\)
afterwards, is a nonanticipative evader strategy from \(z(t_1)\). By
steps~(a)--(d) there is a pursuer control \(u'\) whose escape time from
\(z(t_1)\) is at least \(U_0(z(t_1))-\varepsilon\). Since the evader has not
escaped on \([0,t_1]\), the concatenated control gives
\(J_{z_0}\ge t_1+U_0(z(t_1))-\varepsilon\ge s_0-s_0t_1/R-\varepsilon\). Letting
\(t_1,\varepsilon\to0\) gives \(V_E(z_0)\ge U(z_0)\).

\emph{(f) Minimum-fold states.} Let \(z_0\) be a minimum-fold state; by
reflection \(x,y\ge0\), and \(U(z_0)=T_L\). Let \(\mathcal A\) be a
nonanticipative evader strategy, fix \(T\) with \(T_F<T<T_L\), and put
\(\varepsilon=T_L-T\). Let \(u^0\) be the constant control with headings
\(n_{1F}\), \(n_{2F}\) of Lemma~\ref{lem:foldopening}, and let
\(e^0=\mathcal A[u^0]\). Choose inspection times
\[
 t_0=0,\qquad t_{j+1}=t_j+\min\big\{\bar\ell,\tfrac12(T_F-t_j)\big\},\qquad
 \bar\ell=\min\big\{\tfrac12\Delta,\tfrac18m\varepsilon\big\},
\]
so that \(t_j\uparrow T_F\). The pursuers do not switch at these times;
they only serve to locate where \(e^0\) first differs from \(e_F\).

Suppose that \(e^0=e_F\) almost everywhere on \([0,t_j]\). Applying
Lemma~\ref{lem:foldopening}(i) and~(iii) on the successive intervals
\([t_i,t_{i+1}]\), \(i<j\), whose lengths are smaller than the current
contact time and the current \(\Delta\), shows that the evader is detected
on \([0,t_j]\) and that \(z(t_j)\) is a minimum-fold state with the same
\(k\), hence the same headings \(n_{iF}\), contact time \(T_F-t_j\), and a
later root \(T_L^j\) with
\begin{equation}\label{eq:foldnoloss}
 t_j+T_L^j\ \ge\ T_L.
\end{equation}

\emph{Case 1: \(e^0\ne e_F\) on a set of positive measure in \([0,T_F]\).}
Let \(j\) be the first index for which this happens in \([t_j,t_{j+1}]\),
and put \(\ell_j=t_{j+1}-t_j\). Lemma~\ref{lem:foldopening}, applied at
\(z(t_j)\), gives detection on \([t_j,t_{j+1}]\) and, by (iii), a strictly
positive contact residual from \(z(t_{j+1})\) up to the horizon
\(\widetilde T_{L,j}(\ell_j)-\ell_j\), where \(\widetilde T_{L,j}\) is \eqref{eq:shiftedlaterroot} for the state
\(z(t_j)\). Hence, by \eqref{eq:shiftedlaterroot} and \eqref{eq:foldnoloss},
\[
 t_{j+1}+U_0(z(t_{j+1}))\ \ge\ t_j+\widetilde T_{L,j}(\ell_j)
 \ \ge\ t_j+T_L^j-\frac{2\ell_j}{m}\ \ge\ T_L-\frac\varepsilon4 .
\]
As in step~(e), the map
\(u'\mapsto\mathcal A[u^0|_{[0,t_{j+1}]}\oplus u'](t_{j+1}+\cdot)\) is a
nonanticipative evader strategy from \(z(t_{j+1})\), and by
steps~(a)--(d) there is a control \(u'\) whose escape time from
\(z(t_{j+1})\) is at least \(U_0(z(t_{j+1}))-\varepsilon/4\). The concatenated
control \(u\) leaves the evader's response on \([0,t_{j+1}]\) unchanged and
gives \(J_{z_0}(\mathcal A[u],u)\ge T_L-\varepsilon/2>T\).

\emph{Case 2: \(e^0=e_F\) almost everywhere on \([0,T_F]\).} The evader is
detected on \([0,T_F)\), and at time \(T_F\) it stands on the corner of the
two physical disks:
\[
 z(T_F)=\big(R\sqrt{1-k^2},\,0,\,Rk\big),
\]
a double-boundary point with \(b=Rk\), nonusable because \(\mu k<1\). By
\eqref{eq:nonusableboundaryU} and \eqref{eq:foldnoloss}, in the limit
\(t_j\to T_F\),
\[
 T_F+U(z(T_F))=T_F+\frac{2R(1-\mu k)}{m}=\lim_{j\to\infty}\big(t_j+T_L^j\big)
 \ \ge\ T_L.
\]
Step~(e), applied to the evader strategy
\(u'\mapsto\mathcal A[u^0|_{[0,T_F]}\oplus u'](T_F+\cdot)\) from \(z(T_F)\),
gives a control \(u'\) with escape time at least
\(U(z(T_F))-\varepsilon/2\), and the concatenated control gives
\(J_{z_0}(\mathcal A[u],u)\ge T_L-\varepsilon/2>T\).

In both cases \(\sup_uJ_{z_0}(\mathcal A[u],u)\ge T\). Since \(T<T_L\) and
\(\mathcal A\) are arbitrary, \(V_E(z_0)\ge T_L=U(z_0)\).
\end{proof}

\setlength{\bibsep}{1pt plus 1pt}
\bibliographystyle{elsarticle-num}
\bibliography{refs}
\end{document}